\documentclass[conference,compsoc]{IEEEtran}

\usepackage{caption}
\usepackage{amsmath,amsfonts, amssymb}
\usepackage{subfigure}
\usepackage{hyperref}
\usepackage{url}

\usepackage{graphicx}
\usepackage{textcomp}
\usepackage{xcolor}
\usepackage{amsthm}
\usepackage{colortbl}
\usepackage[misc]{ifsym}
\usepackage{xspace}
\usepackage{multirow}
\usepackage{threeparttable}
\usepackage{subfigure}
\usepackage{booktabs}
\usepackage{tabularx}

\usepackage{enumitem}
\usepackage{booktabs}
\usepackage[linesnumbered,ruled,vlined]{algorithm2e}
\usepackage[normalem]{ulem}
\usepackage{dblfloatfix}	
\usepackage[euler]{textgreek}

\usepackage{framed}
\definecolor{formalshade}{rgb}{0.95,0.95,0.97}
\definecolor{darkblue}{rgb}{0.14,0.22,0.52}
\newenvironment{formal}{
  
  \MakeFramed{\advance\hsize-\width\FrameRestore}
  \noindent\hspace{-4.55pt}
}
{
  \endMakeFramed%
}

\SetKwInput{KwInput}{Input}                
\SetKwInput{KwOutput}{Output}              

\def\eg{\emph{e.g.,}\xspace}

\def\ie{\emph{i.e.,}\xspace}
\def\etal{\emph{et al.}\xspace}
\def\vs{\emph{vs.}\xspace}

\newcommand{\todo}[1]{}
\renewcommand{\todo}[1]{{\color{red} TODO: {#1}}}

\SetCommentSty{mycommfont}

\usepackage{cleveref}
\crefformat{section}{\S\textcolor{purple}{#2#1#3}} 
\crefformat{subsection}{\S\textcolor{purple}{#2#1#3}}
\crefformat{subsubsection}{\S\textcolor{purple}{#2#1#3}}
\crefformat{figure}{#2{\color{blue}#1}#3}
\crefformat{table}{#2{\color{blue}#1}#3}
\crefformat{appendix}{Appendix \textcolor{purple}{#2#1#3}}

\newcommand{\tool}{\textsc{CAUSE}\xspace}

\begin{document}

\title{Edge Unlearning is Not ``on Edge''!\\ An Adaptive Exact Unlearning System on Resource-Constrained Devices}
\author{
\IEEEauthorblockN{Xiaoyu Xia\IEEEauthorrefmark{1}, Ziqi Wang\IEEEauthorrefmark{1}, Ruoxi Sun\IEEEauthorrefmark{2}, Bowen Liu\IEEEauthorrefmark{3}, Ibrahim Khalil\IEEEauthorrefmark{1}, Minhui Xue\IEEEauthorrefmark{2}}

\IEEEauthorblockA{\IEEEauthorrefmark{1} RMIT University, Australia \\ \IEEEauthorrefmark{2} CSIRO’s Data61, Australia \\ \IEEEauthorrefmark{3} Nanjing University, China
}
}

\maketitle

\begin{abstract}
The right to be forgotten mandates that machine learning models enable the erasure of a data owner's data and information from a trained model.
Removing data from the dataset alone is inadequate, as machine learning models can memorize information from the training data, increasing the potential privacy risk to users. To address this, multiple machine unlearning techniques have been developed and deployed. Among them, approximate unlearning is a popular solution, but recent studies report that its unlearning effectiveness is not fully guaranteed. 
Another approach, exact unlearning, tackles this issue by discarding the data and retraining the model from scratch, but at the cost of considerable computational and memory resources. However, not all devices have the capability to perform such retraining. In numerous machine learning applications, such as edge devices, Internet-of-Things (IoT), mobile devices, and satellites, resources are constrained, posing challenges for deploying existing exact unlearning methods.
In this study, we propose a \underline{C}onstraint-aware \underline{A}daptive \underline{E}xact \underline{U}nlearning \underline{S}ystem at the network \underline{E}dge (CAUSE), an approach to enabling exact unlearning on resource-constrained devices. 
Aiming to minimize the retrain overhead by storing sub-models on the resource-constrained device, CAUSE innovatively applies a Fibonacci-based replacement strategy and updates the number of shards adaptively in the user-based data partition process. To further improve the effectiveness of memory usage, CAUSE leverages the advantage of model pruning to save memory via compression with minimal accuracy sacrifice. 
The experimental results demonstrate that CAUSE significantly outperforms other representative systems in realizing exact unlearning on the resource-constrained device by 9.23\%-80.86\%, 66.21\%-83.46\%, and 5.26\%-194.13\% in terms of unlearning speed, energy consumption, and accuracy. 
\end{abstract}

\section{Introduction}
\label{sec:introduction}

Machine learning models, particularly Deep Neural Networks (DNNs), often rely on a large amount of data collected from individuals~\cite{de2021critical}. 
However, the over-collection of private information has become a significant concern for regulators in numerous countries, leading to the implementation of related laws and regulations, such as the American California Consumer Privacy Act (CCPA)~\cite{harding2019understanding} and the European Union's General Data Protection Regulation (GDPR)~\cite{mantelero2013eu}.
Privacy regulations require service providers to delete users' training data upon request, known as \textit{the Right to be Forgotten}. 
Unlike straightforward data deletion from a database, removing data from deep models is complex due to their intricacies and the randomness inherent in training algorithms~\cite{thudi2022unrolling}. Evidence demonstrates that knowledge of data used in the training process may persist within the model, as shown by the impact of specific attacks like membership inference attacks~\cite{maini2024tofu}. This has prompted a new research topic focusing on model privacy, \ie machine unlearning, to comply with the mentioned regulations. Machine unlearning aims to assist service providers in effectively removing users' data, including their contributions to the model during the training process. Specifically, machine unlearning techniques aim to safeguard privacy by removing specific data samples from trained models upon request or after a set timescale. 

Currently, there exist two major research approaches for model unlearning: exact unlearning~\cite{bourtoule2021machine, chen2022recommendation, yan2022arcane} and approximate unlearning~\cite{gupta2021adaptive, lin2023erm}. 
Exact unlearning involves directly removing the data to be unlearned and retraining the model from scratch, known as naive retraining. While effective, this method often incurs significant cost overhead, such as high computational requirements and increased memory consumption. 
In contrast, approximate unlearning adjusts model parameters to eliminate the contribution of specific data, simulating the effects of retraining. While approximate unlearning offers cost advantages, security flaws raise concerns about its reliability~\cite{shumailov2021manipulating, thudi2022necessity} and vulnerability to attacks~\cite{hu2024duty, hu2024learn}. A recent study~\cite{maini2024tofu} discovered that representative approximate unlearning methods fail to guarantee the removal of learned data, indicating that models unlearned using these systems still retain the knowledge of the forgotten data to some extent. These findings corroborate concerns discussed at the AI Seoul Summit~\cite{bengio2024international}, 
co-hosted by the Republic of Korea and the United Kingdom, regarding the limitations of approximate learning systems. This highlights a significant gap that challenges the feasibility of deploying approximate unlearning in the industry.

The proliferation of data-intensive applications, including machine learning models, and the increasing demand for low-latency highlight the importance of shifting services from the cloud to resource-constrained devices at the network edge~\cite{taleb2017multi, xia2020online}, such as AI-powered satellites~\cite{erwin2024saic}, edge devices, Internet-of-Things, and mobile devices.
For example, Europe's plan to launch over 1,000 new satellites annually aims to enhance its computation capabilities in space~\cite{mattioli2023identifying}. These satellites and other auxiliary devices capture vast amounts of data that fuel critical applications such as weather forecasting and traffic monitoring. Yet, challenges arise when data, especially sensitive information such as that captured by satellites during conflicts like the war in Ukraine~\cite{borowitz2022war}, needs to be selectively forgotten from machine learning models to comply with privacy demands, necessitating the urgent deployment of machine unlearning systems on resource-constrained devices.
However, these devices, particularly satellites, face serious vulnerability to manmade threats~\cite{willbold2023space}, also declared by the European Union Agency for Cybersecurity~\cite{mattioli2023identifying}. To ensure effectiveness and reliability, especially considering the evidence of existing limitations and vulnerabilities in approximate unlearning, exact unlearning should be prioritized for handling unlearning requests on resource-constrained devices at the network edge. This involves retraining the model with the requested data directly removed from the training datasets.
However, resource-constrained devices cannot easily and flexibly autoscale like cloud servers to handle computation- or memory-intensive workloads. This makes implementing machine unlearning methods extremely challenging on such devices, for the following three main reasons:
\begin{itemize}[leftmargin=*, noitemsep]
    \item \textbf{Limited resources.} Memory consumption is always a major challenge of exact unlearning systems~\cite{bourtoule2021machine, chen2022recommendation, yan2022arcane} since multiple sub-models have to be stored in the memory. This becomes more serious at resource-constrained devices, \eg on satellites, edge devices, Internet-of-Things, and mobile devices. 
    Taking the state-of-the-art edge computing service, Amazon Wavelength~\cite{amazon_wavelength}, as an example, the highest configuration offers computational resources such as 8 CPUs, 1 GPU, and 32GB memory, which can be easily exhausted when retraining machine learning models.
    In addition, multiple service providers can reserve memory resources on those devices simultaneously~\cite{xia2022formulating, xia2021constrained}, further lowering the available memory resources for saving model parameters for performing exact unlearning.
    \item \textbf{Energy Consumption.} Energy is another significant challenge. For instance, IoT devices typically have small battery capacities; devices with higher energy capacities, like satellites, also have a limited volume of energy harvested, such as solar power. However, exact unlearning requires a significant amount of energy to retrain models. This extensive demand for energy makes implementing exact unlearning on resource-constrained devices particularly problematic.    \item \textbf{Dynamic nature of the edge network environment.} In the dynamic network edge, the departure and arrival of users or specific data captures may involve unpredictable and dynamic learning and unlearning requests. It is difficult for resource-constrained devices to handle dynamic requests since various users may request their partial data to be forgotten at any time. 
\end{itemize}

In this study, we present \underline{C}onstraint-aware \underline{A}daptive Exact \underline{U}nlearning \underline{S}ystem at the network \underline{E}dge, \textbf{\tool}, a novel adaptive exact unlearning approach to enable \textit{the right to be forgotten} on resource-constrained devices, providing privacy protection while ensuring model performance with low energy consumption. 
Motivated by pilot experiment results, we reveal the relationship between the retraining process and resource consumption (as shown in Figure~\ref{fig:linear_energy_rsn}). Specifically, we observe that controlling the number of retraining samples plays a critical role in reducing energy consumption and increasing retraining efficiency. Based on this observation, we have innovatively designed \tool with four main modules, including:
\begin{itemize}[leftmargin=*,noitemsep]
    \item \textbf{User-centered data partition (UCDP).}
    \tool employs the UCDP strategy to divide the training data into approximately balanced shards.
    Existing data partitioning methods for machine unlearning are not directly applicable at the network edge due to the heterogeneity of data collected from multiple users and the specificity of unlearning requests to individual user data. Inspired by class-based data partition in~\cite{yan2022arcane}, UCDP innovatively partitions data based on the origin, \ie users. In this way, UCDP allows \tool to locate the retraining shard directly based on the origin of data, and improves the efficiency of \tool in unlearning since the data from the same user is partitioned in the same shard. 
    \item \textbf{Resource-constrained model pruning (RCMP).} To reduce the memory consumption for storing more sub-models at different learning points, \tool employs RCMP to iteratively prune the trained sub-models, which significantly decreases the model  size while introducing only a slight decrease in the accuracy, \eg $>$50\% memory reduction with $<$3\% accuracy decrease (Table~\ref{tab:pruning}).
    \item \textbf{Fibonacci-based replacement (FiboR).} Although \tool employs model pruning to reduce the memory usage for sub-models, the memory capacity of resource-constrained devices still struggles in storing trained sub-models over time. This challenge arises from the unpredictability of future learning or unlearning requests. Unlike most studies on unlearning that overlook the necessity of memory management, \tool incorporates a novel FiboR approach for memory optimization, strategically replacing older sub-models with newly trained ones in a non-linear ``jumping'' fashion to manage memory space more efficiently.
    \item \textbf{Shard controller (SC).} With limited memory resources, frequent replacement operations are inevitable over time, leading to substantial retraining overheads, especially when the sub-models that were closely trained on the data designated for unlearning have been replaced. To overcome this shortage, \tool employs SC to dynamically adjust the number of trained sub-models, \ie the number of shards, to manage resource usage effectively over time. This approach further boosts the efficiency of \tool in unlearning without sacrificing accuracy.
\end{itemize}

To the best of our knowledge, \tool is the \textit{first} mechanism designed to achieve exact unlearning on resource-constrained devices by improving retraining efficiency and reducing energy consumption, while taking into account constrained memory resources and dynamic changes in the network over time.
We believe that \tool provides a valuable step forward in advancing exact unlearning frameworks on resource-constrained devices.

\section{Preliminaries}

\noindent\textbf{Model update requests}. Let $\mathcal{D}$ be the dataset comprising elements from a specified data domain $\Re$. The machine learning model update requests can be divided into two types, such as addition and deletion. These update requests are formally defined below, similar to how they are defined in~\cite{gupta2021adaptive, neel2021descent}. 
Let $\varsigma = (\phi, d)$ be the update request where $\phi$ indicates the type of update —-- either `add' or `delete' from the set $\Phi = \{add, delete\}$, and $d \in \Re$ represents the subset of data from $\Re$ to be either added to or removed from the dataset. In this framework, an $add$ request corresponds to a task of learning with new training data, while a $delete$ request corresponds to the task of unlearning and removing data. Consider a sequence of such update requests, denoted by $\zeta = \{ \varsigma_1, \varsigma_2, \varsigma_3, ... \}$ where $\varsigma_i \in \Phi \times \Re$ for request index $i \in I$. The process of updating the model with a single request is denoted by $\mathcal{D} \circ \varsigma$ and when applied sequentially across a series of updates, the dataset evolves as  
$\mathcal{D} \circ \zeta = ((((\mathcal{D} \circ \varsigma_1)\circ \varsigma_2)\circ \varsigma_3)\circ ...)$.
This sequential operation ensures that each update request is applied to the dataset in turn, modifying the dataset progressively. Upon processing all updates in the sequence $\zeta$, the updated model can be obtained, \ie $\mathcal{M} = \mathcal{K}(\mathcal{D} \circ \zeta)$, where $\mathcal{K}(\cdot)$ is the model training algorithm, such as ResNet~\cite{he2016deep}, VGG~\cite{simonyan2015very}, and MobileNetV2~\cite{sandler2018mobilenetv2}.

\begin{figure}[ht]
\centering
\vspace{-1 em}
\includegraphics[width=0.65\linewidth]{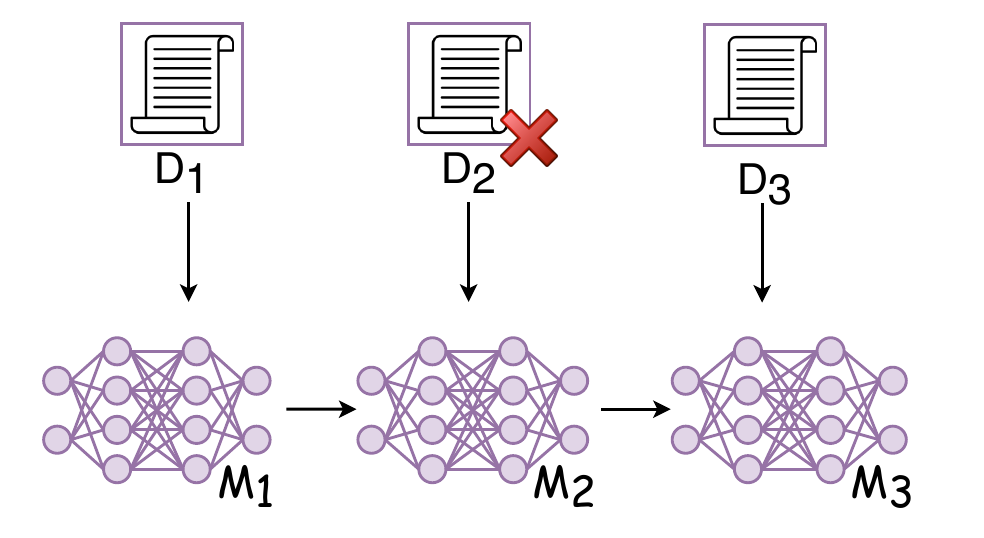}
\vspace{-.5 em}
\caption{Machine learning and unlearning}
\label{fig:learning_unlearning}
\vspace{-.5 em}
\end{figure}

\noindent\textbf{Machine unlearning}. At the network edge, update requests arrive sequentially and are processed in order. As each update request arrives, the update algorithm $\mathcal{U}_\mathcal{L}$ modifies the state of the machine learning model. These requests typically originate from users. When a user issues a machine unlearning request to remove a specific data subset $\mathcal{D}_r \in \mathcal{D}$, the device executes a data deletion operation $\phi_{delete}$. This operation modifies the training dataset to $\mathcal{D}'$, effectively removing $\mathcal{D}_r$ from it, \ie
$\mathcal{D}' \leftarrow \mathcal{D} \cup \neg \mathcal{D}_r$.
The objective of machine unlearning is to eliminate any influence that the unlearned data subset $\mathcal{D}_r$ had on the model $\mathcal{M}$. This is achieved by retraining the model to create a new model $\mathcal{M}'$:
$\mathcal{M}' = \mathcal{K}( (\mathcal{D} \circ \zeta) \circ \varsigma') = \mathcal{K}(\mathcal{D} \cup \neg \mathcal{D}_r \circ \zeta)$,
where $\varsigma' = \mathcal{D}_r \times \phi_{delete}$.
In cases requiring exact unlearning, a straightforward way involves retraining the model from scratch using the modified dataset $\mathcal{D}'$. For example, deploying a deep learning model like ResNet directly on $\mathcal{D}'$ ensures that the new model $\mathcal{M}'$ behaves as if it had never been exposed to the unlearned data subset $\mathcal{D}_r$. This process allows for the model to be refreshed and updated without retaining any information of the data specified for removal, aligning with privacy requirements and ensuring the model's integrity in dynamic environments.

\section{System Design Motivation}
\label{sec:design_motivation}

As discussed in \S\ref{sec:introduction}, implementing exact unlearning on resource-constrained devices poses significant challenges due to the limited resources, energy consumption, and dynamic nature of learning and unlearning requests. To address these challenges effectively, it is crucial to understand the relationship between the retraining process and the resources. Building on this understanding, we can then design a tailored exact unlearning system optimized for resource-limited environments.

Given the shard number $\mathcal{S}$, existing exact unlearning systems, such as SISA~\cite{bourtoule2021machine} and ARCANE~\cite{yan2022arcane} only store $\mathcal{S}$ sub-models. However, this is inefficient in dynamic network edge environments. As depicted in Figure~\ref{fig:learning_unlearning}, a device on the network edge receives training data over time, \ie $\mathcal{D}_1$, $\mathcal{D}_2$, and $\mathcal{D}_3$, sequentially from a user. Within the frameworks of SISA and ARCANE, a newly trained model supersedes the previous one, with $\mathcal{M}_2$ replacing $\mathcal{M}_1$, and subsequently, $\mathcal{M}_3$ replacing $\mathcal{M}_2$, leaving only $\mathcal{M}_3$ in memory. If a request is made to unlearn data $\mathcal{D}_r$ from $\mathcal{D}_2$, the model need to be retrained on datasets $\mathcal{D}_1 \cup \mathcal{D}_2 \cup \mathcal{D}_3 \cup \neg \mathcal{D}_r$. Retaining the model from $\mathcal{M}_1$ could eliminate the need to retrain any data from $\mathcal{D}_1$, showcasing a potential efficiency gain.

\begin{figure}[ht]
\vspace{-1 em}
\begin{minipage}{\linewidth}
\centering
\begin{minipage}{0.45\linewidth}
    \subfigure[Retraining Time \vs $\mathcal{B}$]{
    \includegraphics[width=0.9\linewidth]{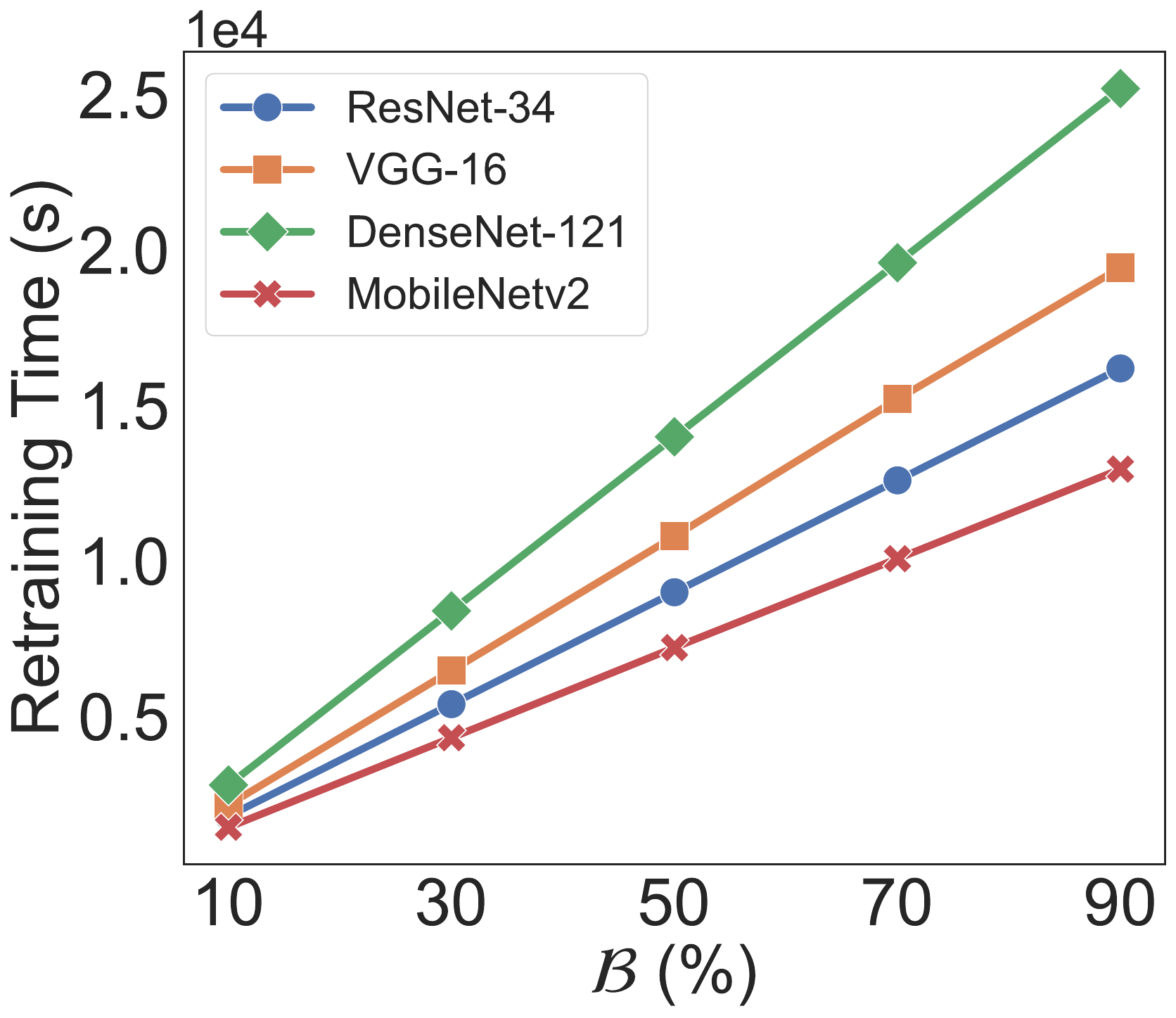}
    \label{fig:linear_time_rsn}}
\end{minipage}\hspace{0.8 em}
\begin{minipage}{0.45\linewidth}
    \subfigure[Energy Consumption \vs $\mathcal{B}$]{
    \includegraphics[width=0.9\linewidth]{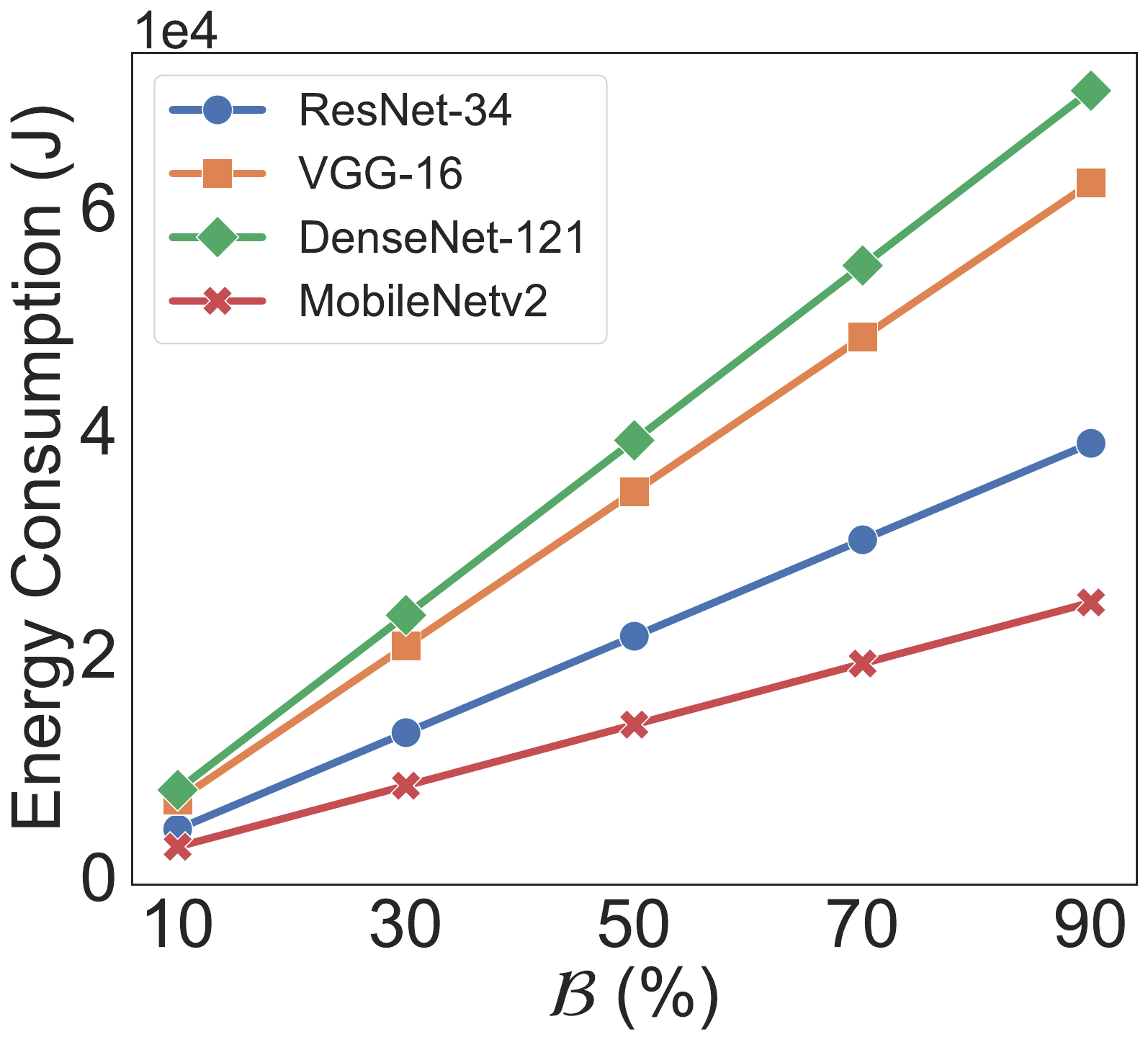}\label{fig:linear_energy_rsn}}
\end{minipage}

\vspace{-0.5 em}
\caption{Impact of various retraining ratios on retraining time and energy consumption.
}

\end{minipage}
\end{figure}

Here, we evaluated the performance impact of various retraining ratios $\mathcal{B}$, i.e., the ratio of the retraining sample number over the total sample number, through experiments using ResNet-34, VGG-16, DenseNet-121, and MobileNetV2 trained on the CIFAR-10 dataset with NVIDIA Jetson Orin Nano.  
Figure~\ref{fig:linear_time_rsn} demonstrates a linear relationship between the processing time and the number of retraining samples across all models. This correlation suggests that reducing the number of retraining samples can significantly decrease processing time, enhancing the overall efficiency of the system. Interestingly, the energy consumption is also linearly related to the number of retraining samples, as shown in Figure~\ref{fig:linear_energy_rsn}.

\smallskip
\noindent\textbf{Remark.} The discussion and observations underscore the pivotal role of reducing the number of retraining samples by optimizing memory usage to realize the exact unlearning on resource-constrained devices. By strategically minimizing the number of retraining samples required upon an unlearning request, we can significantly accelerate the retraining process and reduce energy consumption. This is the key motivation behind the design of \tool in \S\ref{sec:cause}.

\section{\tool System}
\label{sec:cause}

In this section, we first overview \tool and then present the four major components of \tool, \ie the resource-constrained model pruning (RCMP), user-centered data partition (UCDP), Fibonacci-based replacement (FiboR), and shard controller (SC). The notations and symbols used in this paper are summarized in Table~\ref{tab:notations}.

\begin{table}[ht]
\vspace{-0.5 em}
\small
\renewcommand{\arraystretch}{1.}
\caption{{Summary of Notations}}

\label{tab:notations}
\centering
\begin{tabular}{p{1.8cm}p{5.4cm}}
\toprule

\textbf{Notation} & \textbf{Description}\\
\midrule

$\mathcal{D}$ & Dataset\\

$\mathcal{D}_s$ & Data subset $s$\\

$\mathcal{D}_r$ & unlearned data subset\\

$f(\cdot)$ & Function of Fibonacci Sequence\\

$\mathcal{M}_s$ & Sub-model $s$\\

$\mathcal{N}_{mem}$ & Normalized memory resource \\

$p, \gamma$ & Control parameters in the shard controller \\

$\mathcal{S}$ & Number of data shards \\

$\mathcal{S}_t$ & Number of data shards defined by the shard controller in round $t$ \\

$size(\cdot)$ & Function of size calculation\\

$t$ & Training round $t$ \\

$T$ & Set of training rounds \\

$\mathcal{U}$ & Set of users \\

$u_k$ & User $k$ \\

$\delta$ & Pruning rate\\

$\varsigma_i$ & Unlearning request $i$ \\

$\zeta$ & Set of unlearning requests \\

\bottomrule
\end{tabular}
\vspace{-1 em}
\end{table}

\subsection{System Overview}

\begin{figure}[t]
    \centering
    \includegraphics[width=1.\linewidth]{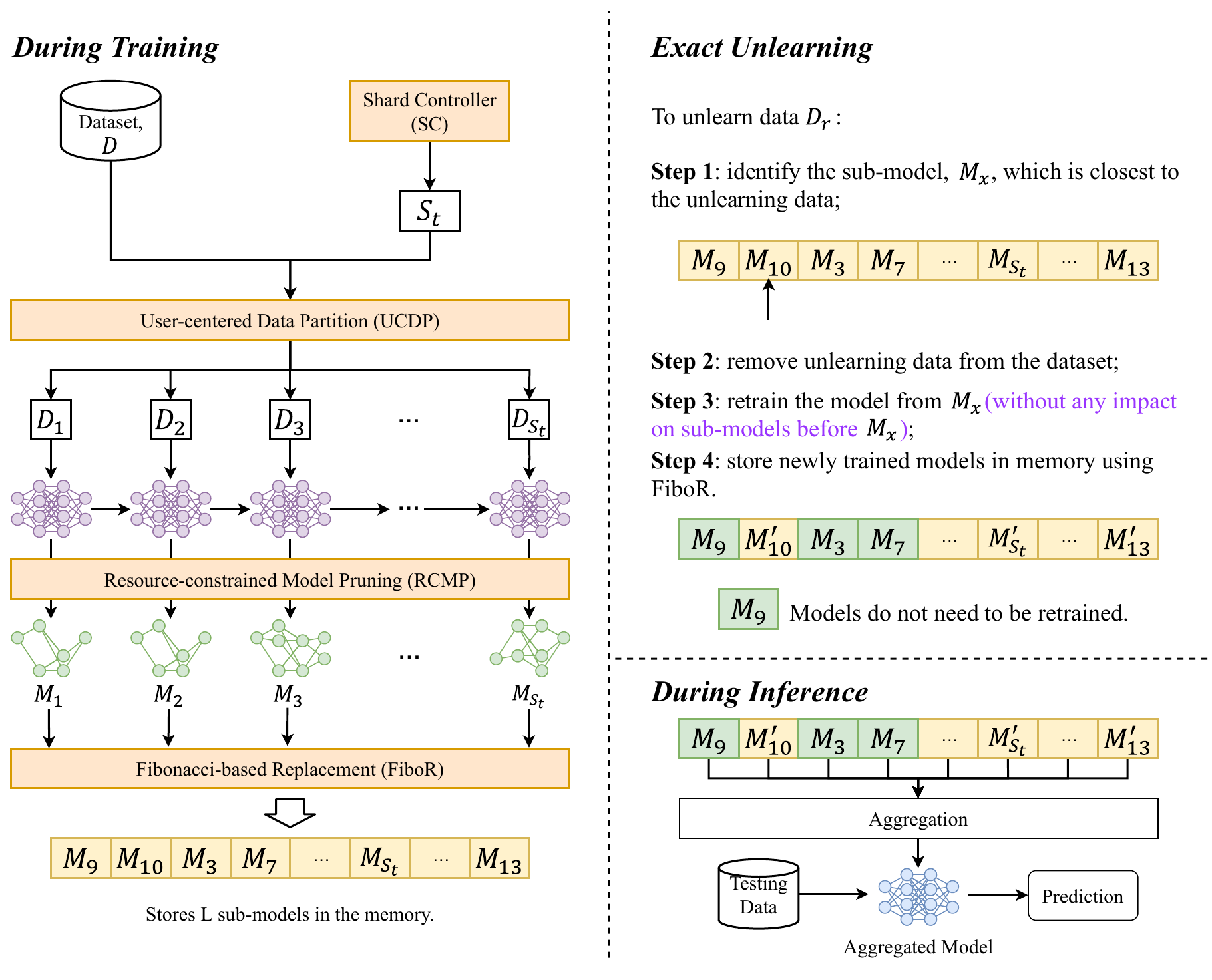}
    \caption{\tool system overview.}
    \label{fig:overview}
    \vspace{-1 em}
\end{figure}

The system overview of \tool is presented in 
Figure~\ref{fig:overview}. Similar to existing exact learning systems such as SISA~\cite{bourtoule2021machine} and ARCANE~\cite{yan2022arcane}, \tool trains multiple sub-models based on a set of disjoint shards (or subsets) of the dataset. At the network edge, the device usually receives training data from multiple users. The attributes of the data obtained from these users can exhibit significant or minimal variation. Furthermore, when a user initiates a request for data unlearning, the request generally focuses on removing the data associated with that user. Taking into account this fact in the resource-constrained exact unlearning, \tool implements the user-centered data partition (UCDP) algorithm to divide the training data into multiple data shards, organizing them according to their origin, namely, the specific users who supplied the data. 
Compared to the cloud and data center environment, the major challenge at the network edge is the limited resources, especially memory resources. To utilize the memory effectively, \tool prunes the trained sub-models by resource-constrained model pruning (RCMP) to reduce the memory needed for saving an individual sub-model. However, as time goes on, the memory would be exhausted by keeping storing sub-models. This is still a research gap in exact unlearning. To cover this gap, the Fibonacci-based replacement (FiboR) algorithm decides the way to replace previously trained sub-models with newly trained ones. Based on our observation of the relationship between the number of shards and accuracy, a shard controller (SC) is implemented in \tool to decrease the shard number based on the exponential weighted moving average (EWMA) function. We will now introduce RCMP~(\S\ref{subsec:pruning}), UCDP (\S\ref{subsec:data_partition}), FiboR (\S\ref{subsec:replacement}) and SC (\S\ref{subsec:shard_controller}) in detail.

\begin{figure*}
    \centering
    \includegraphics[width=0.8\linewidth]{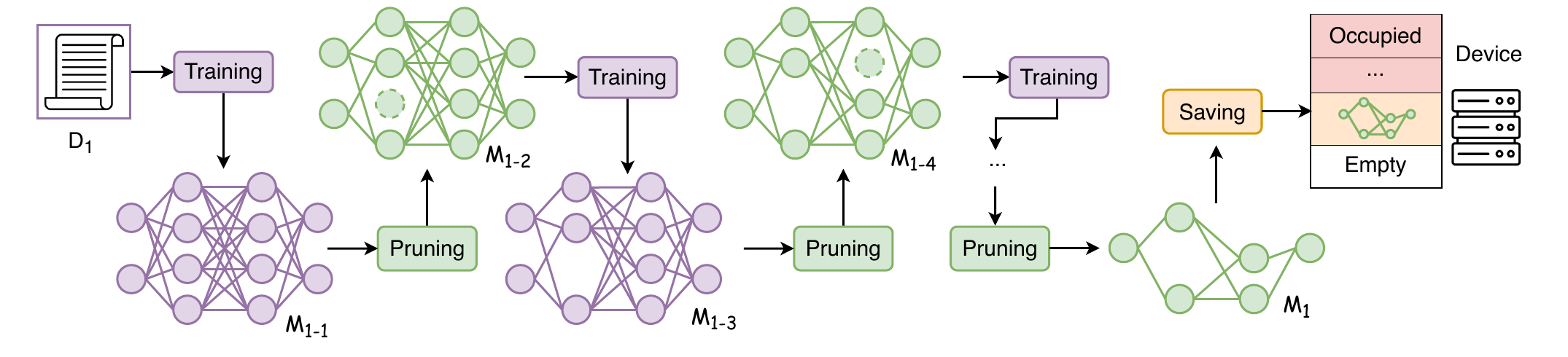}
    \vspace{-0.5 em}
    \caption{Resource-constrained model pruning.}
    \label{fig:pruning}
    \vspace{-.5 em}
\end{figure*}

\begin{table*}[!t]
\centering
\renewcommand{\arraystretch}{1.25}
\caption{Evaluations on Model Performance with Different Pruning Rates}
\label{tab:pruning}
\resizebox{\textwidth}{!}{%
\begin{tabular}{lccccccccccccc}
\toprule
\multirow{2}{*}{Baselines} & \multirow{2}{*}{Dataset} & \multirow{2}{*}{PR (\%)} & \multicolumn{3}{c}{Accuracy (\%)} & \multicolumn{3}{c}{Params (M)} & \multicolumn{3}{c}{Model File Size (MB)} & \multicolumn{2}{c}{Time(s)} \\
\cmidrule(r){4-6} \cmidrule(lr){7-9} \cmidrule(lr){10-12} \cmidrule(lr){13-14}
 & & & Original & Pruned & Degradation (\%) & Original & Pruned & Degradation (\%) & Original & Pruned & Degradation (\%) & Retrain & Prune \\
\midrule
\multirow{5}{*}{VGG-16} & \multirow{5}{*}{CIFAR-10} & 10 & 67.40 & 65.84 & 2.315 & 15.05 & 12.040 & 19.984 & 53.02 & 48.9879 & 7.601 & 750.09 & 0.5729 \\
& & 30 & 67.40 & 66.00 & 2.077 & 15.05 & 9.036 & 39.952 & 53.02 & 40.8060 & 23.037 & 716.54 & 0.4045 \\
& & 50 & 67.40 & 66.13 & 1.884 & 15.05 & 6.031 & 59.921 & 53.02 & 31.1364 & 41.274 & 749.84 & 0.8214 \\
& & 70 & 67.40 & 64.66 & 4.065 & 15.05 & 3.026 & 79.889 & 53.02 & 19.7250 & 62.797 & 750.31 & 0.4046 \\
& & 90 & 67.40 & 43.40 & 35.608 & 15.05 & 0.171 & 98.862 & 53.02 & 5.3768 & 89.859 & 749.84 & 0.4103 \\
\midrule
\multirow{5}{*}{ResNet-34} & \multirow{5}{*}{CIFAR-10} & 10 & 71.92 & 70.76 & 1.613 & 23.61 & 15.7276 & 33.386 & 85.82 & 65.3355 & 21.205 & 749.22 & 2.2910 \\
& & 30 & 71.92 & 71.69 & 0.319 & 23.61 & 12.4481 & 47.276 & 85.82 & 56.3608 & 32.028 & 739.90  & 2.0324 \\
& & 50 & 71.92 & 72.56 & +0.890 & 23.61 & 9.1686 & 61.166 & 85.82 & 45.5214 & 45.101 & 753.44 & 2.0485 \\
& & 70 & 71.92 & 72.75 & +1.154 & 23.61 & 4.7748 & 79.776 & 85.82 & 30.1478 & 63.641 & 746.37 & 2.1464 \\
& & 90 & 71.92 & 57.42 & 20.161 & 23.61 & 0.3144 & 98.668 & 85.82 & 8.4376 & 89.824 & 737.20 & 4.8679 \\
\midrule
\multirow{5}{*}{DenseNet-121} & \multirow{5}{*}{CIFAR-100} & 10 & 56.83 & 57.02 & +0.334 & 7.14 & 5.670 & 20.588 & 26.24 & 21.7729 & 17.024 & 971.86 & 5.2532 \\
& & 30 & 56.83 & 56.36 & 0.827 & 7.14 & 4.275 & 40.119 & 26.24 & 17.5123 & 33.261 & 968.31 & 5.3697 \\
& & 50 & 56.83 & 56.44 & 0.686 & 7.14 & 2.881 & 54.650 & 26.24 & 13.0240 & 50.366 & 983.78 & 4.6839 \\
& & 70 & 56.83 & 55.89 & 1.654 & 7.14 & 1.487 & 79.181 & 26.24 & 8.1287 & 69.022 & 957.20 & 5.0167 \\
& & 90 & 56.83 & 16.50 & 70.970 & 7.14 & 0.162 & 97.774 & 26.24 & 2.4860 & 90.525 & 982.17 & 5.3000 \\
\midrule
\multirow{5}{*}{\shortstack[l]{MobileNetV2}} & \multirow{5}{*}{CIFAR-10} & 10 & 78.79 & 80.86 & +2.614 & 2.18 & 1.738 & 20.254 & 7.71 & 7.2262 & 6.150 & 211.49 & 0.8017 \\
& & 30 & 78.79 & 80.91 & +2.677 & 2.18 & 1.327 & 39.134 & 7.71 & 6.1087 & 20.663 & 212.01 & 0.7976 \\
& & 50 & 78.79 & 80.55 & +2.221 & 2.18 & 0.915 & 58.015 & 7.71 & 4.7611 & 38.165 & 211.25 & 0.7997 \\
& & 70 & 78.79 & 79.46 & +0.838 & 2.18 & 0.504 & 76.895 & 7.71 & 3.1716 & 58.810 & 212.42 & 0.8011 \\
& & 90 & 78.79 & 10.00 & 87.310 & 2.18 & 0.113 & 94.831 & 7.71 & 1.1935 & 84.499 & 219.33 & 0.8047 \\
\bottomrule
\end{tabular}
}
\end{table*}

\subsection{Resource-Constrained Model Pruning}
\label{subsec:pruning}

Model pruning has emerged as a favored solution in machine learning to address the over-parameterization challenge in deep neural networks~\cite{liu2018rethinking}. The conventional process of model pruning involves three main phases: training, pruning, and fine-tuning. Generally, heuristic techniques are employed to prune the network model, like low-rank decomposition and sparsity regularization~\cite{liu2019metapruning}, to identify and remove insignificant network components. Originally, model pruning is designed as a practical approach to improving computational efficiency and achieving faster inference. Recently, machine learning studies~\cite{jiang2023model, jiang2022fedmp} in edge computing demonstrate the effectiveness of model pruning in saving memory on resource-constrained devices.

Figure~\ref{fig:pruning} shows an example of the resource-constrained model pruning (RCMP). Compared to conventional structure pruning methods, e.g., one-shot pruning in OMP, RCMP recursively restructures the pruning process by dividing it into a series of 'prune-and-retrain' sub-processes. This approach gradually achieves the desired model sparsity. In each sub-process, a small portion of the model's parameters is pruned, followed by a retraining phase to maintain structural stability and accuracy. This method prevents the model from being degraded by intensive one-shot pruning and enhances overall performance~\cite{zhu2017prune}.

Given a training dataset $\mathcal{D}_1$ existing on the device, the machine learning model $\mathcal{M}_1$, \eg a deep neural network model, is trained with a defined number of epochs. Let $\mathcal{M}_{1-1}$ denote the stage 1 of $\mathcal{M}_1$, \ie the model trained after the first epoch. Through various techniques, such as loss-based approaches or gradient-based approaches, the redundant or abnormal parameters, channels, or filters are identified, laying the groundwork for the pruning phase. After that, these identified factors are systematically removed from the model. Then, the updated model undergoes a fine-tuning step to mitigate the performance degradation resulting from the removal. As a result, a new model $\mathcal{M}_{1-2}$ is obtained. In this paper, we refer to the whole process of identification, removal, and fine-turning as the \textit{pruning} process, \eg the process from $\mathcal{M}_{1-1}$ to $\mathcal{M}_{1-2}$. This pruning process is executed every time after the training process. Finally, the trained and pruned model $\mathcal{M}_1$ is stored in the memory on the device for future unlearning. 

Typically, model pruning removes identified parameters by setting their values to 0. However, at the dynamic network edge, the memory resource on the device is usually limited. Thus, \tool deletes those identified parameters, channels, or filters to further reduce memory usage for saving more model parameters when an unlearning request arrives. To investigate the impact of model pruning, we conducted a set of experiments by running models VGG-16, ResNet-34 and MobileNetV2 on CIFAR10, and DenseNet-121 on CIFAR100, following~\cite{chen2021chasing,zhu2021vision}. Details can be found in Table~\ref{tab:exp_pruning} in Appendix~\ref{apx:pruning_settings}. In the experiments, we measured the performance of model pruning in terms of accuracy, parameter numbers, model file size, and pruning time, with various pruning rates $\delta = 10\%, 30\%, 50\%, 70\%, 90\%$. Detailed experimental results are demonstrated in Table~\ref{tab:pruning}. In terms of the reduction in model size including the number of parameters and file size, the pruning rate is nearly linear related to the size deduction for all the tested models. With the pruning rate $\delta = 70\%$, the model file size is reduced by 62.80-74.21\% for all the tested models. 
In addition, it can also be observed that there is no significant decrease in the overall accuracy of all the test models when the pruning rate is less than or equal to 70\%, \ie $\delta \leq 70\%$. In fact, in certain cases, such as with ResNet-34 at $\delta=50\%$ and $\delta=70\%$, accuracy slightly increases. Once the pruning rate arrives at 90\%, the degradation of the model accuracy becomes significant. 

According to the results in Table~\ref{tab:pruning}, adopting a suitable model pruning ratio can effectively reduce the file size of a model, thereby reducing memory usage with only a minor effect on the model's accuracy. Implementing this pruning strategy introduces some additional computational overhead, yet this increase is minimal, particularly in comparison to the overall time required for training the model. For instance, employing a 70\% pruning ratio results in the following computational overheads: 0.40 seconds for pruning versus 750.31 seconds for training a VGG-16 on CIFAR-10, 2.15 seconds for pruning versus 746.37 seconds for training a ResNet-34 on CIFAR-10, 7.02 seconds for pruning versus 957.20 seconds for training a DenseNet-121 on CIFAR-100, and 0.80 seconds for pruning versus 212.42 seconds for training a MobileNetV2 on CIFAR-10.
Given these statistics, the trade-offs associated with model pruning --- specifically in terms of accuracy and the slight increase in execution overhead --- are reasonable and acceptable by most, if not all, service providers, particularly when weighed against the cost savings and the enhancement in unlearning efficiency it offers. As a result, \tool incorporates RCMP into the training process on edge devices, as detailed in Figure~\ref{fig:pruning}. This approach saves memory resources for storing a single sub-model while maintaining training effectiveness, making it a valuable strategy for managing models in constrained environments such as edge devices and satellites.

\smallskip
\noindent\textbf{Remark.} It is important to acknowledge that different service providers may have various accuracy standards. Therefore, the choice of pruning rate $\delta$ should be made by each service provider based on its specific need and priority. In the following evaluations, the pruning rate $\delta = 70\%$ is selected as default, while this execution overhead of RCMP is ignored, given the minimal overhead from model pruning demonstrated in Table~\ref{tab:pruning}.

\subsection{User-Centered Data Partition}
\label{subsec:data_partition}

In machine unlearning, the training dataset is typically divided into separate and non-overlapping segments or ``shards'' to train individual sub-models in a manner that confines the impact of specific data samples to these sub-models~\cite{bourtoule2021machine, chen2022recommendation, yan2022arcane}. This approach ensures that, when there is a need to forget or ``unlearn'' certain data, only those sub-models associated with the shards containing the unlearned data need to be retrained. In this way, the efficiency of such retraining can be improved significantly. 
The efficiency gains in data partition are directly proportional to the size of the data shard: smaller shards result in lower computational costs associated with the unlearning process. 
However, this approach also means that each sub-model has access to a smaller training dataset, which can indirectly reduce the overall accuracy of the aggregated model~\cite{xu2024machine}.

Currently, the data partition methods are designed from multiple perspectives, such as class-based~\cite{yan2022arcane}, data similarity-based~\cite{chen2022recommendation, qian2022patient} and uniform-based~\cite{bourtoule2021machine} partitions.
However, traditional methods of data partitioning, which are effective in isolating training samples for machine unlearning, cannot be directly implemented at the network edge. This is because a device at the network edge typically collects data from multiple users, and the nature of this data can range from being very similar to quite distinct among these users. In the meanwhile, the unlearning request from a user only contains the data from this user in most cases. Inspired by the data partitioning method proposed in~\cite{chen2022recommendation}, we propose a user-centered data partition (UCDP) algorithm, as presented in Algorithm \ref{alg:ucdp}.

\begin{algorithm}[t]
\footnotesize
\DontPrintSemicolon
  \KwInput{shard number $\mathcal{S}$, user set $\mathcal{U}$, data set $\mathcal{D} = \{\mathcal{D}_1, \mathcal{D}_2, \dots, \mathcal{D}_{|\mathcal{U}|}\}$ 
  }
  \KwOutput{shards $\mathcal{D}' = \{\mathcal{D}'_1, \mathcal{D}'_2, \dots, \mathcal{D}'_{\mathcal{S}}\}$ }

  \If{$\mathcal{S} < |\mathcal{U}|$\label{line_check_n_user}} 
  {
    {$\Bar{\vartheta} = \frac{\sum_{k=1}^{|\mathcal{U}|} size(\mathcal{D}_k)}{|\mathcal{U}|}$}\label{line_cal_avg}
    
    {select $\mathcal{S}$ users $\{ u'_1, u'_2, \dots, u'_{\mathcal{S}} \}$ randomly from $\mathcal{U}$\label{line_select_user}}

    {$\hat{\mathcal{U}} \leftarrow \{ u'_1, u'_2, \dots, u'_{\mathcal{S}} \}$\label{line_setup_u}}

    {$\mathcal{D}' \leftarrow \{\mathcal{D}'_1 \leftarrow \{\mathcal{D}'_{u'_1}\}, \mathcal{D}'_2 \leftarrow \{\mathcal{D}'_{u'_2}\}, \dots, \mathcal{D}'_{\mathcal{S}} \leftarrow \{\mathcal{D}'_{u'_{\mathcal{S}}}\} \}$\label{line_setup_d}}

    \While{$\mathcal{D} \neq \varnothing$\label{line_while}}
    {
            
        \For{$1 \leq s \leq \mathcal{S}$}
        {          
          {select the user $ u_k $ to achieve $\min \{ \big \lfloor  \frac{size(\mathcal{D}'_s) + size(\mathcal{D}_k)}{(|\mathcal{D}'_s|+1)} - \Bar{\vartheta} \big \rfloor_+ ~|~ u_k \in \mathcal{U} \cup \neg \hat{\mathcal{U}} \}$ \label{line_min}} 
    
          {$\mathcal{D}'_s \leftarrow \mathcal{D}'_s \cup \mathcal{D}_k $}
          
          {$\hat{\mathcal{U}}  \leftarrow \hat{\mathcal{U}} \cup u_k$}
          
          {$\mathcal{D}  \leftarrow \mathcal{D} \cup \neg \mathcal{D}_k$}\label{line_setup_dataset}
        }
        
    }
   
  }
  \Else
  {
    \For{$1 \leq s \leq |\mathcal{U}|$\label{line_for_user}}
    {
        $\mathcal{D}'_s \leftarrow \mathcal{D}_s$\label{line_end}
    }
  }
\caption{User-centered Data Partition}
\label{alg:ucdp}
\end{algorithm}

Given the shard number $\mathcal{S}$, UCDP first checks the number of users $|\mathcal{U}|$ who contribute data to the model training in the current round (Line~\ref{line_check_n_user}). If $|\mathcal{U}|$ is no more than $\mathcal{S}$, UCDP partitions the training data $\mathcal{D}$ into $|\mathcal{U}|$ data shards, and each data shard only contains data from a single user (Lines~\ref{line_for_user} to~\ref{line_end}). Otherwise, UCDP allocates users' data into $\mathcal{S}$ shards similar to the knapsack problem (Lines~\ref{line_cal_avg} to~\ref{line_setup_dataset}). UCDP first calculates the average size $\Bar{\vartheta}$ of training data contributed by the users (Line~\ref{line_cal_avg}). Then, $\mathcal{S}$ users are randomly selected as set $\hat{\mathcal{U}}$ from all the users $\mathcal{U}$ who contribute training data (Lines~\ref{line_select_user} to~\ref{line_setup_u}), and their data are selected as the original datasets for $\mathcal{S}$ shards (Line~\ref{line_setup_d}). Then, UCDP allocates the remaining data into those $\mathcal{S}$ shards iteratively (Lines~\ref{line_while} to~\ref{line_setup_dataset}), achieving an approximate balance among shards. Specifically, $\mathcal{D}_k$ is assigned to shard $s$ if, after updating the data shard $s$ with $\mathcal{D}_k$, the data size per user is the closest to $\Bar{\vartheta}$ (Line~\ref{line_min}).

\begin{figure}[!tbp]
    \vspace{-1. em}
    \centering
    \subfigure[Accuracy \vs $\mathcal{S}$ on CIFAR-10]{
    \includegraphics[width=0.43\linewidth]{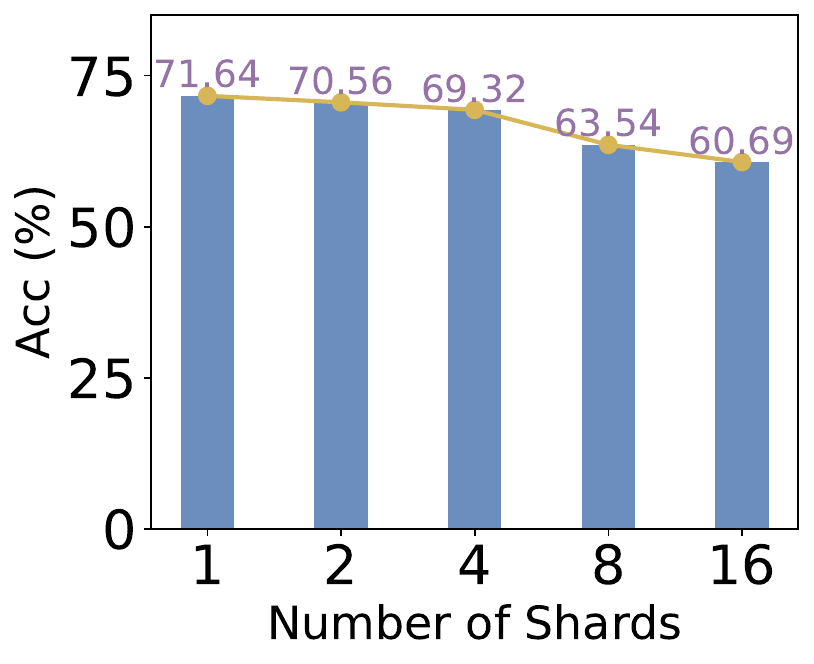} \label{fig:efficiency_set11}} 
    \subfigure[Accuracy \vs $\mathcal{S}$ on SVHN]{
    \includegraphics[width=0.43\linewidth]{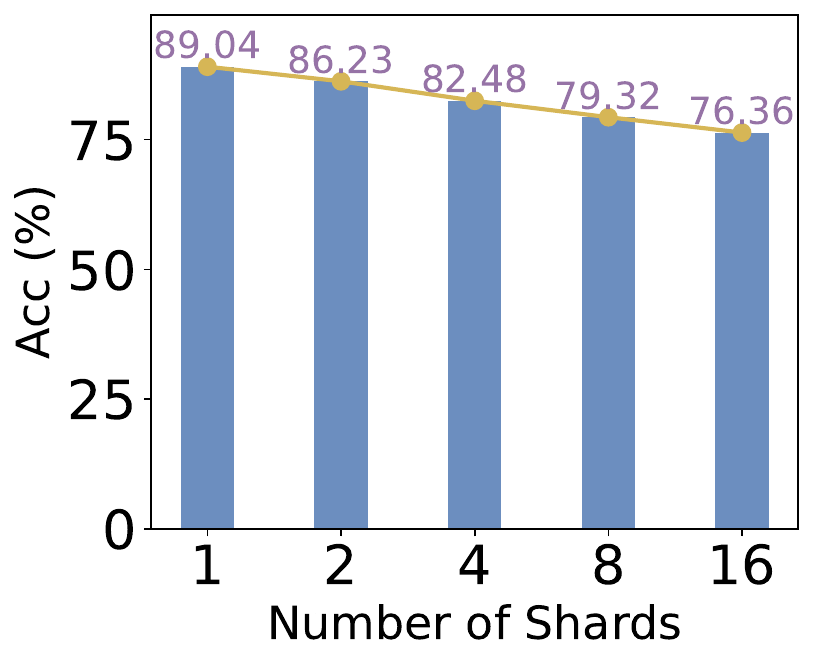} \label{fig:efficiency_set12}} \\
    \vspace{-0.5 em}
    \caption{Accuracy with various shard numbers.}
    \label{fig:motivation_shards}
    \vspace{-1. em}
\end{figure}

Typically, the model owner, \eg service providers, has prior knowledge regarding the training and inference processes. Consequently, this model owner can approximately determine and specify the number of data shards, \ie $\mathcal{S}$, based on its retraining and model accuracy requirements. To further understand the relationship between the number of shards $\mathcal{S}$ and the model accuracy, we conducted a set of experiments by training ResNet-34 on CIFAR-10 and SVHN with $\mathcal{S}=1, 2, 4, 8, 16$ under the default settings presented in~\S\ref{subsec:exp_setup}. Figure~\ref{fig:motivation_shards} shows the results. Unsurprisingly, with the increase from 1 data shard to 16 data shards, the accuracy decreases by 15.28\% (from 0.7164 to 0.6069) on CIFAR-10 and by 14.24\% (from 0.8904 to 0.7636) on SVHN. This occurs because each sub-model is trained on a smaller dataset, consequently leading to a lower overall accuracy of the aggregated results. This is also aligned with observations in existing unlearning studies~\cite{bourtoule2021machine, yan2022arcane}.

\smallskip
\noindent\textbf{Remark.} Despite the possibility of overlapping data among different users, it is essential to treat those overlapping data as distinct. This approach ensures that the privacy concerns of individual users are respected, even in cases where their data might be similar or identical to that of others.

By leveraging UCDP, \tool efficiently processes a user's unlearning request by locating and retraining only a single data shard and its corresponding sub-model. This approach substantially reduces the unlearning overhead and accelerates the entire process.
\tool also has the capability to further partition the data shard into multiple data slices, same as SISA~\cite{bourtoule2021machine}. However, this would necessitate \tool to store additional sub-models, thus increasing the burden on the resource-limited edge device. Given that the slicing methodology is not the primary focus of this research and there are no changes to the slicing algorithm introduced in SISA~\cite{bourtoule2021machine}, we omit this slicing process in this paper.

\subsection{Fibonacci-based Replacement}
\label{subsec:replacement}

The memory resources on a device at the network edge are usually very limited. Although different techniques are designed in~\S\ref{subsec:pruning} and~\S\ref{subsec:data_partition} to save memory usage for storing more sub-models, it is impractical for most, if not all, service providers to save all the sub-models over time. 

\begin{figure}[t]
    \centering
    \includegraphics[width=0.9\linewidth]{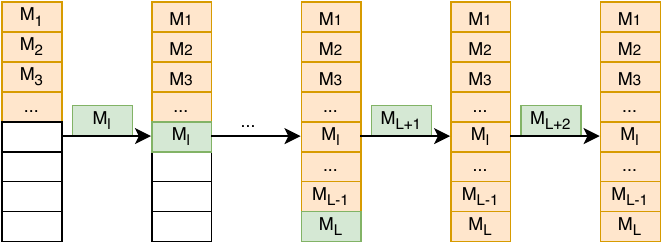}
    \caption{Memory updates without replacement.}
    \label{fig:no_replacement}\vspace{-0.5 em}
\end{figure}

Existing unlearning studies mainly focus on the efficiency and effectiveness of the unlearning performance, while ignoring the available memory limitation on the devices, as discussed in~\S\ref{subsec:background_exact_unlearning}. As a result, the replacement of sub-models stored on a resource-constrained device is not particularly mentioned in existing unlearning studies. Figure~\ref{fig:no_replacement} shows an example of the memory updates in traditional machine unlearning systems, considering the assumption that the memory can store $\mathcal{L}$ sub-models, due to the contained memory resource on a device at the network edge. Before the memory resources are fully utilized by storing sub-models, each newly trained sub-model, \ie $\mathcal{M}_l, \forall l \leq \mathcal{L}$, is directly stored in memory, as shown in Figure~\ref{fig:no_replacement}. However, once the memory is exhausted by storing $\mathcal{M}_1, \mathcal{M}_2, \dots, \mathcal{M}_l, \dots, \mathcal{M}_{\mathcal{L}}$, newly trained sub-models will no longer be stored in memory due to the lack of a sub-model replacement strategy. In this case, the sub-models trained after $\mathcal{M}_{\mathcal{L}}$ cannot be stored in the memory. Once the unlearning request of data $\mathcal{D}_r$ with $(\mathcal{D}_1 \cup \mathcal{D}_2 \cup \cdots \cup \mathcal{D}_{\mathcal{L}}) \cap \mathcal{D}_r = \varnothing$ arrives, the impacted sub-model has to be retrained from its latest version in $\{\mathcal{M}_1, \mathcal{M}_2, \dots, \mathcal{M}_\mathcal{L}\}$. Therefore, this strategy will dramatically lower the retraining efficiency over time, weakening the benefits from exact unlearning.

\begin{figure}[t]
    \centering
    \includegraphics[width=0.9\linewidth]{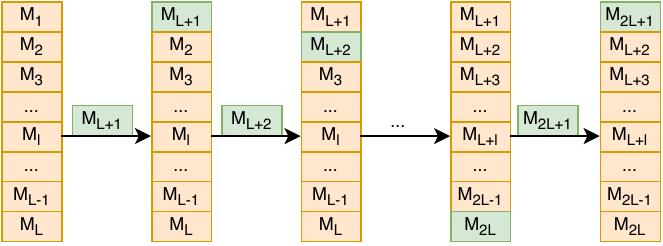}
    \caption{Memory updates with FIFO replacement.}
    \label{fig:fifo}
    \vspace{-1. em}
\end{figure}

\begin{figure*}
\begin{minipage}[t]{.67\linewidth}
\includegraphics[width=\linewidth]{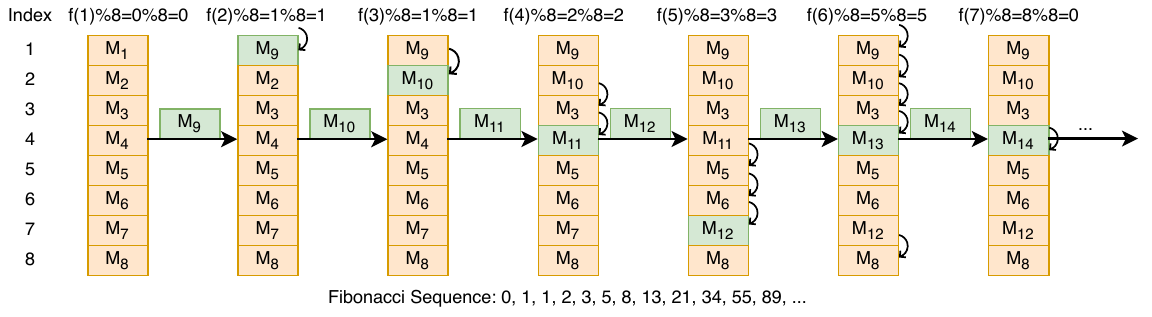}
\captionof{figure}{An example of memory updates with Fibonacci-based replacement.}
\label{fig:fibonacci}
\end{minipage}
\begin{minipage}[t]{0.33\linewidth}
\includegraphics[width=\linewidth]{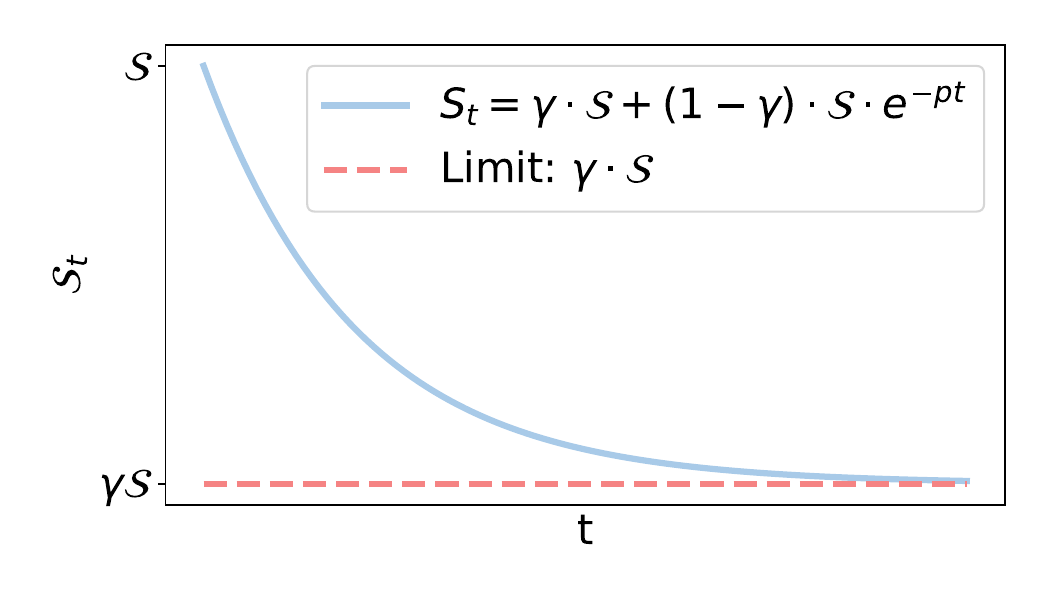}
\caption{Shard control function.}
\label{fig:sc_function}
\end{minipage}
\vspace{-.5 em}
\end{figure*}

Cache or memory replacement strategies have been widely investigated in the last decades~\cite{podlipnig2003survey}. One of the most representative replacement strategies is First-In-First-Out (FIFO). Figure~\ref{fig:fifo} depicts the sub-model storing and replacing process with the FIFO replacement strategy. Before the memory resources are fully used up by storing sub-models, the sub-model storing process is exactly the same as that in Figure~\ref{fig:no_replacement}. After the memory is exhausted by storing $\mathcal{M}_1, \mathcal{M}_2, \dots, \mathcal{M}_l, \dots, \mathcal{M}_{\mathcal{L}}$, the FIFO replacement strategy replaces the sub-model $\mathcal{M}_1$ in the memory by the newly trained sub-model, \eg $\mathcal{M}_{\mathcal{L}+1}$. With new sub-models trained, sub-models from $\mathcal{M}_1$ to $\mathcal{M}_{L}$ are replaced by from $\mathcal{M}_{\mathcal{L}+1}$ to $\mathcal{M}_{2L}$, accordingly. After that, the newly trained sub-models start to replace from $\mathcal{M}_{\mathcal{L}+1}$ to $\mathcal{M}_{2L}$ in order. The advantage of this FIFO-based replacement is always keeping the newest sub-models in the memory. In this way, when a request to unlearn recently trained data arrives, the sub-model is more likely to be retrained from a recently stored sub-model. This significantly reduces the time needed for retraining, compared to the strategy in Figure~\ref{fig:no_replacement}. However, such a strategy has its significant shortcomings. In the case that there is a request to forget data learned a considerable time ago, the corresponding sub-model must undergo retraining from the start point of the entire training life. This is because the sub-model that originally learned the data to be forgotten is replaced by a newly trained sub-model. Consequently, FIFO-based algorithms result in significant retraining overheads.

Theoretically, the advantage of storage replacement strategies in improving unlearning speed lies in the temporal sparsity of stored sub-models, crucial due to the unpredictability of unlearning requests. Static and FIFO strategies often store sub-models trained within a narrow time window, resulting in poor temporal sparsity and suboptimal performance. Thus, a ``jumping'' style for replacing existing sub-models stored in memory is needed.

\begin{algorithm}[t]
\footnotesize
\DontPrintSemicolon
  \KwInput{normalized memory resource $\mathcal{N}_{mem}$ and the set of newly trained sub-models $\wp\mathcal{M}$
  }

  $\Delta \mathcal{N}_{mem}$ = $\mathcal{N}_{mem}$

  $I_{replace}$ = 1

  $I_{FiboR}$ = 0

  \For{$\mathcal{M}_x \in \wp\mathcal{M}$}
  {
    \If{$\Delta \mathcal{N}_{mem} \geq size(\mathcal{M}_x)$}
    {
        store $\mathcal{M}_x$ into the memory
        
        $\Delta \mathcal{N}_{mem}$ = $\Delta \mathcal{N}_{mem}$ - $size(\mathcal{M}_x)$
    }
    \Else
    {
        $I_{replace}$ = [ $I_{replace}$ + $f(I_{FiboR})$ \%  $\mathcal{N}_{mem}$ ] \%  $\mathcal{N}_{mem}$

        replace the sub-model at $I_{replace}$ by $\mathcal{M}_x$

        $I_{FiboR}$++
    }
  }
\caption{Fibonacci-based Replacement}
\label{alg:fbior}
\end{algorithm}

The Fibonacci sequence~\cite{fibonacci} has long been utilized across a variety of applications such as data queries~\cite{brodal2012two}, synchronization control~\cite{kautz1965fibonacci} and network optimization~\cite{fredman1987fibonacci}, leveraging its benefits of temporal sparsity. Implementing Fibonacci in the context of unlearning allows for closer alignment to the start point of retraining in storage across time and shard dimensions for arbitrary unlearning requests. Inspired by the Fibonacci Sequence, we designed a Fibonacci-based replacement (FiboR) algorithm to utilize the memory resources for effective and efficient exact unlearning.

The pseudo-code is presented in Algorithm~\ref{alg:fbior}. After a set of sub-models $\wp\mathcal{M}$ are trained, FiboR starts to perform the model placement and replacement operations. Without loss of generality, FiboR normalizes the memory resources by the number of sub-models that can be stored, and initialize the remaining memory resource $\Delta \mathcal{N}_{mem}$ = $\mathcal{N}_{mem}$. Here, FiboR uses two indexes, \ie $I_{replace}$ and $I_{FiboR}$ for indicating the memory place for storing the newly trained sub-model and indicating the current index in the Fibonacci Sequence, respectively. If the remaining resource is enough, FiboR just directly stores the newly trained model $\mathcal{M}_x \in \wp\mathcal{M}$  in the memory (Lines 5-7). Otherwise, FiboR uses the Fibonacci function $f(\cdot)$ to define the memory address for replacement by  $[I_{replace} + f(I_{FiboR}) \%  \mathcal{N}_{mem}]  \%  \mathcal{N}_{mem}$ (Lines 9-11). FiboR iteratively executes this process for all new sub-models.

Figure~\ref{fig:fibonacci} shows an example of memory updates with the Fibonacci-based replacement strategy. Real-world scenarios are much more complicated compared to the example illustrated in Figure~\ref{fig:fibonacci}. However, the purpose of this example is to clearly demonstrate the functioning of FiboR. In this example, we assume that the available memory resources can only store 8 sub-models. Before the memory resources are exhausted by storing sub-models, the sub-model storing process is exactly the same as that in Figure~\ref{fig:no_replacement} and Figure~\ref{fig:fifo}. After the memory is exhausted by storing $\mathcal{M}_1, \mathcal{M}_2, \dots, \mathcal{M}_8$, FiboR starts to replace the stored sub-models according to the Fibonacci Sequence. Once $\mathcal{M}_9$ is trained, FiboR calculates the f-index by $f(0)\%8 = 0$, where $f(0)$ is the number of index 0 in the Fibonacci Sequence. In this way, the f-index does not move and stays at index 1 of the memory. Thus, $\mathcal{M}_1$ is selected to be replaced by $\mathcal{M}_9$. Similarly, once $\mathcal{M}_{10}$ is trained, FiboR calculates the f-index by $f(1)\%8 = 1$ and the replacement index moves from index 1 to index 2. Consequently, $\mathcal{M}_2$ is selected to be replaced by $\mathcal{M}_{10}$. Following such rules, $\mathcal{M}_{11}, \mathcal{M}_{12}, \mathcal{M}_{13}$ and $\mathcal{M}_{14}$ replace $\mathcal{M}_4, \mathcal{M}_7, \mathcal{M}_{11}$ and $\mathcal{M}_{13}$ in the memory, accordingly. Finally, sub-models $\mathcal{M}_3, \mathcal{M}_5, \mathcal{M}_6, \mathcal{M}_8, \mathcal{M}_9, \mathcal{M}_{10}, \mathcal{M}_{12}$ and $\mathcal{M}_{14}$ are stored after updates. In this way, for forgetting data participating in the training of any sub-models in $\mathcal{M}_1, \dots, \mathcal{M}_{14}$, the sub-model can be easily identified, close to the retraining point, reducing the corresponding retraining overhead.

\noindent\textbf{Remark}. While random replacement is also a ``jump" strategy, its temporal sparsity is inherently unstable. FiboR, however, leverages the Fibonacci sequence, replacing storage locations at different frequencies in an observable cyclic pattern. Some positions in the queue are surely replaced less frequently, allowing them to retain older sub-models longer. For example, with a storage capacity of 10 sub-models, the replacement pattern repeats every 60 rounds. In each cycle, positions 5, 7, and 9 are replaced 4 times, less frequently than the 6 times expected in random replacement over 60 rounds. Furthermore, this cyclic nature ensures that after a certain number of iterations, most, if not all, sub-models are replaced, allowing for a sufficient mix of new models. 
FiboR's ability to maintain a diverse mix of older and newer sub-models significantly enhances overall performance, outperforming random, static, and FIFO strategies. To investigate the advantage of FiboR compared to random replacement, we conducted the evaluation of various replacements with \tool under the default experiment setup in~\cref{sec:evaluation}. The results show that the retrained sample number achieved by \tool with FiboR is 143,226 while it is 154,193 with the random replacement. 

\subsection{Shard Controller}
\label{subsec:shard_controller}

\tool improves the unlearning efficiency by conducting Fibonacci-based replacement With the implementation of FiboR (\S\ref{subsec:replacement}). However, due to \textit{limited memory resources}, frequent replacement operations are inevitable over time. This leads the system into the dilemma of FIFO-based replacement strategies as shown in Figure~\ref{fig:fifo}. Combining the observation in \S\ref{subsec:data_partition}, we found that reducing the value of the shard number, \ie $\mathcal{S}$, will lead to \textit{the improvement in the model accuracy}, while simultaneously reducing the number of replacement operations in memory. Hence, updating the value of $\mathcal{S}$ dynamically in a decreasing manner might offer a solution to alleviate the dilemma mentioned above. This idea aligns with the moving average~\cite{moving_average} in statistics. 
In addition, considering the issue caused by lazy replacement strategies, for example shown in Figure~\ref{fig:no_replacement}, the value of $\mathcal{S}$ should not be too small to fall into the low update frequency.

Thus, we design a shard controller (SC) based on the exponential weighted moving average (EWMA)~\cite{hunter1986exponentially} to decrease the value of $\mathcal{S}$ over time in an exponential manner. Specifically, we design the dynamic shard function \eqref{eq:sc_function} as follows:
\vspace{-0.5em}
\begin{equation}
\label{eq:sc_function}
    \mathcal{S}_t = \gamma \cdot \mathcal{S} + (1 - \gamma) \cdot  \mathcal{S} \cdot e^{-pt}, 
    \vspace{-0.5em}
\end{equation}
where $\mathcal{S}$ is the original shard number and $\mathcal{S}_t$ is the value of the shard number in the $t^{th}$ training round.

In \eqref{eq:sc_function}, the parameter $\gamma$, ranging between [0, 1], sets the minimum threshold for the shard number, and $p$ dictates the extent of reduction. Specifically, when $\gamma$ is set to 1, the value of $\mathcal{S}_t$ equals to $\mathcal{S}$ consistently over time. The function's behavior is depicted in Figure~\ref{fig:sc_function}. According to the formulation of \eqref{eq:sc_function}, the value of $\mathcal{S}_t$ is ranged within the interval $[\gamma, 1]\mathcal{S}$. This mechanism enables a gradual reduction of $\mathcal{S}_t$ from the original value $\mathcal{S}$ down to $\gamma \mathcal{S}$, modulated by the parameter $p$ to achieve our design goal.

\begin{table}[ht]
\caption{SC Performance Comparison}
\label{tab:evaluation_sc}
\resizebox{\columnwidth}{!}{%
\begin{tabular}{lccccc}
\toprule
\textbf{Shard} & \textbf{1} & \textbf{2} & \textbf{4} & \textbf{8} & \textbf{16}\\
\midrule
Accuracy & & & & & \\
\midrule
\tool & 	0.7164 &	0.7055 &	0.6931 &	0.6254 &	0.6068\\
\midrule
\tool-No-SC & 0.7164 &	0.6683 &	0.6456 &	0.5809 & 	0.5515 \\
\midrule
\multicolumn{3}{l}{Retrained Sample Number (RSN)} & & & \\
\midrule
\tool & 	586,482 &	288,546 & 	147,698 &	76,568 &	67,732\\
\midrule
\tool-No-SC & 586,482 & 	328,311 & 	163,855 & 	82,797 &	68,324 \\
\bottomrule
\end{tabular}
}
\vspace{-1 em}
\end{table}

\textbf{Remark}. SC gradually reduces the number of shards, allowing sub-models stored in limited memory to retain more learning data. This approach increases the likelihood of retraining on existing sub-models within \tool, thereby enhancing unlearning speed. Although this may seem counterintuitive to UCDP, SC actually improves unlearning speed under resource constraints. In addition, a sub-model learns more precise information from a larger dataset than from aggregating multiple sub-models trained on smaller sub-datasets. By gradually reducing the number of shards, SC contributes to greater accuracy. Here, we conducted experiments comparing \tool and \tool-No-SC, and the results in Table~\ref{tab:evaluation_sc} prove the motivation of SC.

\subsection{Resource-Constrained Adaptive Exact Unlearning System}
\label{subsec:cause}

Through integrating RCMP, UCDP, FiboR, and SC, \tool can utilize the limited resources more efficiently to store sub-models, in order to realize exact unlearning on the resource-constrained device. The pseudo-code of \tool is presented in Algorithm \ref{alg:cause}.

In the beginning, \tool first calculates the number of data shards to be divided, \ie $\mathcal{S}_t$, in this training round based on \eqref{eq:sc_function} of SC (Line 2). Then, \tool executes UCDP to partition training data into $\mathcal{S}_t$ shards (Line 3). For each data shard, \tool runs model training and RCMP to obtain the corresponding sub-model with pruning rate $\delta$ (Line 4). Once the training process is complete, \tool stores those newly trained sub-models in the memory via the execution of FiboR (Line 5).
Once a set of unlearning requests $\zeta$ arrive, \tool processes each request $\varsigma_i \in \zeta$ iteratively (Lines 8-12). For each request, \tool identifies the sub-model $\mathcal{M}_x$ most closely to the unlearned data in $\varsigma_i$ before the targeted data is learned. In addition, \tool removes the unlearned data from the training dataset and retrains the sub-model $\mathcal{M}'_x$ from $\mathcal{M}_x$. Once the retraining is complete, \tool needs to store the newly trained sub-model $\mathcal{M}'_x$ in the memory. Here, \tool can directly replace $\mathcal{M}_x$ by $\mathcal{M}'_x$ without any impact on other sub-models. However, if the number of unlearning requests is high, this operation could introduce frequent replacement issues as discussed in ~\S\ref{subsec:replacement}. Thus, \tool adopts the same strategy, \ie FiboR, to store the newly trained sub-models to address it.

\noindent\textbf{Aggregation}. Same as the aggregation strategy used in existing exact unlearning systems~\cite{bourtoule2021machine, yan2022arcane}, \tool employs a label-based majority vote mechanism to aggregate results from its sub-models. This approach is chosen to optimize the combined predictive performance of the sub-models without involving the training data. This method ensures that \tool efficiently synthesizes the individual insights of each sub-model, leveraging their collective intelligence to enhance accuracy, while also adhering to data privacy in machine unlearning scenarios.

\begin{algorithm}[t]
\footnotesize
\DontPrintSemicolon
  \KwInput{shard number $\mathcal{S}$, time set $\mathcal{T}$, user set $\mathcal{U}$, data set $\mathcal{D}$, normalized memory resource $\mathcal{N}_{mem}$, pruning rate $\delta$}

  \For{$1 \leq t \leq \mathcal{T}$}
  {
    calculate $\mathcal{S}_t$ in SC based on \eqref{eq:sc_function} 

    partition $\mathcal{D}$ into $\mathcal{S}_t$ shards based on Algorithm \ref{alg:ucdp} 

    train the $\mathcal{S}_t$ sub-models and execute pruning with $\delta$ \label{stp:train}

    store newly trained $\mathcal{S}_t$ sub-models in the memory based on Algorithm \ref{alg:fbior}

    \If{unlearning requests $\zeta$ arrive} {
      \For{$\varsigma_i \in \zeta$}
      {
        identify the sub-model $\mathcal{M}_x$ most closely to the unlearned data $\mathcal{D}_r$ in $\varsigma_i$ before $\mathcal{D}_r$ is learned in~\cref{stp:train}
        
        $\mathcal{D}_x \leftarrow \mathcal{D}_r \cup \neg \mathcal{D}_x$ 

        retrain $\mathcal{M}'_x$ from the updated data $\mathcal{D}_x$

        delete all sub-models containing any learning information in the request, \ie $\exists d \in \mathcal{D}_r$

        store the newly trained $\mathcal{M}_x$ in the memory based on Algorithm \ref{alg:fbior}
      }
    }
  }

\caption{Resource-constrained Adaptive Exact Unlearning System (\tool)}
\label{alg:cause}
\end{algorithm}

\noindent\textbf{Application Scenarios}. According to a report by IBM, the edge AI market was valued at US\$14.787 billion in 2022~\cite{edgeai}, encompassing online and incremental learning applications such as wearable health monitoring, real-time traffic updates for autonomous vehicles, and smart appliances. 
Edge devices collect sensitive data that ranges from personal information to data on critical issues, such as conflicts and wars, highlighting the growing need for edge unlearning. CAUSE can be implemented across various scenarios to uphold the right to be forgotten directly at the network edge. For IoT-enabled applications, such as health monitoring and traffic updates, a primary limitation of edge unlearning is the low unlearning speed, constrained by limited memory on IoT devices. CAUSE addresses this limitation by significantly enhancing unlearning speed, with improvements of up to 80.85\% in evaluations (\S\ref{sec:evaluation}), thereby resolving speed challenges in these application scenarios.
In edge AI applications deployed on energy-harvesting devices like AI-powered satellites, resource-limited devices face an additional challenge: the high energy demands of exact unlearning processes, further constrained by limited battery capacity and intermittent energy harvesting. CAUSE demonstrates a significant advantage in energy efficiency, reducing energy consumption by 66.21\% to 83.46\% in evaluations. By effectively overcoming these two major limitations—unlearning speed and energy consumption—CAUSE proves suitable for implementation in various exact unlearning scenarios on resource-constrained devices, underscoring its significance and feasibility for both industry and academia.

\section{Evaluation}
\label{sec:evaluation}

In this section, we conduct intensive experiments to demonstrate and analyze \tool's performance, compared to various representative exact unlearning mechanisms. To ensure the validation and reproducibility of our experimental results, both synthesized datasets and the source code are made available at \url{https://github.com/XLab-hub/CAUSE}.

\subsection{Experiment Setup}
\label{subsec:exp_setup}

\noindent \textbf{Benchmarks.}
To evaluate the performance of \tool, 3 representative machine unlearning mechanisms are involved in the performance comparison.
\begin{itemize}[leftmargin=*, noitemsep]
\item \textbf{SISA~\cite{bourtoule2021machine}.} SISA is an exact unlearning system consisting of four phases: Sharded, Isolated, Sliced, and Aggregated training phases. It uniformly partitions data into discrete shards and isolates sub-model learning within these shards. SISA simplifies retraining and reduces computational overhead during unlearning processes. 
\item \textbf{ARCANE~\cite{yan2022arcane}.} ARCANE is an exact unlearning system that operates based on class classifiers. It divides datasets into sub-datasets according to class labels, trains sub-models independently, and employs decision aggregation. In this way, when an unlearning request arises, ARCANE analyzes the class of unlearned data to identify the sub-models requiring retraining.
\item \textbf{OMP~\cite{liu2024model}.} OMP is a machine unlearning mechanism inspired by the one-shot magnitude pruning approach~\cite{liu2024model}. OMP follows a similar framework to SISA for partitioning, training, and aggregating. Different from SISA, OMP integrates the one-shot magnitude pruning process for each trained sub-model, effectively reducing memory usage to save more sub-models.
\end{itemize}

To ensure a fair comparison, we adjust the benchmark systems to align with practical scenarios if necessary. Specifically, following ARCANE's design, we grouped data classes and assigned them to each shard based on the total number of shards in ARCANE. Additionally, we introduce two versions of OMP: the original 95\% sparsity~\cite{liu2024model} (OMP-95), and a 70\% sparsity, consistent with \tool (OMP-70). This ensures the fairness and generality of our evaluations under the same experimental settings. Detailed system settings including training and pruning information can be found in Table~\ref{tab:model_training_and_pruning_details} in Appendix~\ref{apx:settings}.

\subsubsection{Datasets and models}
\label{subsubsec:dm}

As discussed in~\S\ref{sec:introduction}, the dynamic nature of the network edge is a key attribute. At the network edge, data from users—such as images or videos captured by sensors, cameras, or even satellites, is continually updated, and users may request data unlearning at any time. Users can specify requests to delete data from certain periods or specific time slots, adding complexity to the already resource-constrained edge devices.
To conduct realistic scenarios for our experiments, we constructed \textit{synthetic imbalanced datasets} based on the widely-used datasets, including CIFAR-10~\cite{krizhevsky2009learning}, SVHN~\cite{netzer2011reading}, and CIFAR-100~\cite{krizhevsky2009learning}, by randomly shuffling data categories and quantities to model heterogeneous user data, where each user's data are fully different 
in terms of data instances, labels, and sizes. Those synthetic datasets incorporate dynamic user learning and unlearning requests over time, with each training round representing a distinct time slot. Each user can request the unlearning of a randomly generated subset of their data, with the probability of raising the unlearning request based on $\rho_{u}$. When the device receives multiple unlearning requests, it processes them on a first-come-first-served policy. 

To evaluate the performance of all the exact unlearning systems comprehensively, our experiments are conducted using image classification tasks employing ResNet-34~\cite{he2016deep}, VGG-16~\cite{simonyan2015very} and MobileNetV2~\cite{sandler2018mobilenetv2}, along with the synthetic datasets.
The basic information of datasets used to conduct synthetic datasets are as follows:
\begin{itemize}[leftmargin=*, noitemsep]
    \item CIFAR-10~\cite{krizhevsky2009learning}. CIFAR-10 contains 60,000 32$\times$32$\times$3 color images in 10 classes, with 6,000 images in each class. There are 50,000 training images and 10,000 test images in total. 
    \item SVHN~\cite{netzer2011reading}. SVHN contains 604,833 32$\times$32$\times$3 color digital images in 10 classes. There are 630,420 training images and 26,032 test images in total. 
    \item CIFAR-100~\cite{krizhevsky2009learning}. CIFAR-100 contains 60,000 32$\times$32$\times$3 images in 100 classes, with 600 images in each class. There are 500 training images and 100 test images in each class. 
\end{itemize}
The configurations of model training can be found in Table~\ref{tab:dataset_and_backbone_models}.
\begin{table}[ht]
\caption{Dataset and Backbone Models}
\centering
\label{tab:dataset_and_backbone_models}
\resizebox{0.9\columnwidth}{!}{%
\begin{tabular}{l|c|c|c|c}
\toprule
\multirow{2.5}{*}{Settings} & CIFAR-10 & SVHN & CIFAR-100 & CIFAR-10 \\
\cmidrule(r){2-5} 
& ResNet-34 & ResNet-34 & VGG-16 & MobileNetV2 \\
\midrule
Batch Size & 256 & 256 & 512 & 128 \\
\bottomrule
\end{tabular}
} 
\end{table}

\subsubsection{Implementation}

Our experiments are conducted on a system running Ubuntu 22.04.3 LTS 64-bit. We utilize TensorFlow v2.15.0 paired with CUDA 12.2 and Python 3.9 for machine learning tasks, including model pruning and unlearning. For the edge device, an NVIDIA Jetson Orin Nano is deployed for experiments. By default, each model undergoes training across 10 training rounds, \ie $\mathcal{T}=10$, with 80 epochs in each training round. The number of shards is fixed at $\mathcal{L}=4$, and the default memory for training is 2GB, \ie $\mathcal{C}_m=2$. There are 100 users with non-i.i.d data, and the number of unlearning requests is determined by the probability of an edge user's unlearning request $\rho_{u}$ with the default value $0.1$. In \tool, the parameters in SC are set as $p=\gamma=0.5$. In our scalability tests (\S\ref{subsec:exp_scalability}), and specific evaluations on shard numbers (\S\ref{subsec:exp_shard}), and data partitioning strategies (\S\ref{subsec:exp_dp}), we vary the number of shards and the unlearning request probability. These modifications help us explore a variety of edge machine unlearning scenarios, assessing how these factors influence overall performance.

\subsubsection{Performance metrics} \textit{Unlearning speed} is a critical metric for evaluating the performance of exact unlearning. In our experiments, we evaluate this by measuring the number of retrained samples (\textbf{RSN}), leveraging the linear relationship between the processing time and RSN~\cite{bourtoule2021machine,yan2022arcane} and results in~\S\ref{sec:design_motivation}. This ensures that our evaluation remains consistent across devices with varying specifications. Additionally, we evaluate the \textit{accuracy} (\textbf{Acc}) of exact unlearning systems, including \tool, to identify any significant accuracy degradation among the systems. We also measure the \textit{energy consumption} to evaluate the feasibility of exact unlearning systems on resource-constrained devices. 

\subsection{Accuracy Evaluation}

\begin{figure}[t]
\begin{minipage}{\linewidth}
\centering
\subfigure[ResNet-34 on CIFAR-10]{    \includegraphics[width=0.42\linewidth]{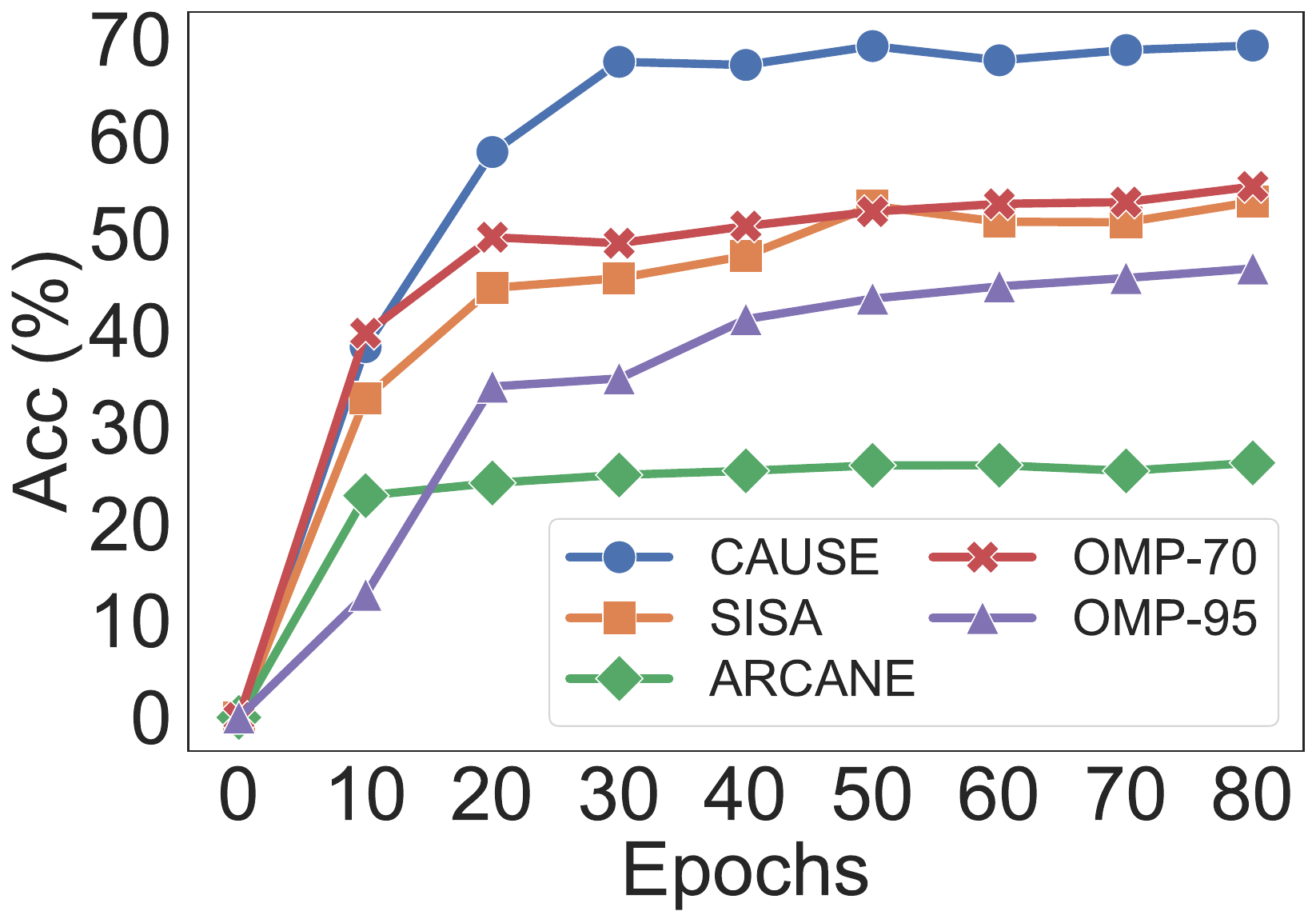}\label{fig:exp_acc_res_cifar10}}
\subfigure[ResNet-34 on SVHN]{
\includegraphics[width=0.42\linewidth]{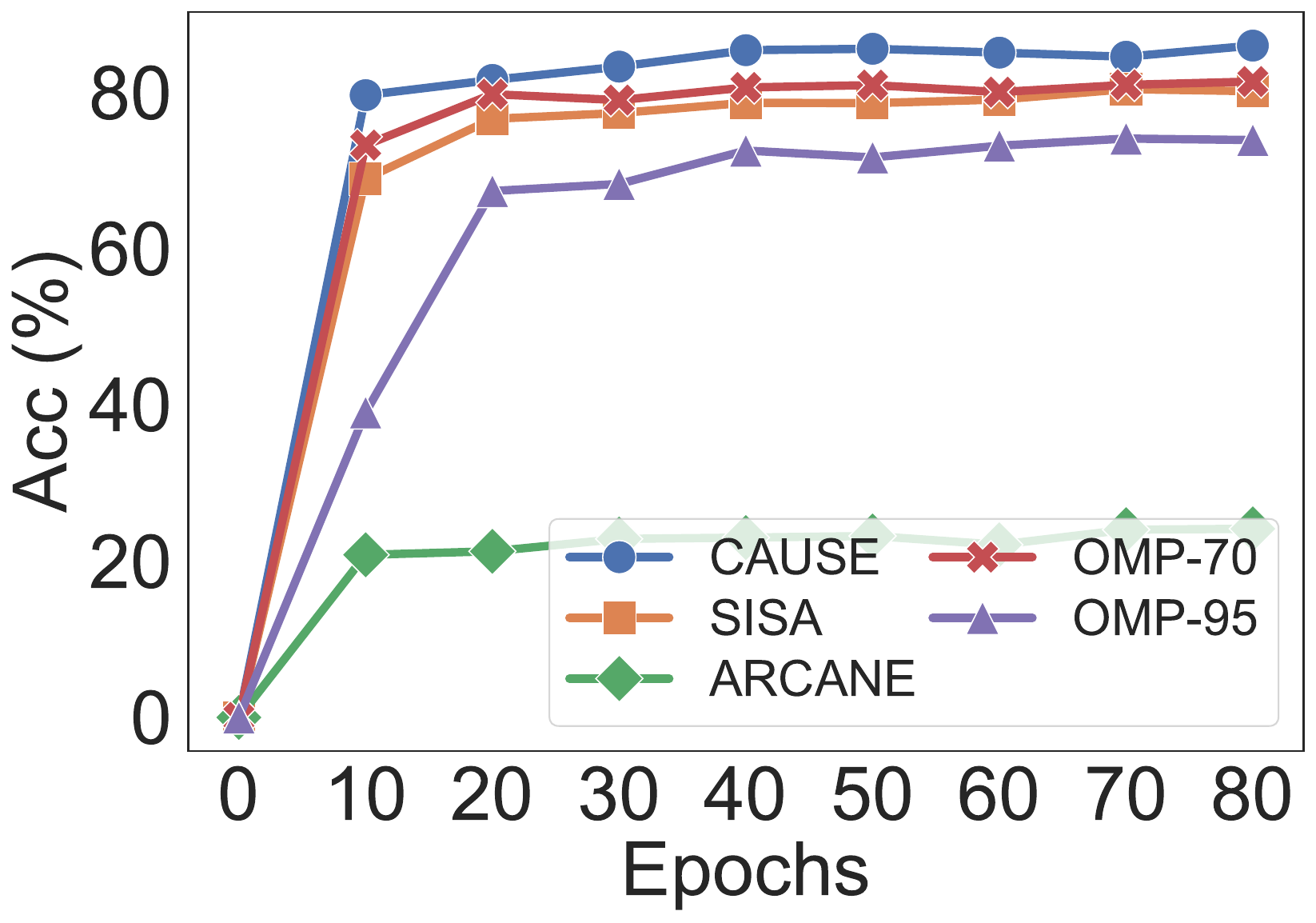}\label{fig:exp_acc_res_svhn}}
\subfigure[VGG-16 on CIFAR-100]{
\includegraphics[width=0.42\textwidth]{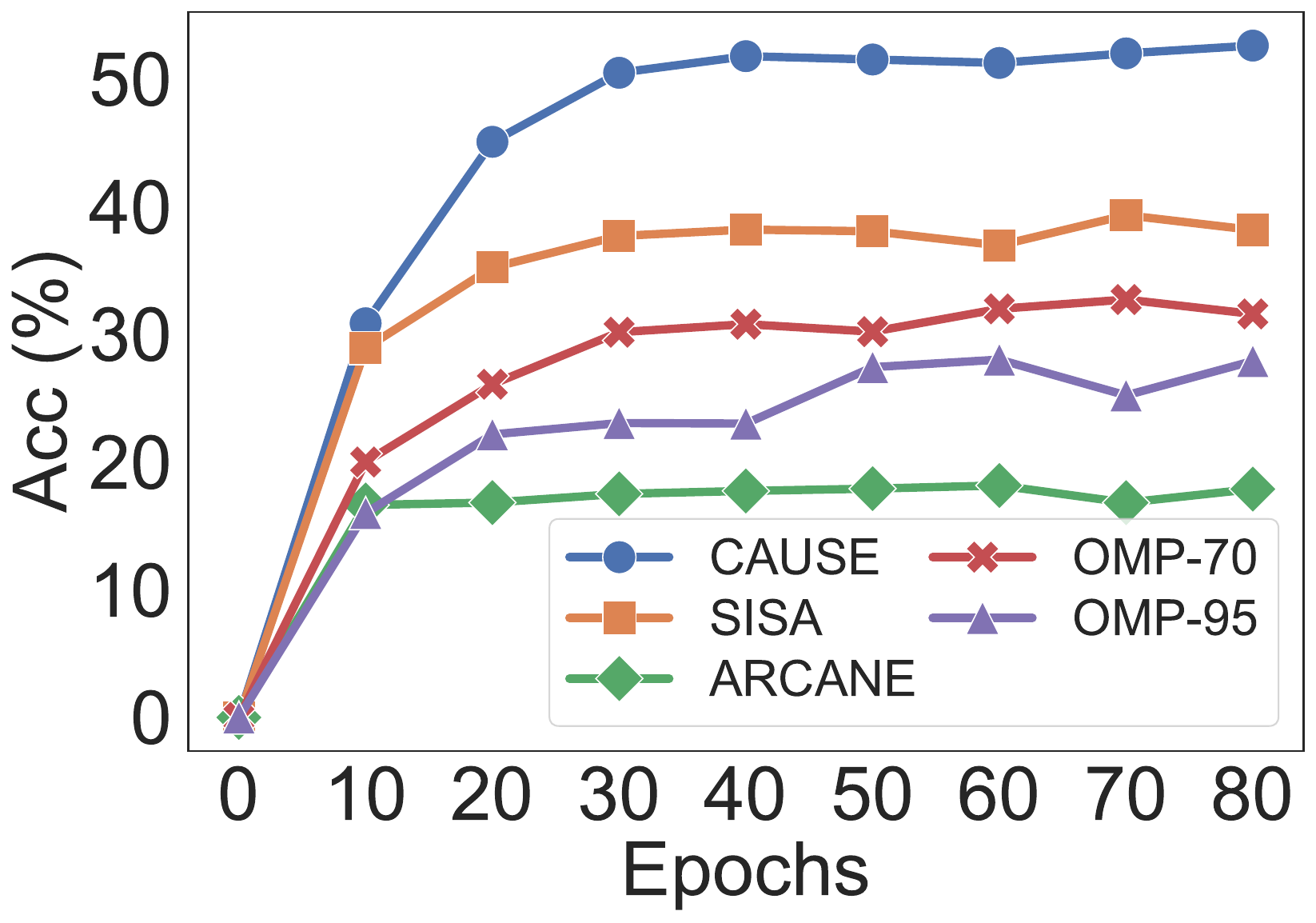}\label{fig:exp_acc_vgg_cifar100}}
\subfigure[MobileNetV2 on CIFAR-10]{
\includegraphics[width=0.42\textwidth]{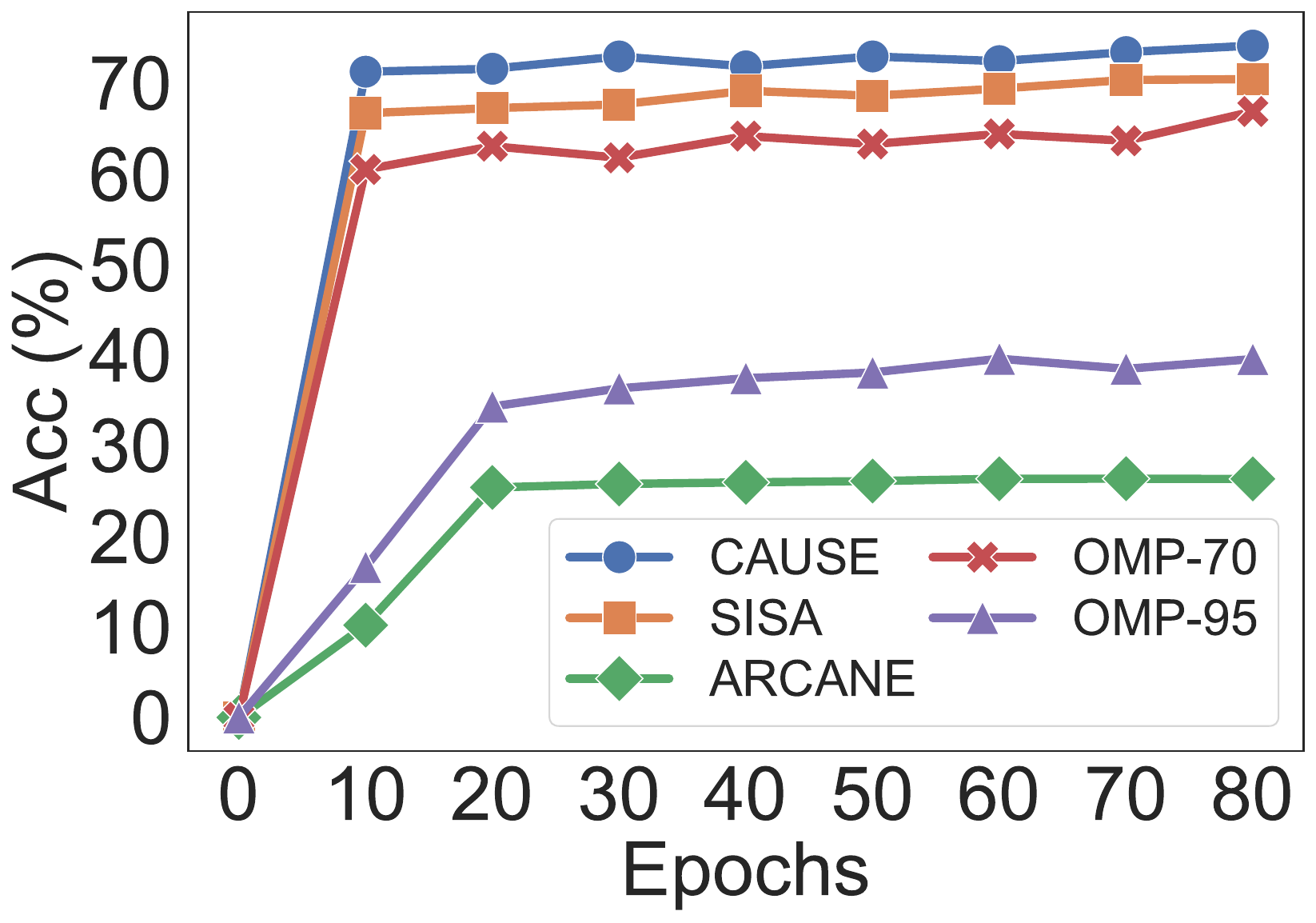}\label{fig:exp_acc_v2_cifar10}}
\vspace{-0.5 em}
\caption{Accuracy comparison over 80 epochs.}
\label{fig:exp_acc_epoch}
\vspace{-1 em}
\end{minipage}
\end{figure}

\begin{figure}[ht]
\centering
\vspace{-1em}
\setlength{\textfloatsep}{5pt}
\begin{minipage}[c]{\linewidth}
    \centering
    
    \subfigure[ResNet-34 on CIFAR-10]{
    \includegraphics[width=0.3\linewidth]{results/exp_top1_acc_vs_epoch_cifar10.pdf}}
    \subfigure[VGG-16 on CIFAR-10]{
    \includegraphics[width=0.3\linewidth]{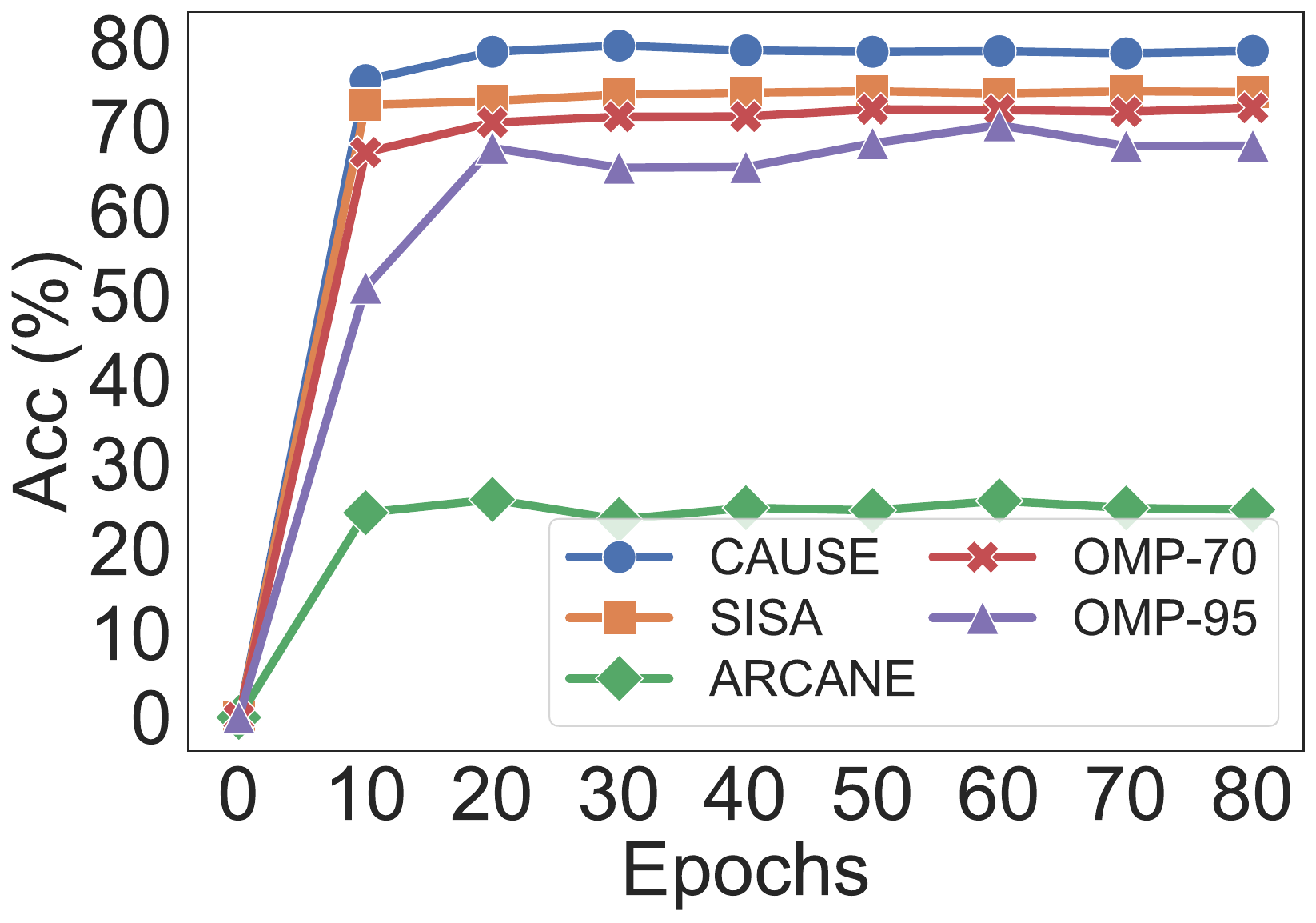}}
    \subfigure[MobileNetv2 on CIFAR-10]{
    \includegraphics[width=0.3\linewidth]{results/exp_top1_acc_vs_epoch_mobilenet.pdf}}
   
    \subfigure[ResNet-34 on SVHN]{
    \includegraphics[width=0.3\linewidth]{results/exp_top1_acc_vs_epoch_svhn.pdf}}
    \subfigure[VGG-16 on SVHN]{
    \includegraphics[width=0.3\linewidth]{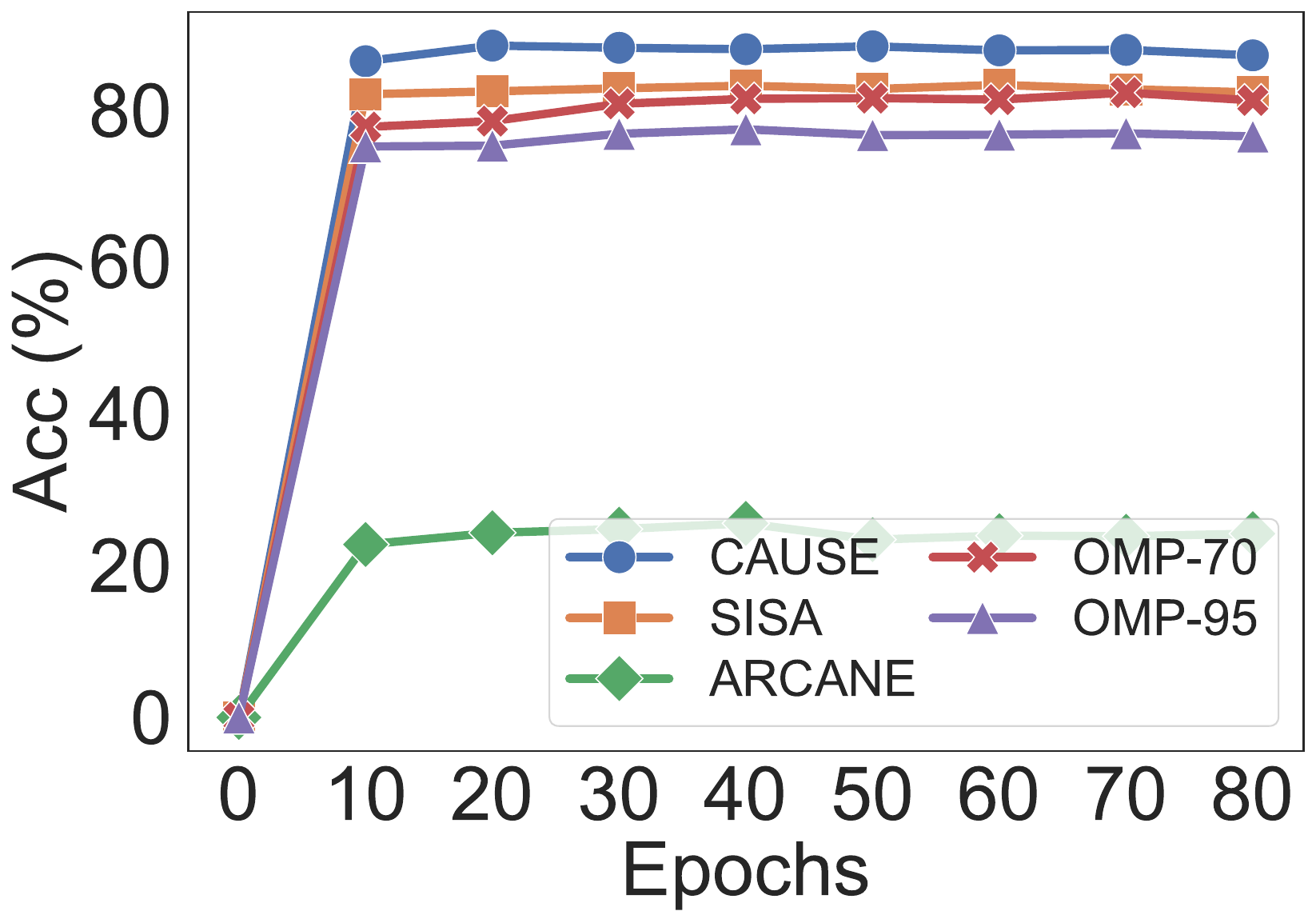}}
    \subfigure[MobileNetv2 on SVHN]{
    \includegraphics[width=0.3\linewidth]{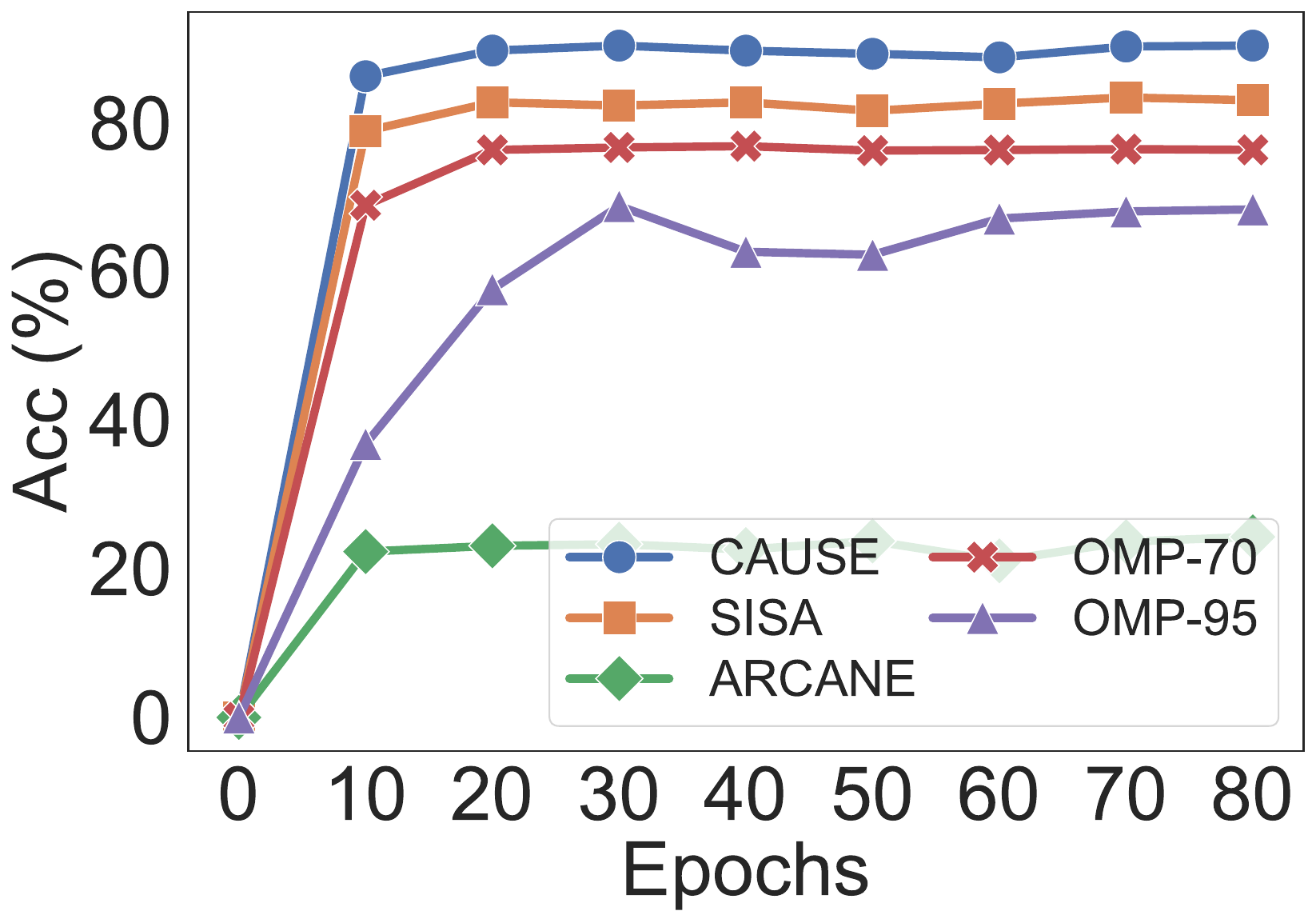}}

    \subfigure[ResNet-34 on CIFAR-100]{
    \includegraphics[width=0.3\linewidth]{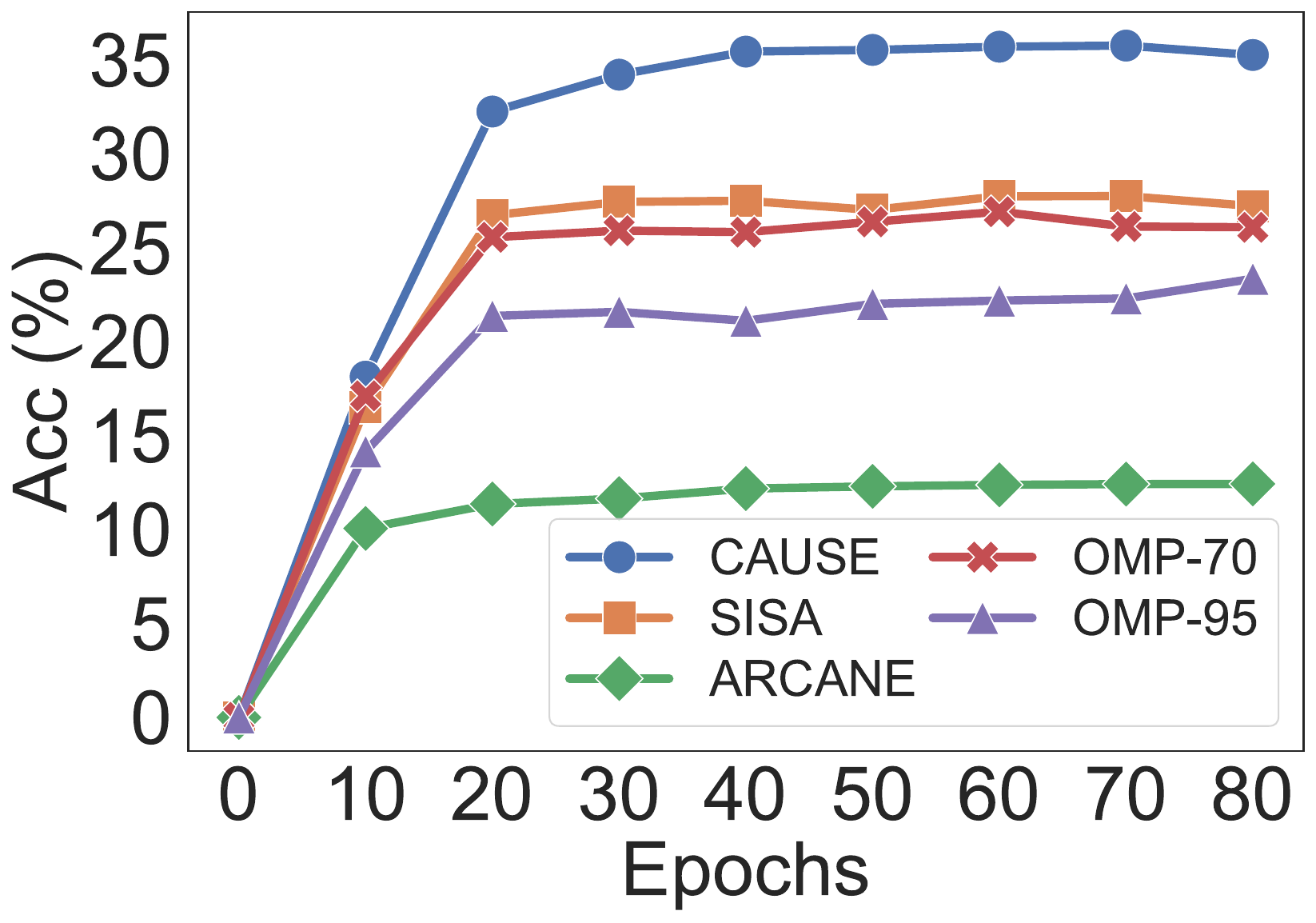}}
    \subfigure[VGG-16 on CIFAR-100]{
    \includegraphics[width=0.3\linewidth]{results/exp_top1_acc_vs_epoch_cifar100.pdf}}
    \subfigure[MobileNetv2 on CIFAR-100]{
    \includegraphics[width=0.3\linewidth]{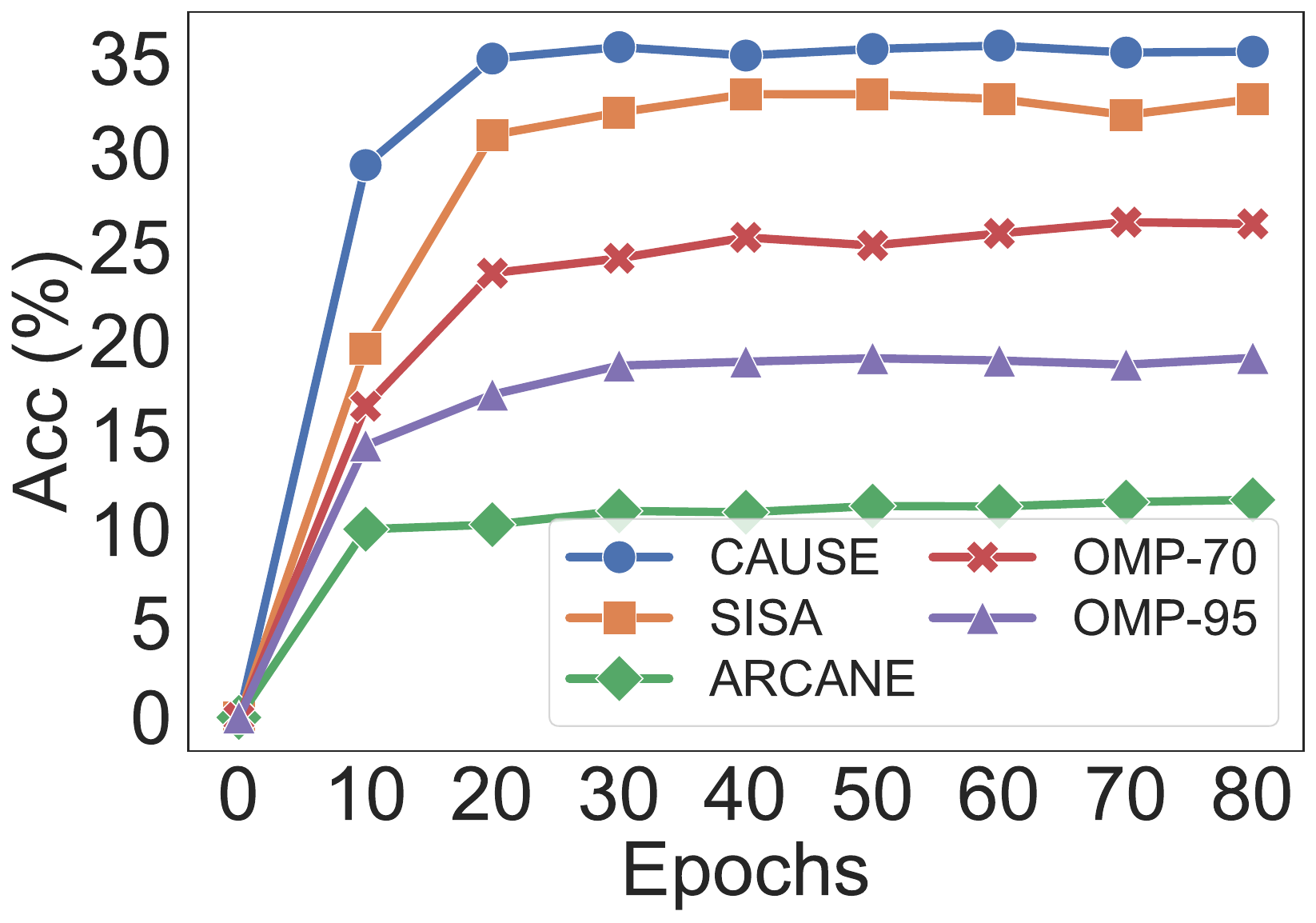}}

    \vspace{-0.5em}
    \caption{{Accuracy performance on CIFAR-100, SVHN, and CIFAR-10.}}
    \label{fig:combined_accuracy}
\end{minipage}
\vspace{-1.5em}
\end{figure}

We trained ResNet-34 on CIFAR-10, ResNet-34 on SVHN, VGG-16 on CIFAR-100, and MobileNetV2 on CIFAR-10 over 80 epochs and present the accuracy of a single model with data shard partitioned by \tool and banchmarks in 
Figure~\ref{fig:exp_acc_epoch}. On average, the accuracy of \tool is 
20.17\% higher than SISA (averaged from 30.32\%, 7.28\%, 37.82\%, and 5.26\%), 
158.50\% higher than ARCANE (averaged from 163.86\%, 94.28\%, 194.13\%, and 181.73\%), 
27.40\% higher than OMP-70 (averaged from 26.56\%, 5.65\%, 66.51\%, and 10.87\%), and 
15.12\% higher than OMP-95 (averaged from 49.63\%, 16.35\%, 88.51\%, and 87.47\%), respectively. The results of other models and data combinations can be found in Figure~\ref{fig:combined_accuracy}.

There are two key observations: (1) \tool consistently achieves significantly higher model accuracy than other systems. This indicates the advantage and importance of UCDP and SC. UCDP strategically partitions data according to user contributions, while SC facilitates the progressive training of increased amounts of data within individual sub-models. This dynamic allows \tool to outperform other systems such as SISA, ARCANE, OMP-70, and OMP-95 in terms of accuracy.
(2) The quality of the dataset plays an important role in the accuracy outcomes of the systems. For instance, when training ResNet-34 on different datasets like CIFAR-10 and SVHN, all systems, including \tool, show markedly better accuracy with SVHN than with CIFAR-10. Notably, when overall system performance is lower (as seen in Figure~\ref{fig:exp_acc_res_cifar10} and Figure~\ref{fig:exp_acc_vgg_cifar100}), the advantages of \tool become even more significant, highlighting its robustness across varying data conditions. In the remainder of this section, we illustrate and analyze the performance by training ResNet-34 on CIFAR-10 as default for demonstration purposes, if there is no specific statement. 

\begin{formal}
\textbf{Takeaway 1:} \textit{CAUSE demonstrates superior accuracy performance since UCDP enables strategic data partitioning based on user contributions, and SC allows progressive training of larger datasets in sub-models.}
\end{formal}

\subsection{Unlearning Speed Evaluation}
\label{subsec:exp_speed}

\begin{figure}[t]
\centering
\includegraphics[width=0.65\linewidth]{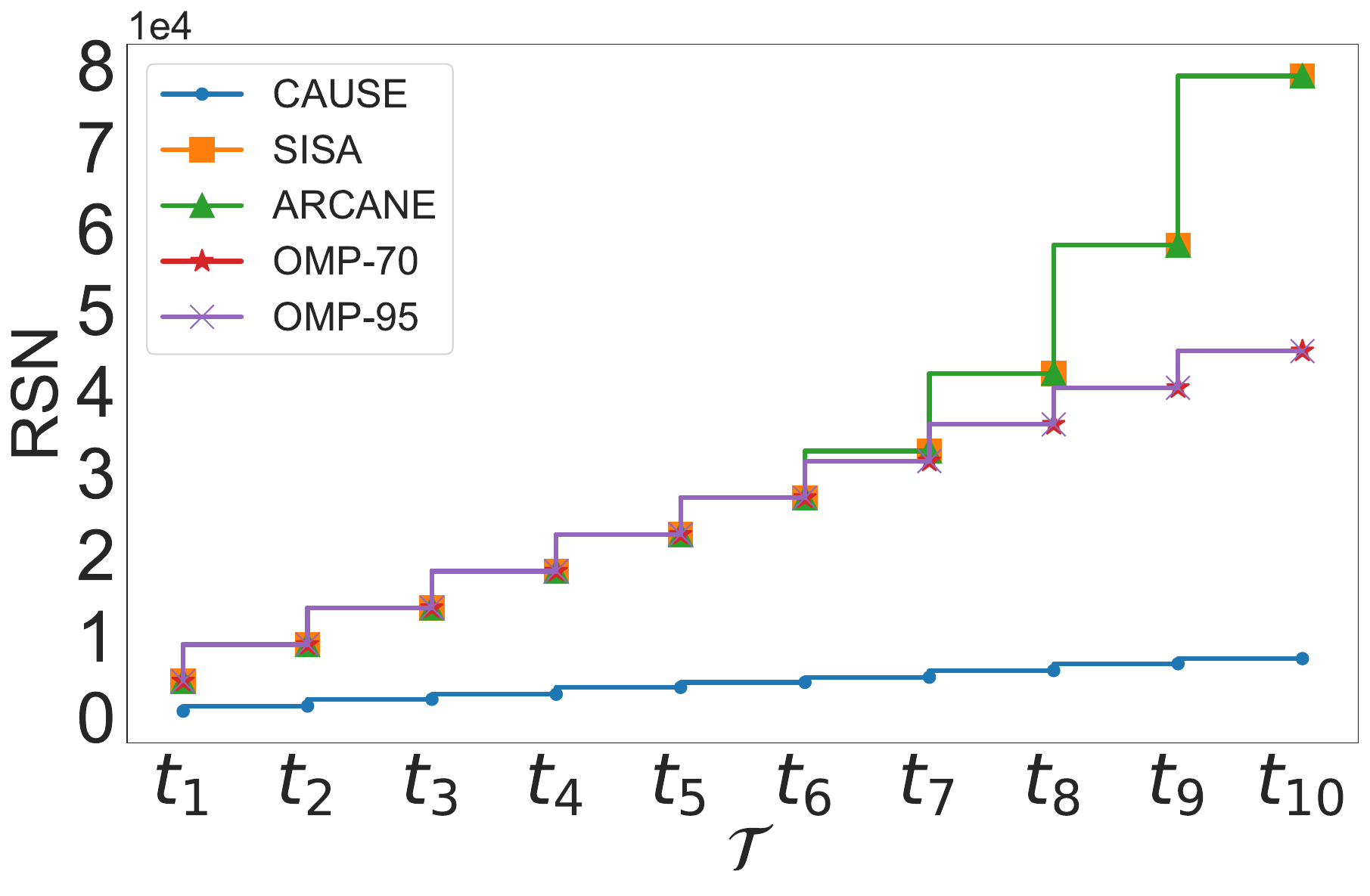}
\vspace{-0.5 em}
\caption{Retrained sample number over training rounds.}
\label{fig:rsn_over_t}
\vspace{-1. em}
\end{figure}

We compare the retraining time over 10 training rounds in Figure~\ref{fig:rsn_over_t}. Same as recent studies~\cite{bourtoule2021machine,yan2022arcane}, the retraining time is measured by the number of retrained samples, as the relationship between training time and the number of samples trained is linear. As time goes on, the number of samples retrained by \tool increases from 825 to 7,269 by 7.81 times. After 10 training rounds, the number of samples retrained by \tool is only 16.15\% of those retrained by OMP-70 and OMP-95 and 9.23\% of those retrained by SISA and ARCANE. 
Initially, in the first round (time slot $t_1$), the advantage of \tool is due to the design of RCMP and UCDP. RCMP helps reduce the memory size required for sub-models, enabling \tool to maintain more sub-models than SISA and ARCANE. In addition, UCDP segments data based on individual users, drastically cutting down the number of sub-models that need retraining when a user requests data unlearning. This advantage persists throughout subsequent training rounds.

Over time, the advantage of \tool becomes increasingly significant. Its superior performance largely stems from its innovative sub-model replacement strategy, FiboR. Unlike SISA, ARCANE, OMP-70, and OMP-95, which lack effective replacement strategies and thus face challenges in managing memory resources, \tool integrates FiboR and SC to save more sparse data information stored in sub-models. Those strategic approaches help \tool save memory and enhance the efficiency of identifying the optimal datasets for retraining. By leveraging these well-structured components, \tool demonstrates its excellent performance in unlearning speed, showcasing its capability to handle dynamic demands efficiently.

\begin{formal}
\textbf{Takeaway 2: }\textit{CAUSE speeds up retraining with much fewer samples involved. RCMP decreases the size of models, allowing more sub-models stored in memory; UCDP minimizes the number of sub-models needing retraining; and FiboR and SC efficiently manage memory and enhance the efficiency of identifying the start point of retraining.}
\end{formal}

\subsection{Energy Consumption Evaluation}
\label{subsec:exp_energy}

\begin{figure}[t]
\centering
\setlength{\textfloatsep}{5pt}
    \centering
    \subfigure[ResNet-34]{
    \includegraphics[width=0.42\linewidth]{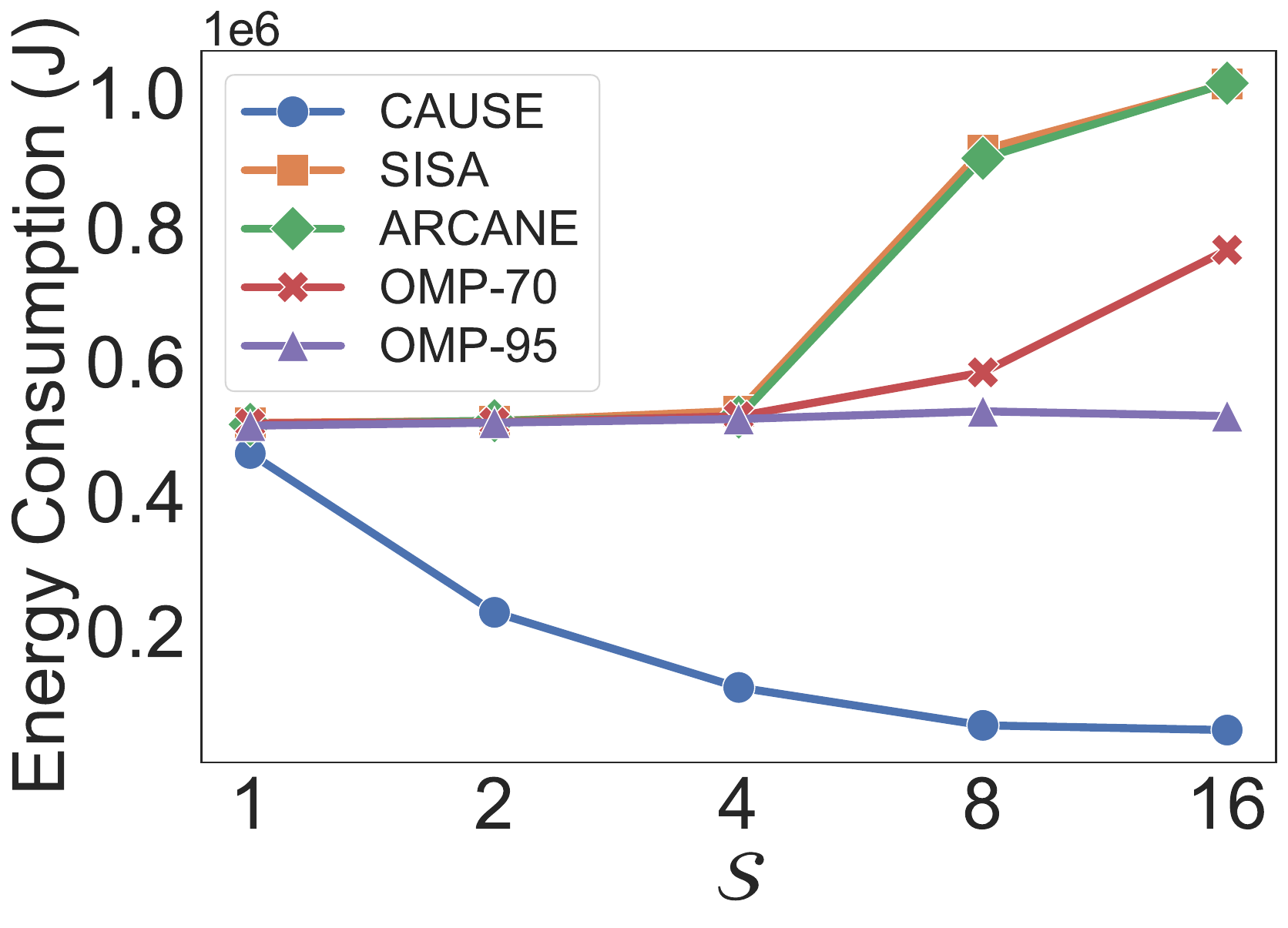}\label{fig:energy_shard_restnet}}\hspace{0.3 em}
    \subfigure[VGG-16]{
    \includegraphics[width=0.42\linewidth]{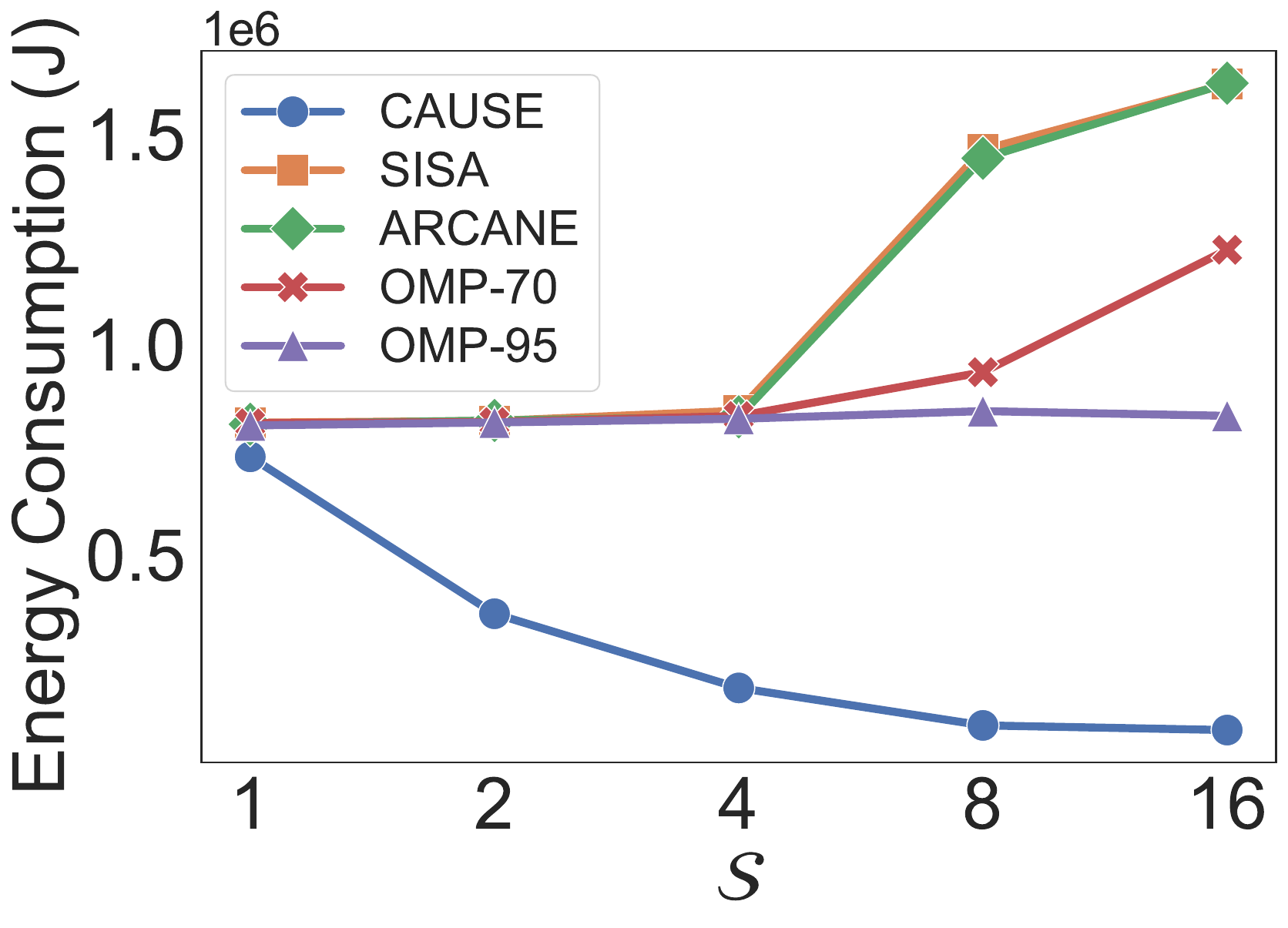}\label{fig:energy_shard_vgg}}\hspace{0.3 em}
    \subfigure[DenseNet-121]{
    \includegraphics[width=0.42\linewidth]{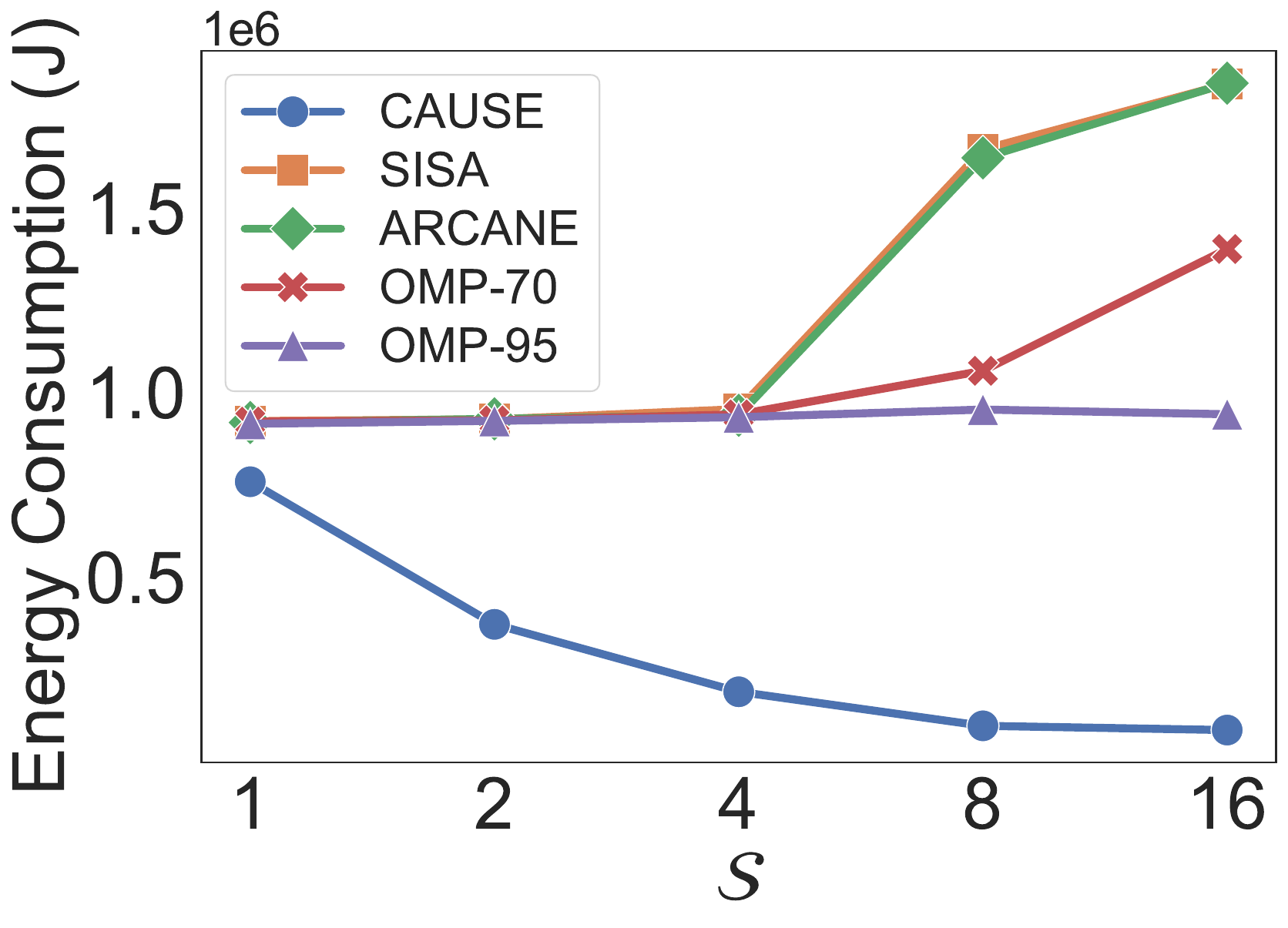}\label{fig:energy_shard_densenet}}\hspace{0.3 em}
    \subfigure[MobileNetV2]{
    \includegraphics[width=0.42\linewidth]{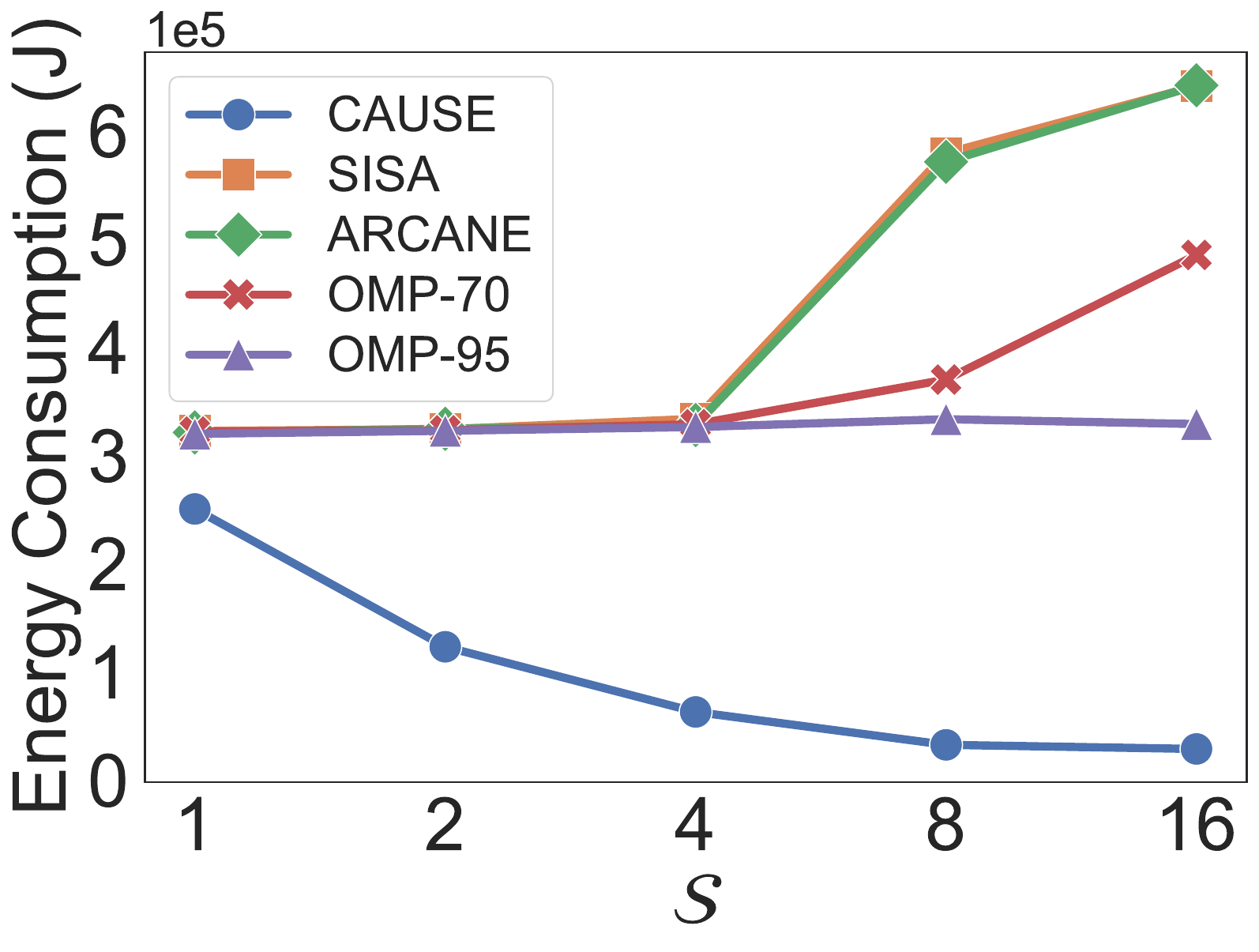}\label{fig:energy_shard_mobilenetv2}}
    \vspace{-0.5 em}
    \caption{Energy consumption comparison on various shard numbers.}
    \vspace{-1 em}
\label{fig:exp_energy_shard}
\end{figure}

\begin{figure}[t]
\centering
\setlength{\textfloatsep}{5pt}
    \centering
    \subfigure[ResNet-34]{
    \includegraphics[width=0.42\linewidth]{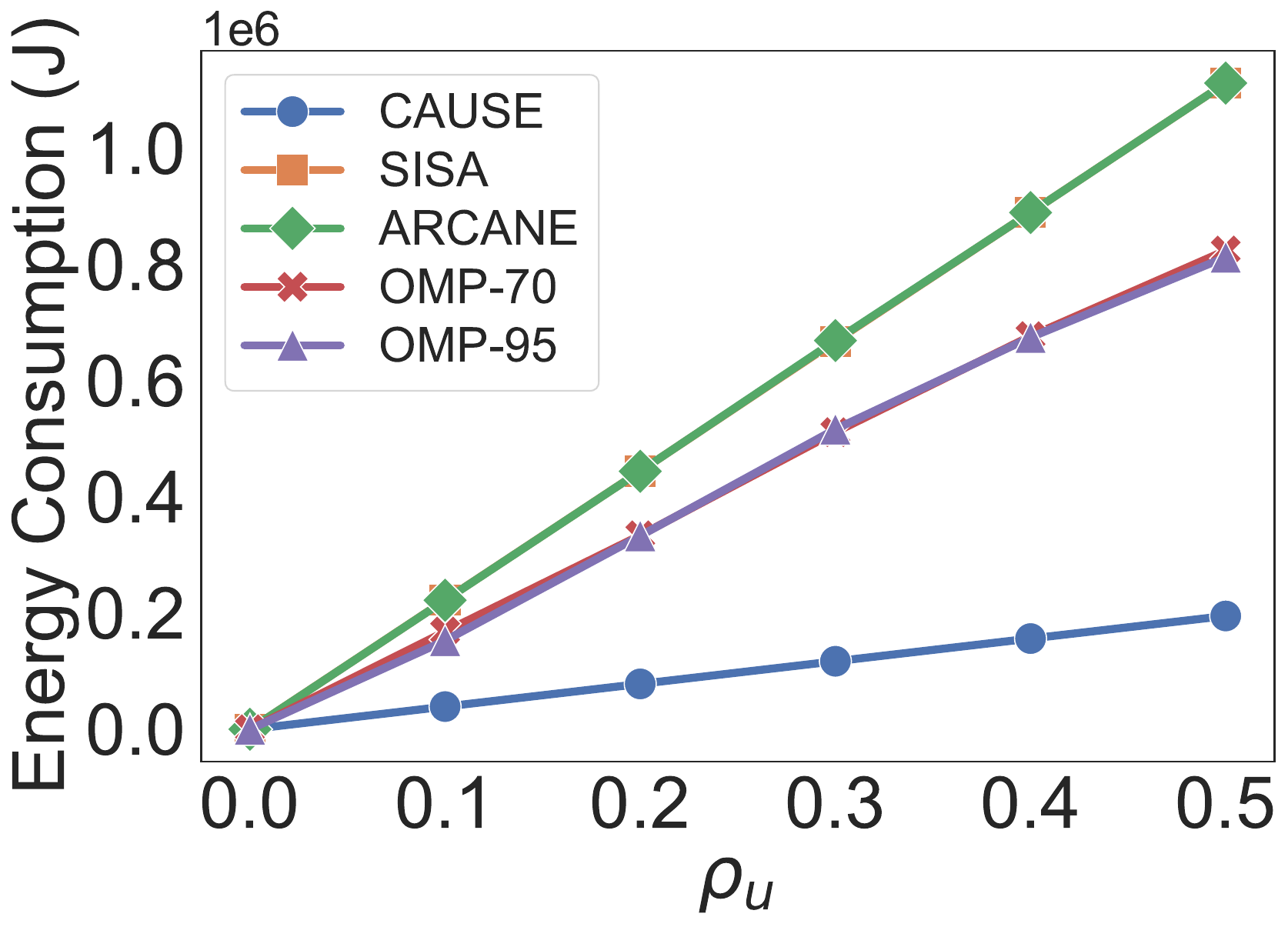}\label{fig:energy_ratio_restnet}}\hspace{0.3 em}
    \subfigure[VGG-16]{
    \includegraphics[width=0.42\linewidth]{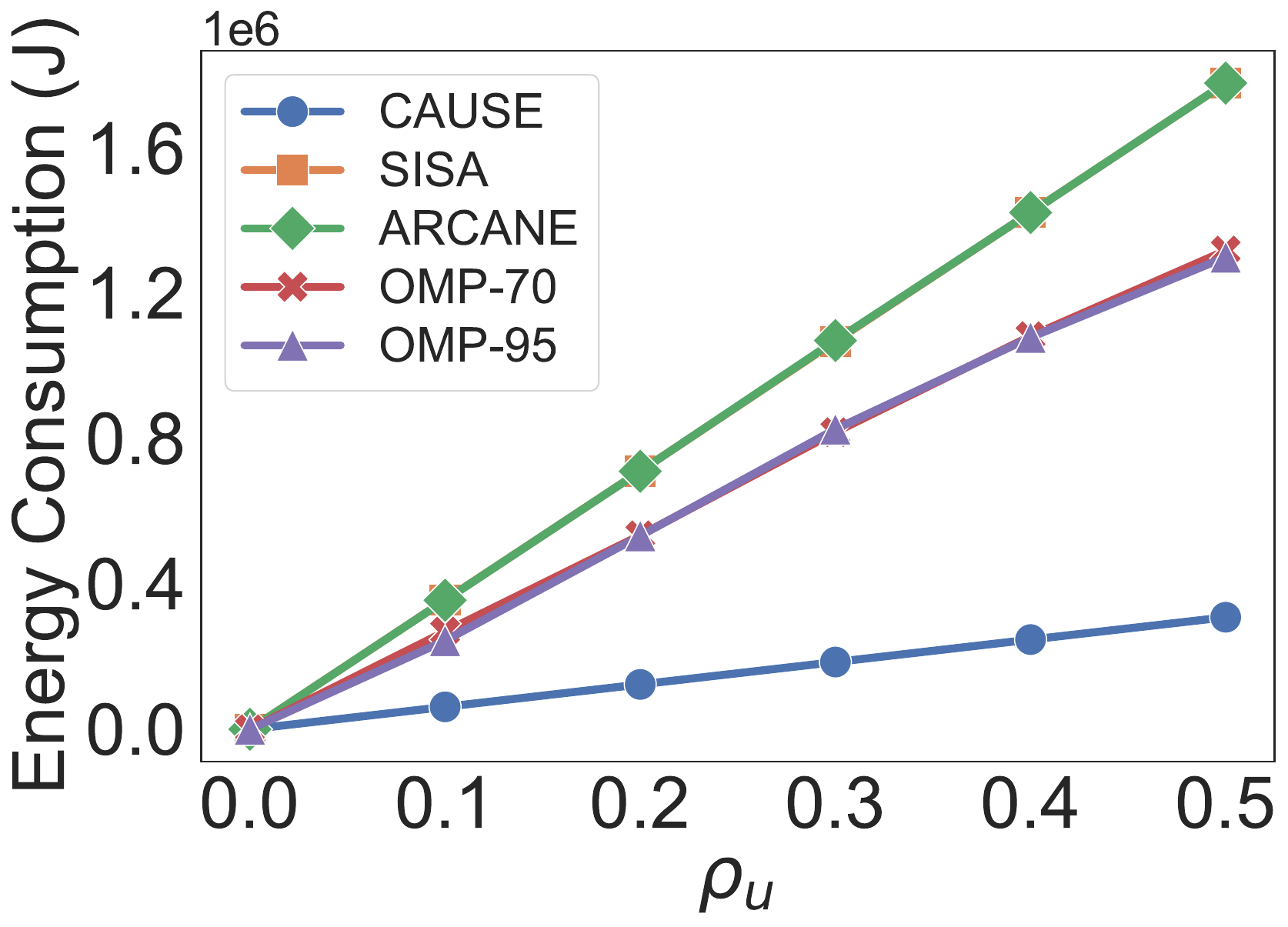}\label{fig:energy_ratio_vgg}}\hspace{0.3 em}
    \subfigure[DenseNet-121]{
    \includegraphics[width=0.42\linewidth]{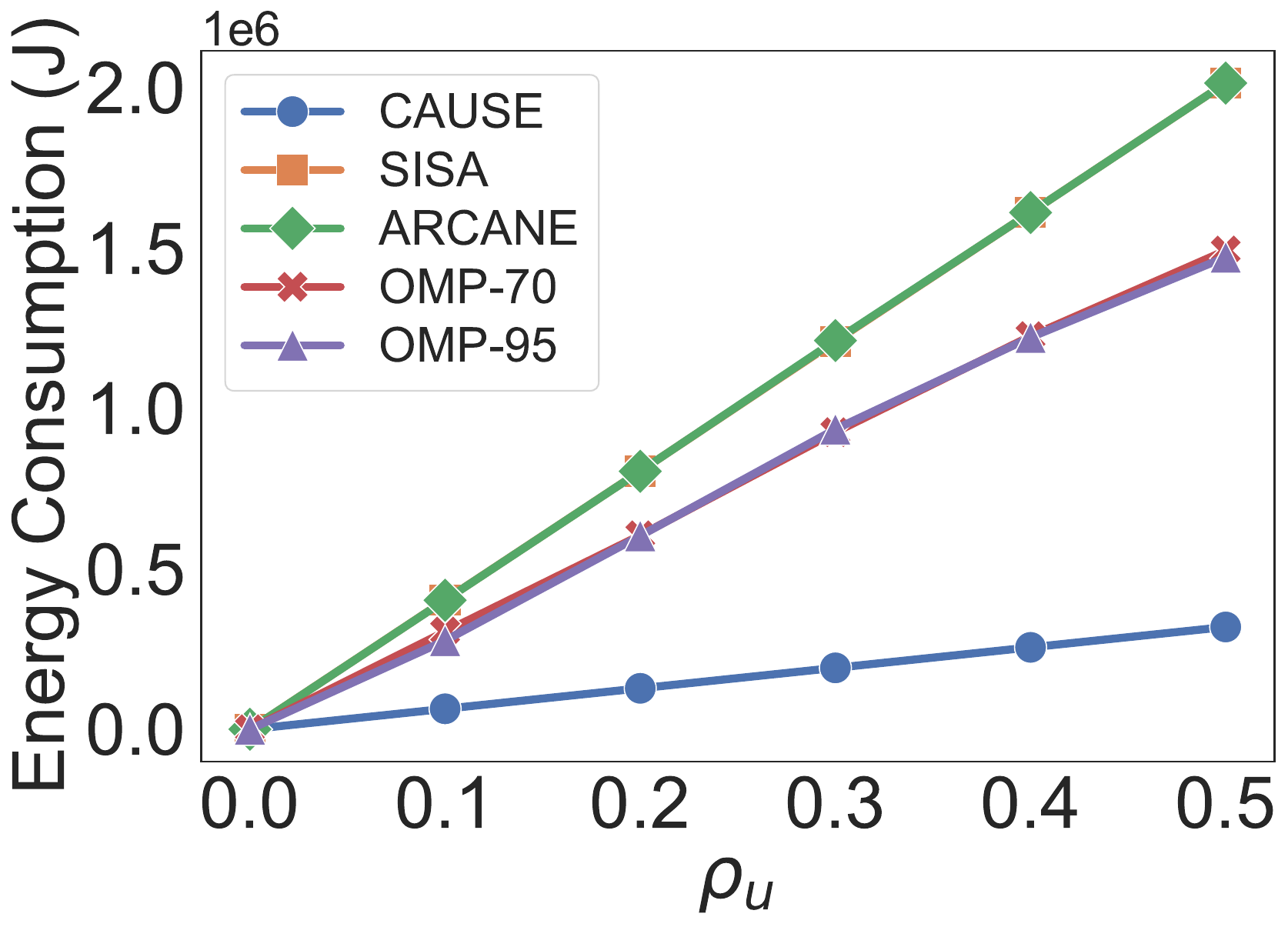}\label{fig:energy_ratio_densenet}}\hspace{0.3 em}
    \subfigure[MobileNetV2]{
    \includegraphics[width=0.42\linewidth]{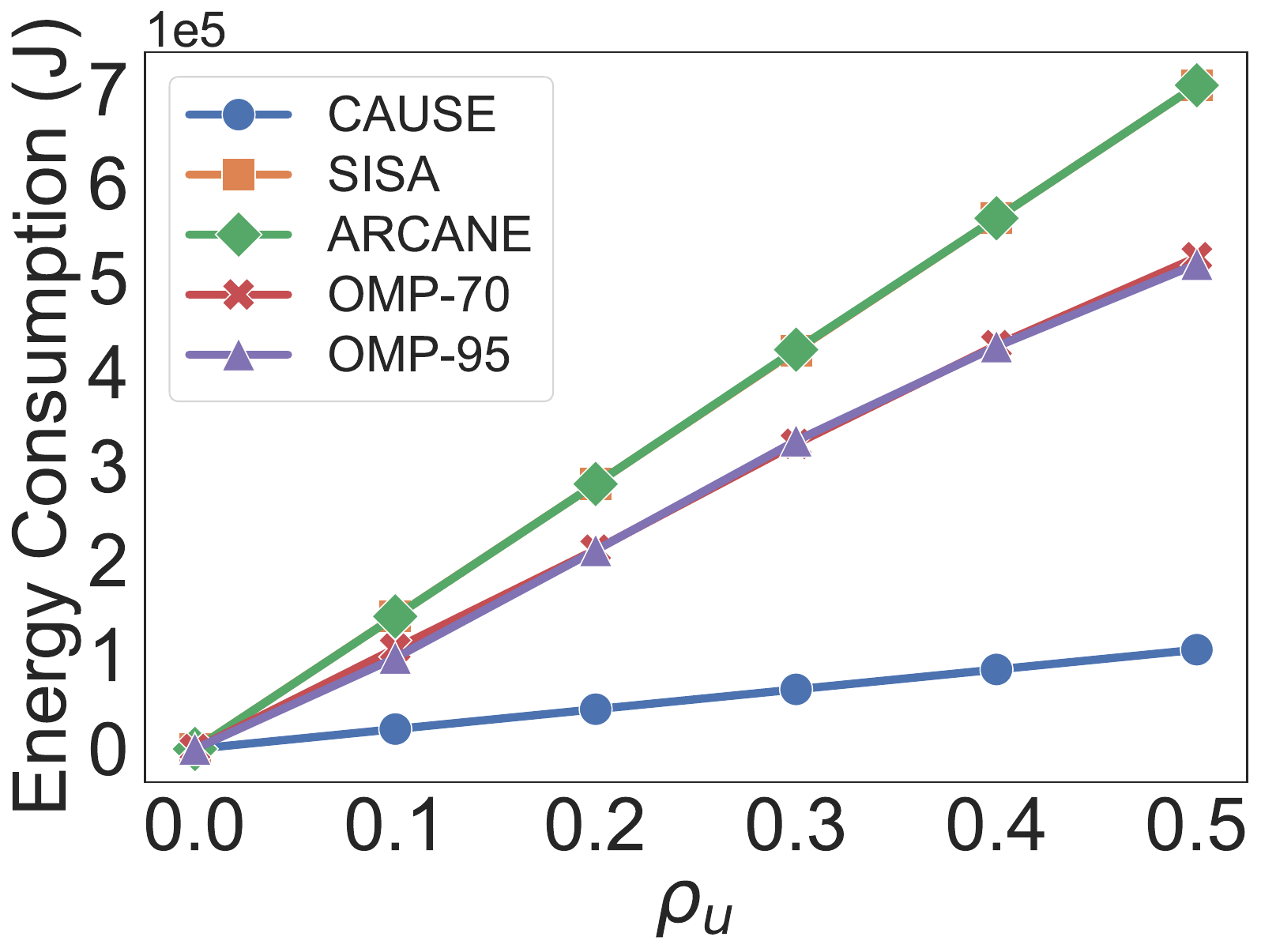}\label{fig:energy_ratio_vmobilenetv2}}
    \vspace{-0.5 em}
    \caption{Energy consumption comparison with various unlearning probabilities.}
    \vspace{-1. em}
\label{fig:exp_energy_ratio}
\end{figure}

We further conduct experiments to compare \tool against various exact unlearning benchmarks in terms of energy consumption, employing ResNet-34, VGG-16, DenseNet-121, and MobileNetV2 on CIFAR-10 with default experimental settings. Figure~\ref{fig:exp_energy_shard} and Figure~\ref{fig:exp_energy_ratio} show the energy consumption of \tool, SISA, ARCANE, OMP-70, and OMP-95 with different numbers of shards and various unlearning probabilities, respectively. As shown in Figure~\ref{fig:exp_energy_shard}, \tool significantly outperforms the other four exact unlearning systems in terms of energy consumption, with default unlearning probability $\rho_u=0.3$. Specifically, with $\mathcal{S}=16$ in the implementations of ResNet-34, VGG-16, DenseNet-121, and MobileNetV2, the energy consumption of \tool are only 25.05\% that of SISA, 25.21\% that of ARCANE, 30.14\% that of OMP-70, and 33.79\% that of OMP-95, on average. With the increase of $\mathcal{S}$, the energy consumption of \tool decreases while that of SISA, ARCANE, OMP-70, and OMP-95 increases. This is the same as the trends demonstrated in Figure~\ref{figs:rsn_vs_s} in~\S\ref{subsec:exp_shard}.
The experimental results resonate the observation in \S\ref{sec:design_motivation}: lower RSN results in lower energy consumption.
Figure~\ref{fig:exp_energy_ratio} depicts the energy consumption of all five exact unlearning systems with various unlearning probabilities and default shard number $\mathcal{S}=8$. In such settings, all five systems consume more energy with a higher unlearning probability. This is because more unlearning processes are required. Again, \tool achieves the lowest energy consumption, and outperforms other systems with large margins, \ie average 83.45\% over SISA, 83.46\% over ARCANE, 78.04\% over OMP-70 and 77.82\% over OMP-95.

\begin{formal}
\textbf{Takeaway 3: }\textit{Compared with benchmarks, CAUSE achieves the lowest energy consumption by a large margin, reducing it by over 66\%.}
\end{formal}

\subsection{Scalability in more Restrictive Scenarios}
\label{subsec:exp_scalability}

\begin{figure}[t]
\centering
\subfigure[RSN \vs memory capacities.]{
\includegraphics[width=0.47\linewidth]{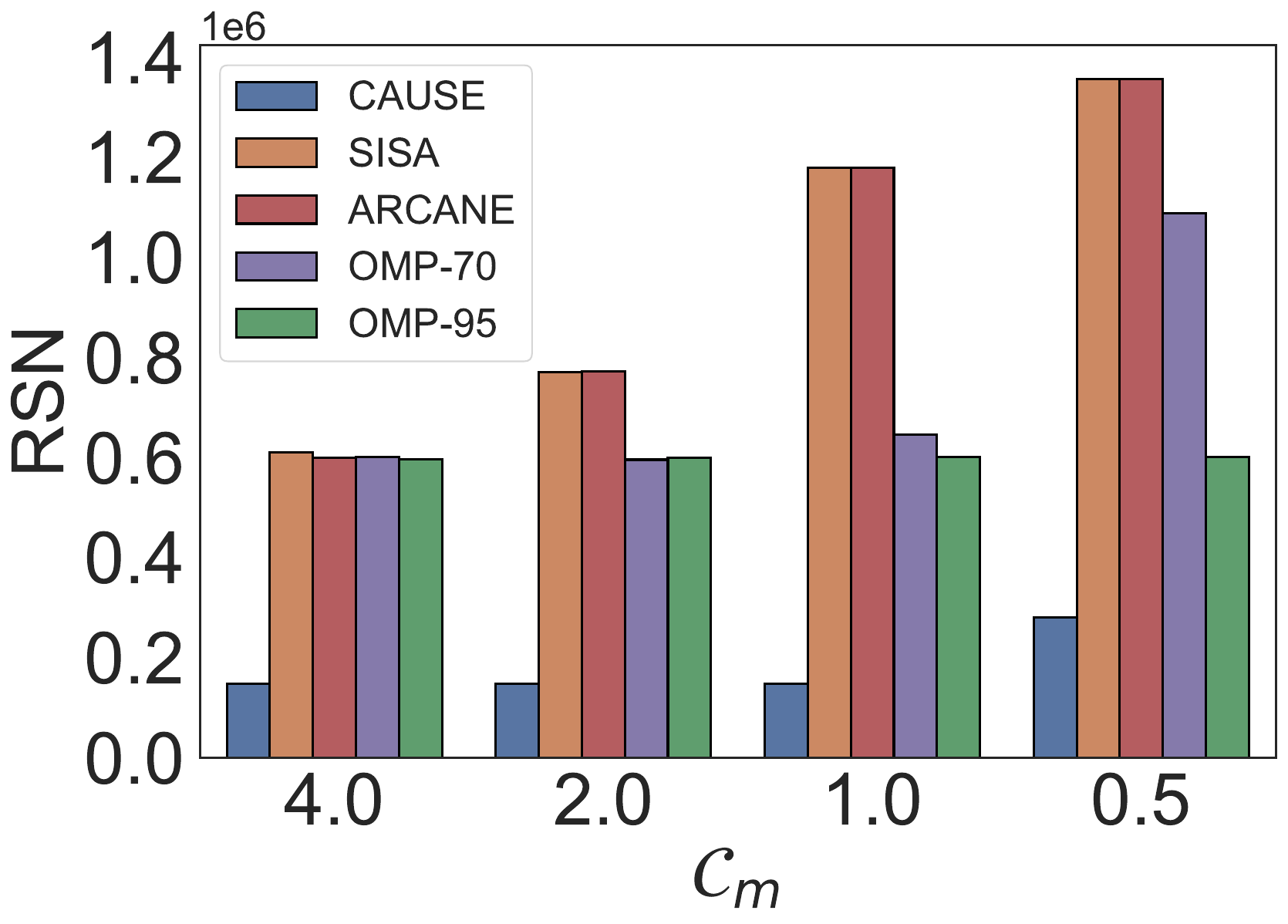}\label{figs:rsn_vs_m}}
\subfigure[RSN \vs unlearning probabilities.]{
\includegraphics[width=0.47\linewidth]{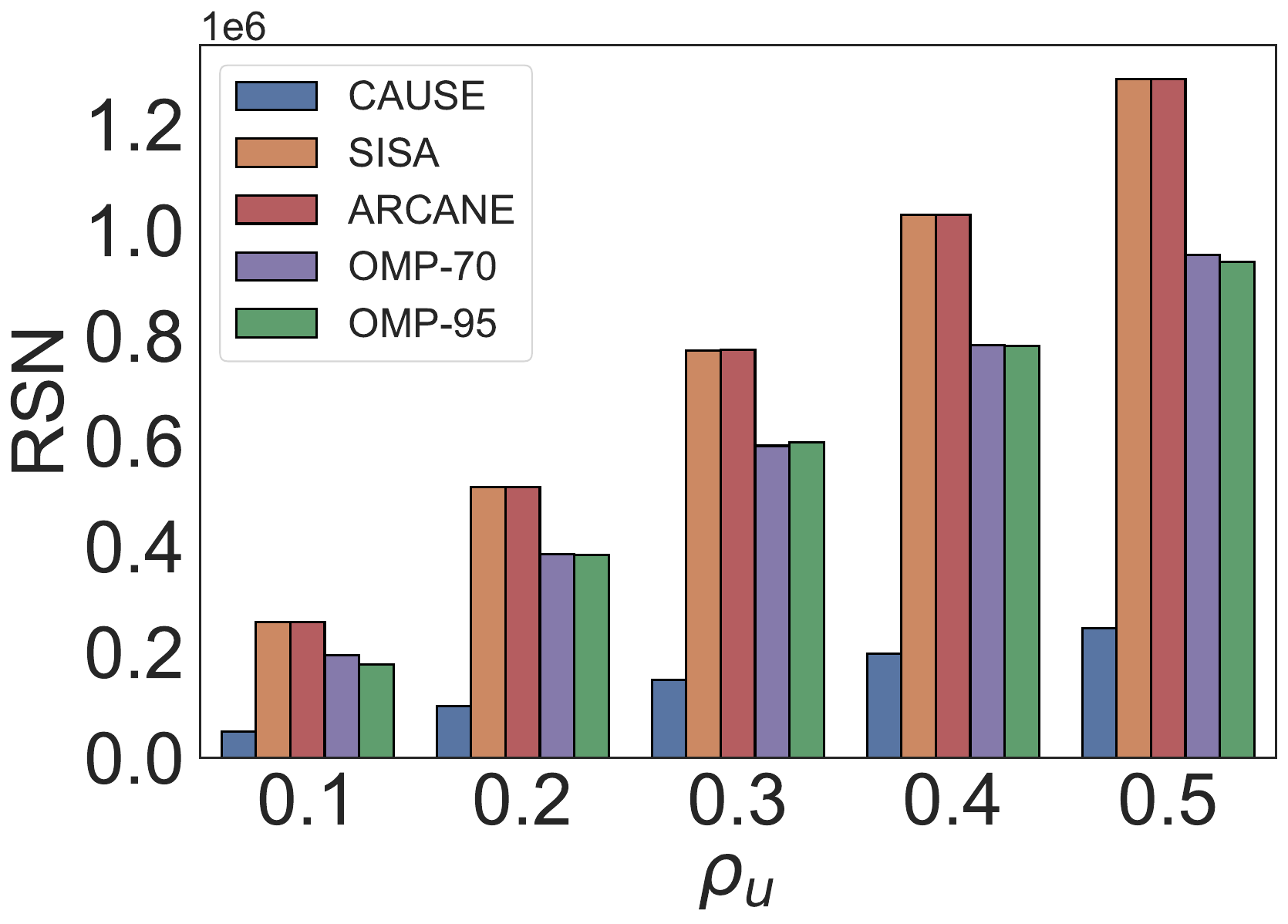}\label{figs:rsn_vs_ur}}
\vspace{-0.5 em}
\caption{A comparison of retrained sample number (RSN).}
\vspace{-1. em}
\end{figure}

Regarding scalability, \tool should be ensured with various memory capacities,
since the limited memory capacity of an edge device determines the number of sub-models that can be stored. 
Another key factor of scalability is the possibility of unlearning requests within time slots. Specifically, we conduct experiments under more restrictive scenarios, varying the memory capacity from 4.0GB to 0.5GB and increasing the unlearning possibility from 0.1 to 0.5 in steps of 0.1.

With various memory capacity settings, compared with benchmarks, \tool requires the lowest number of retrained samples and maintains an advantage of 80.75\% over SISA, 80.65\% over ARCANE, 75.38\% over OMP-70, and 70.93\% over OMP-95 (as shown in Figure~\ref{figs:rsn_vs_m}). As memory capacity decreases, all systems show an increase in the number of samples that need to be retrained due to the reduced number of sub-models that can be stored, which in turn slows down the unlearning process.
Furthermore, Figure~\ref{figs:rsn_vs_ur} illustrates how an increase in the probability of unlearning requests $\rho_{u}$, impacting the frequency of unlearning requests, affects system performance. Once again, \tool outperforms other exact unlearning systems by significant margins, demonstrating an average retraining speed that is 80.86\% faster than SISA, 80.87\% faster than ARCANE, 74.60\% faster than OMP-70, and 74.36\% faster than OMP-95. With the rise in unlearning request probability from 0.1 to 0.5, the number of samples retrained by \tool and other systems also increases dramatically, reflecting a proportional response to the heightened demand for unlearning.

Figures~\ref{figs:rsn_vs_m} and~\ref{figs:rsn_vs_ur} highlight \tool's capacity to handle memory constraints and increased unlearning requests. As discussed in~\S\ref{subsec:exp_speed}, the advantages of \tool in scalability are from the strategic design of its components, which optimize performance across varying operational conditions. This makes \tool particularly effective on resource-constrained devices where resource limitations and dynamic model management are critical concerns.

\begin{formal}
\textbf{Takeaway 4: }\textit{Even in more restrictive scenarios, CAUSE outperforms benchmarks by a large margin, demonstrating that the strategic design of its components optimizes performance across varying operational conditions.}
\end{formal}

\subsection{Ablation Study}
\label{subsec:exp_shard}

\begin{figure}[t]
\centering
\centering
\subfigure[ResNet-34 on CIFAR-10]{
\includegraphics[width=0.2\textwidth]{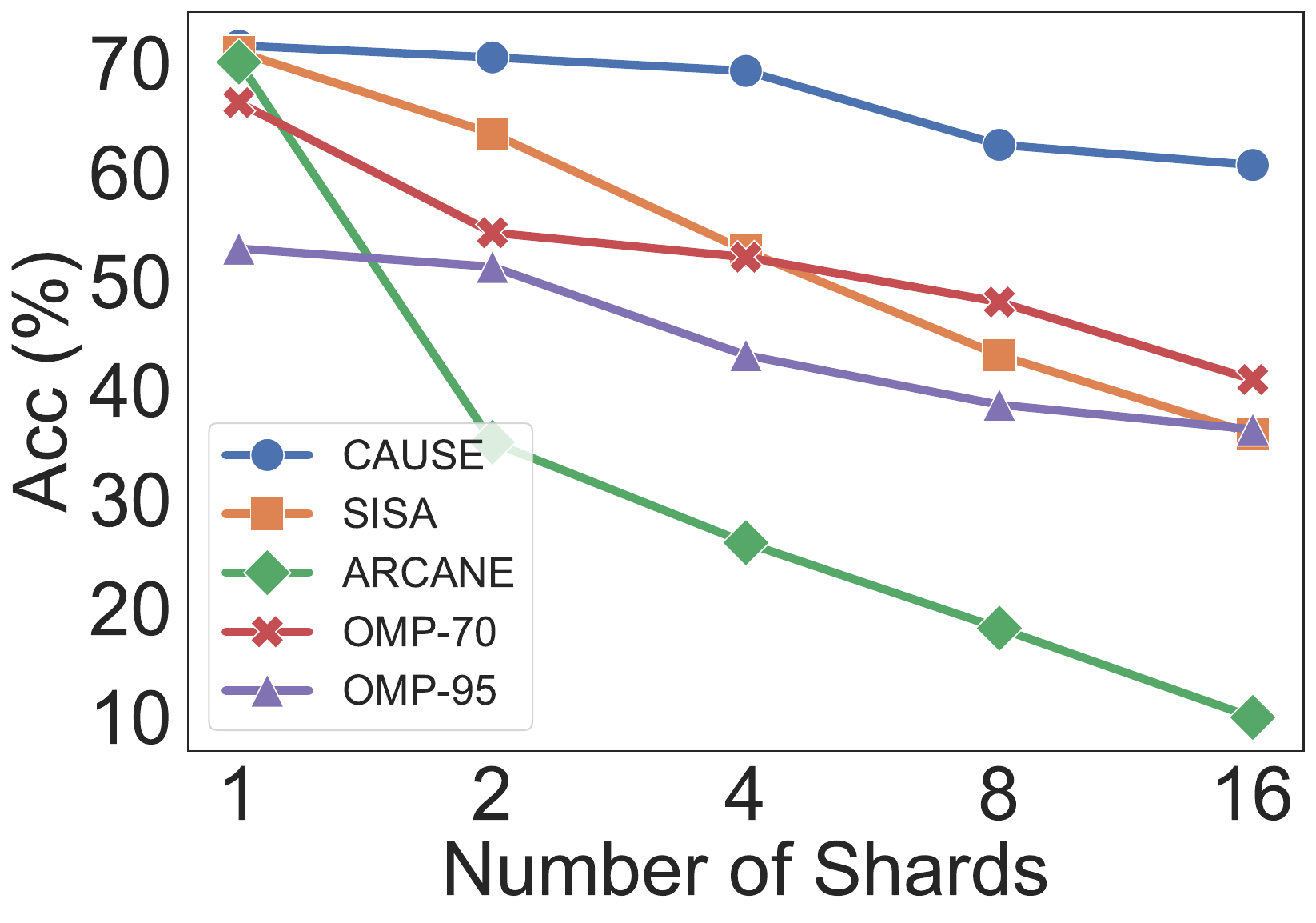}\label{fig:shards_r_c}}\hspace{0.5 em}
\subfigure[ResNet-34 on SVHN]{
\includegraphics[width=0.2\textwidth]{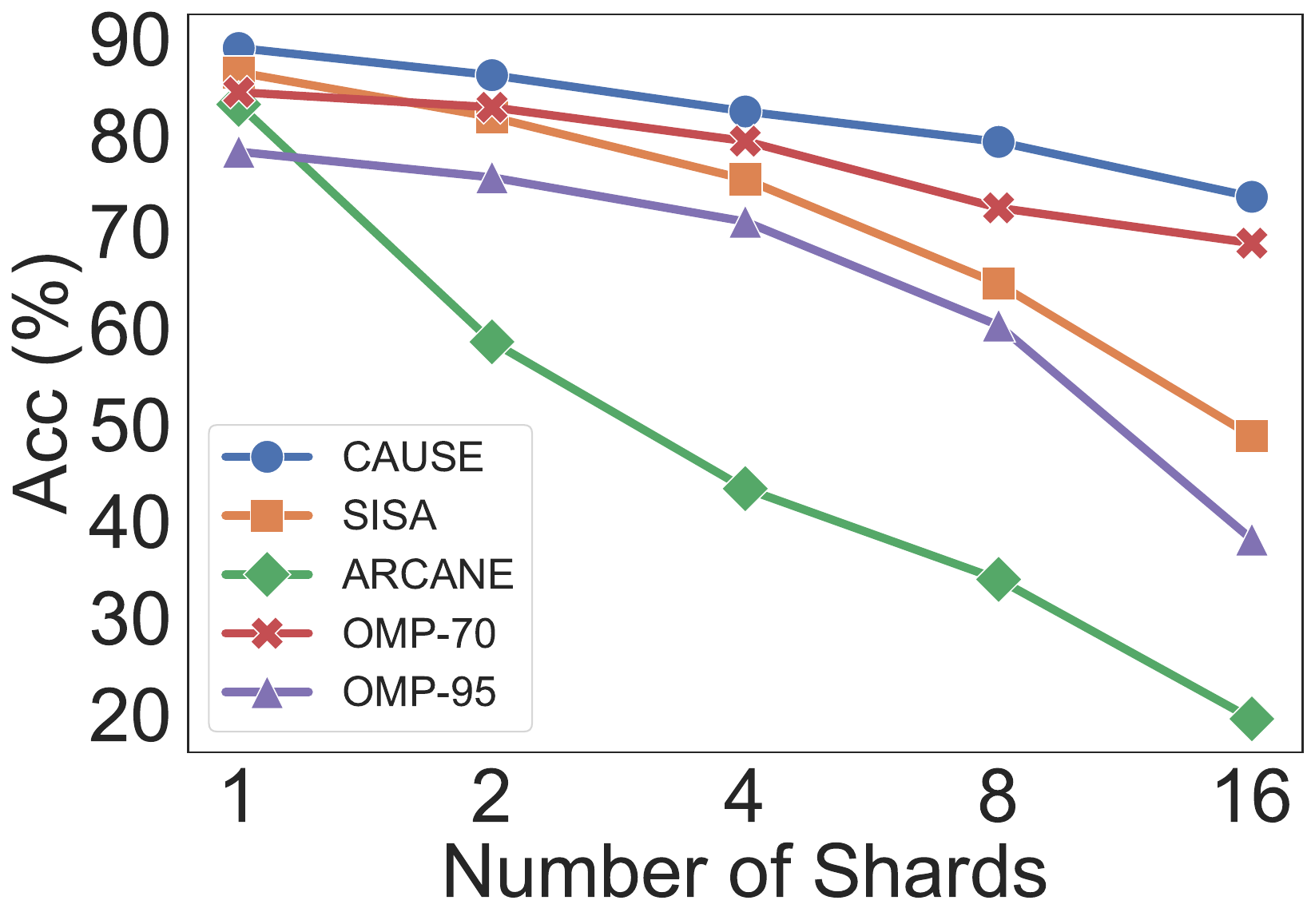}\label{fig:shards_r_s}}
\subfigure[VGG-16 on CIFAR-100]{
\includegraphics[width=0.2\textwidth]{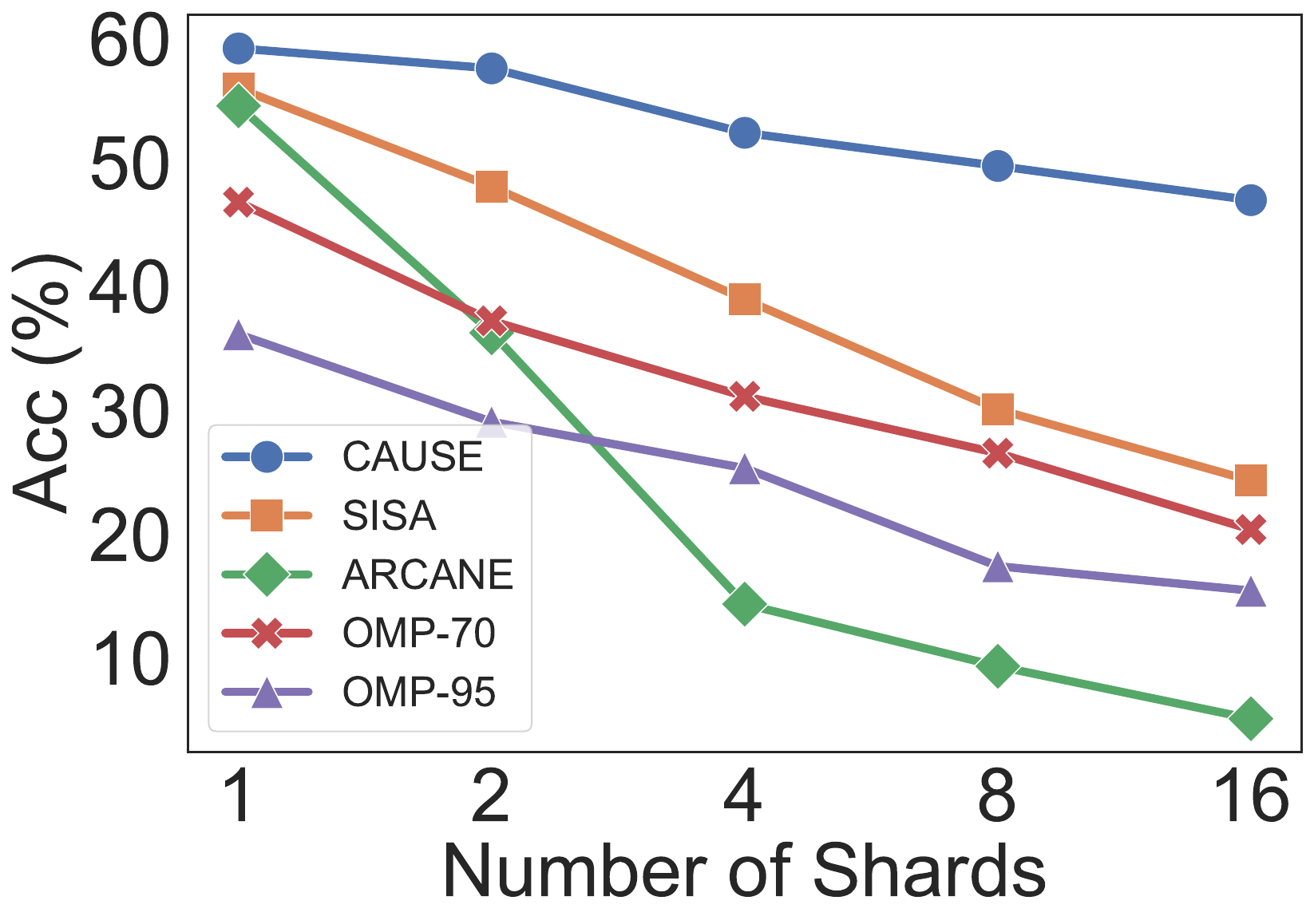}\label{fig:shards_v_c}}\hspace{0.5 em}
\subfigure[MobileNetV2 on CIFAR-10]{
\includegraphics[width=0.2\textwidth]{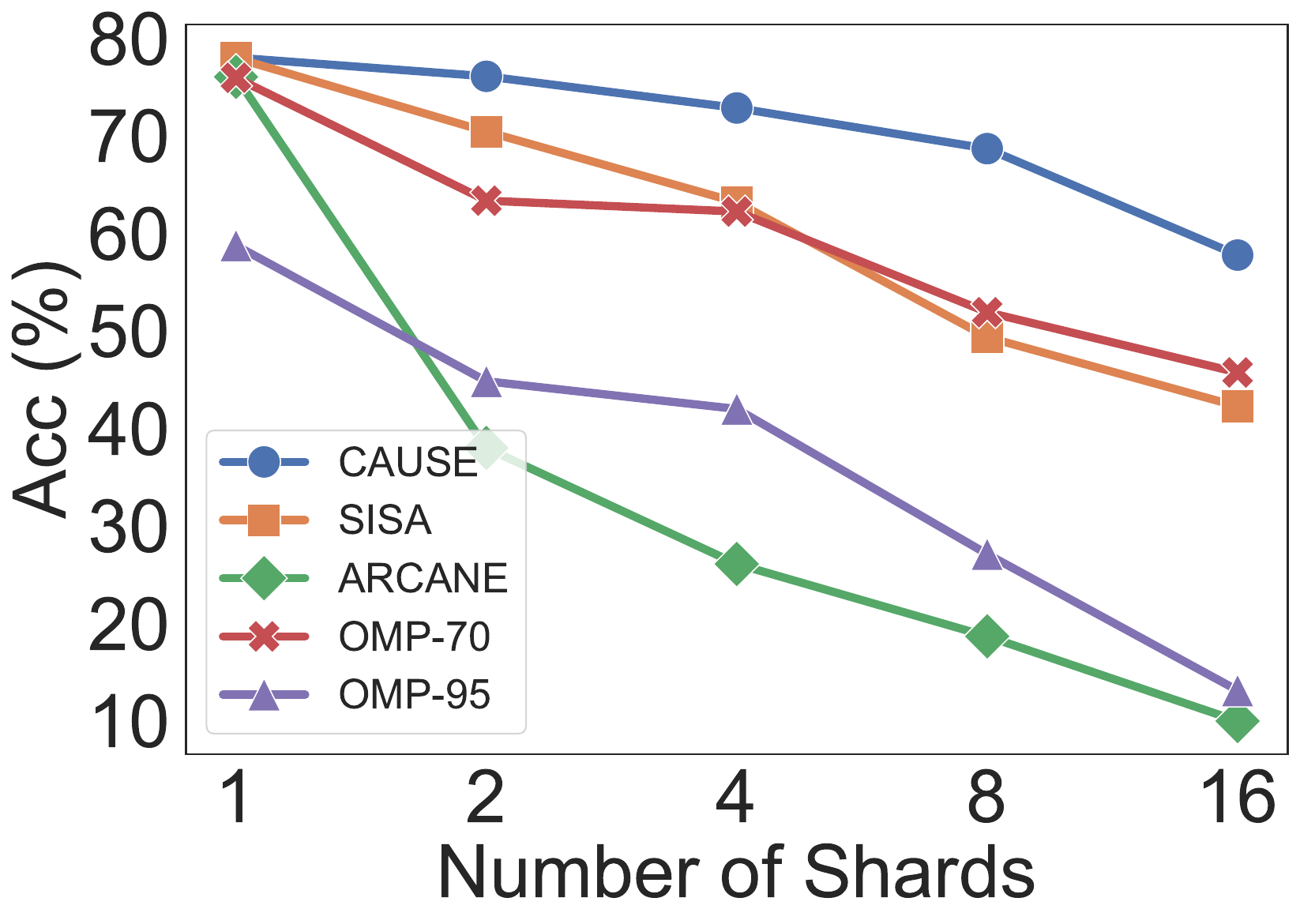}\label{fig:shards_v2_c}}
\vspace{-0.5 em}
\caption{Accuracy comparison with various shard numbers ($\mathcal{S}=1, 2, 4, 8, 16$).}
\vspace{-0.5 em}
\label{fig:exp_shards}
\vspace{-1. em}
\end{figure}

\noindent \textbf{Impact of shard number $\mathcal{S}$.}
Figure~\ref{fig:exp_shards} depicts the impact of shard number $\mathcal{S}$ on the accuracy of various systems under different experimental setups, such as ResNet-34 on CIFAR-10, ResNet-34 on SVHN, VGG-16 on CIFAR-100 and MobileNetV2 on CIFAR-10, running on t3.medium. As previously highlighted in~\S\ref{subsec:data_partition}, a higher shard count tends to reduce accuracy, a trend confirmed by the results depicted in Figure~\ref{fig:motivation_shards}. The results shown in Figure~\ref{fig:exp_shards} further prove this observation. When training ResNet-34 on CIFAR-10 in Figure~\ref{fig:shards_r_c}, the accuracy for all systems declines as $\mathcal{S}$ increases from 1 to 16. For instance, \tool's accuracy drops from 70.56\% to 60.68\%, SISA's from 70.06\% to 36.02\%, ARCANE's from 70.13\% to 10.00\%, OMP-70's from 66.41\% to 41.01\%, and OMP-95's from 53.02\% to 36.43\%. Notably, \tool maintains higher accuracy levels compared to other systems throughout the shard increase. 
With $\mathcal{S}=16$, \tool's relative accuracy advantage becomes even more pronounced: it achieves 1.68$\times$, 6.07$\times$, 1.48$\times$, and 1.67$\times$ the accuracy of SISA, ARCANE, OMP-70, and OMP-95, respectively. This advantage is attributed to SC, which optimally adjusts the shard number to allow individual sub-models to learn more data effectively, thereby enhancing overall accuracy.
With similar performance and advantages, Figure~\ref{fig:shards_r_s}, Figure~\ref{fig:shards_v_c} and Figure~\ref{fig:shards_v2_c} show that \tool can perform training tasks effectively when training ResNet-34 on SVHN, VGG-16 on CIFAR-100 and MobileNetV2 on CIFAR-10. On average, \tool achieves an average of 1.15$\times$, 1.34$\times$ and 1.17$\times$ accuracy achieved by SISA, 1.72$\times$ 2.22$\times$ and 2.09$\times$ accuracy achieved by ARCANE, 1.06$\times$, 1.64$\times$ and 1.18$\times$ accuracy achieved by OMP-70, 1.27$\times$, 2.15$\times$ and 1.90$\times$ accuracy achieved by OMP-95, respectively. This demonstrates the superior performance of \tool in providing accurate training models, compared to other systems. 

\begin{figure}[t]
\centering
\includegraphics[width=0.44\linewidth]{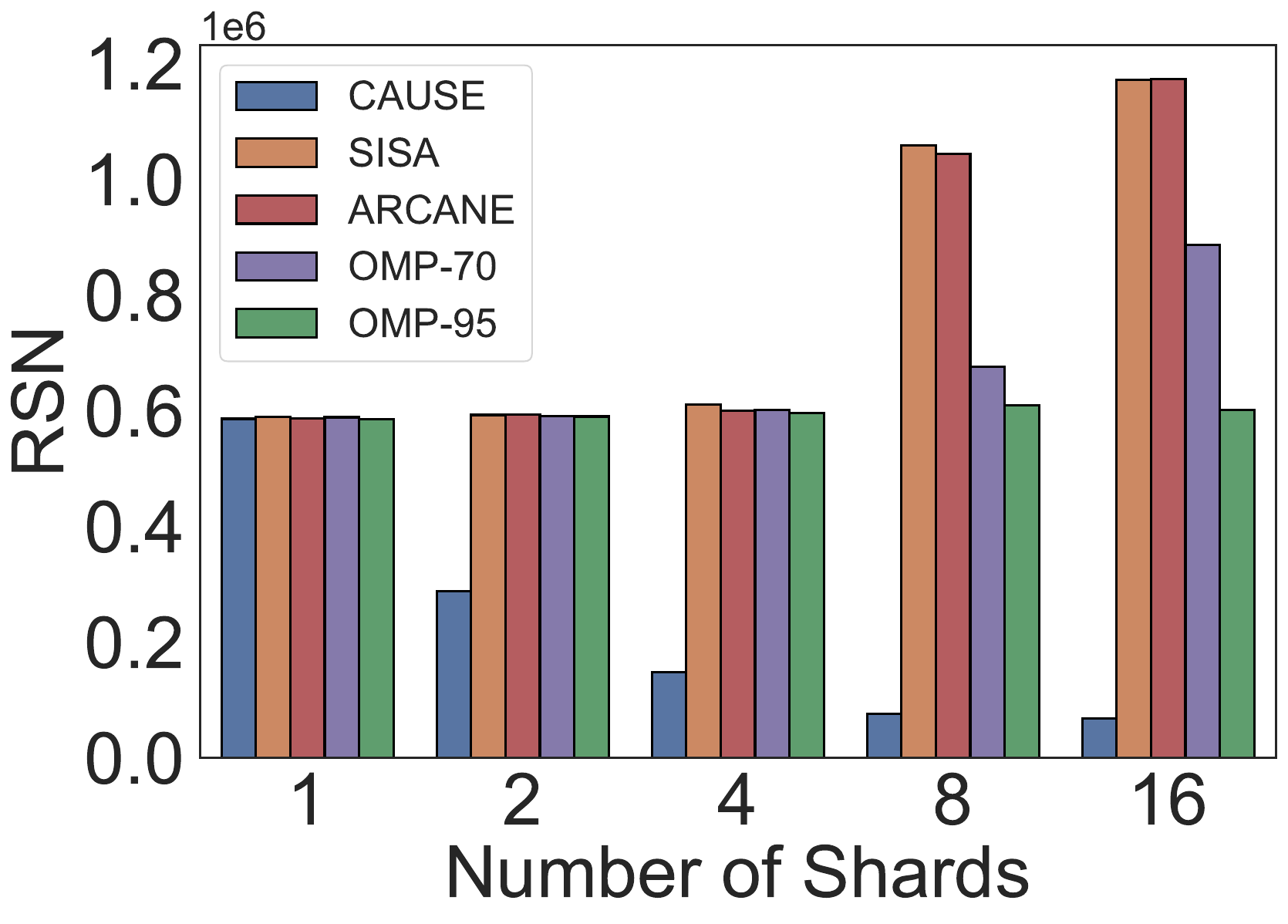}
\captionof{figure}{Retrained sample number comparison with various shard numbers ($\mathcal{S}=1, 2, 4, 8, 16$).}
    \vspace{-1. em}
\label{figs:rsn_vs_s}
\end{figure}

Figure~\ref{figs:rsn_vs_s} shows the impact of shard number $\mathcal{S}$ on the number of retrained samples when training ResNet-34 on CIFAR-10. As $\mathcal{S}$ increases, the number of retained samples by \tool significantly decreases from 586,482 to 67,732 by 88.45\%. In contrast, the numbers for SISA, ARCANE, OMP-70, and OMP-95 show increases.
Interestingly, the trends for SISA and ARCANE are opposite to what has been reported in prior studies~\cite{bourtoule2021machine,yan2022arcane}, likely due to the unique challenges of edge machine unlearning where users typically request to forget only their specific data. Unlike other systems that use uniform or class-based partitioning methods, which can lead to more sub-models requiring retraining upon receiving unlearning requests, the use of UCDP partitions data according to ownership, effectively minimizing such needs on the resource-constrained device. This highlights the adaptability and efficiency of \tool in handling unlearning requests.

\noindent \textbf{Impact of data partition method.}\label{subsec:exp_dp}
As discussed in~\S\ref{subsec:exp_speed}, ~\S\ref{subsec:exp_scalability}, and~\S\ref{subsec:exp_shard}, UCDP is a key factor contributing to \tool's superior unlearning speed compared to other systems. To further explore UCDP's role in enhancing \tool's performance, we conducted an ablation study. This involved comparing the standard \tool with two variants: \tool-U, which replaces UCDP with a uniform-based data partition method from~\cite{bourtoule2021machine}, and \tool-C, which adopts a class-based partition strategy from~\cite{yan2022arcane}.

\begin{figure}[t]
\centering
\setlength{\textfloatsep}{5pt}
\begin{minipage}[c]{\linewidth}
\centering
    \subfigure[Accuracy \vs shard numbers]{
    \includegraphics[width=0.44\linewidth]{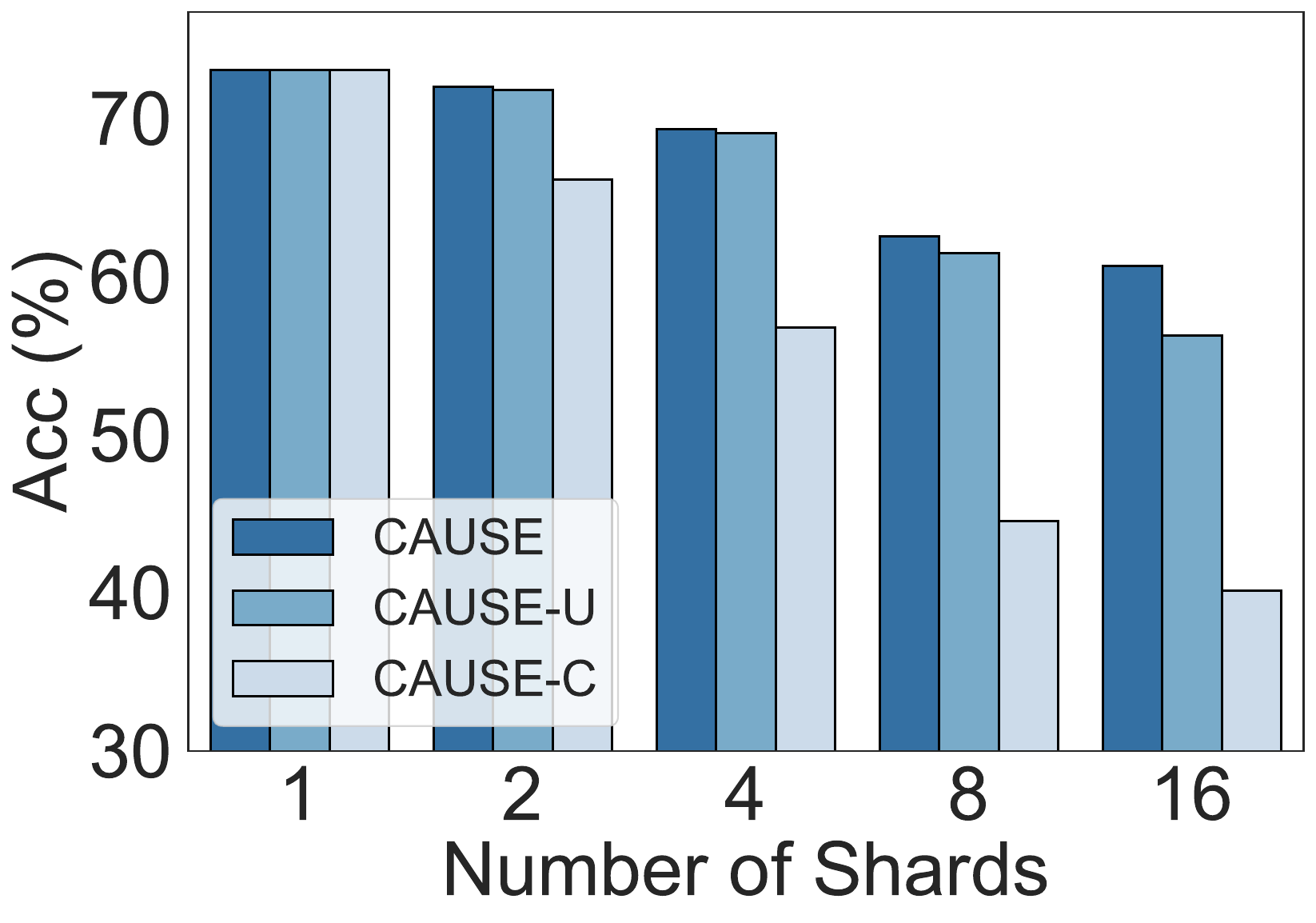}\label{fig:dp_s}}
    \subfigure[RSN \vs shard numbers]{
    \includegraphics[width=0.44\linewidth]{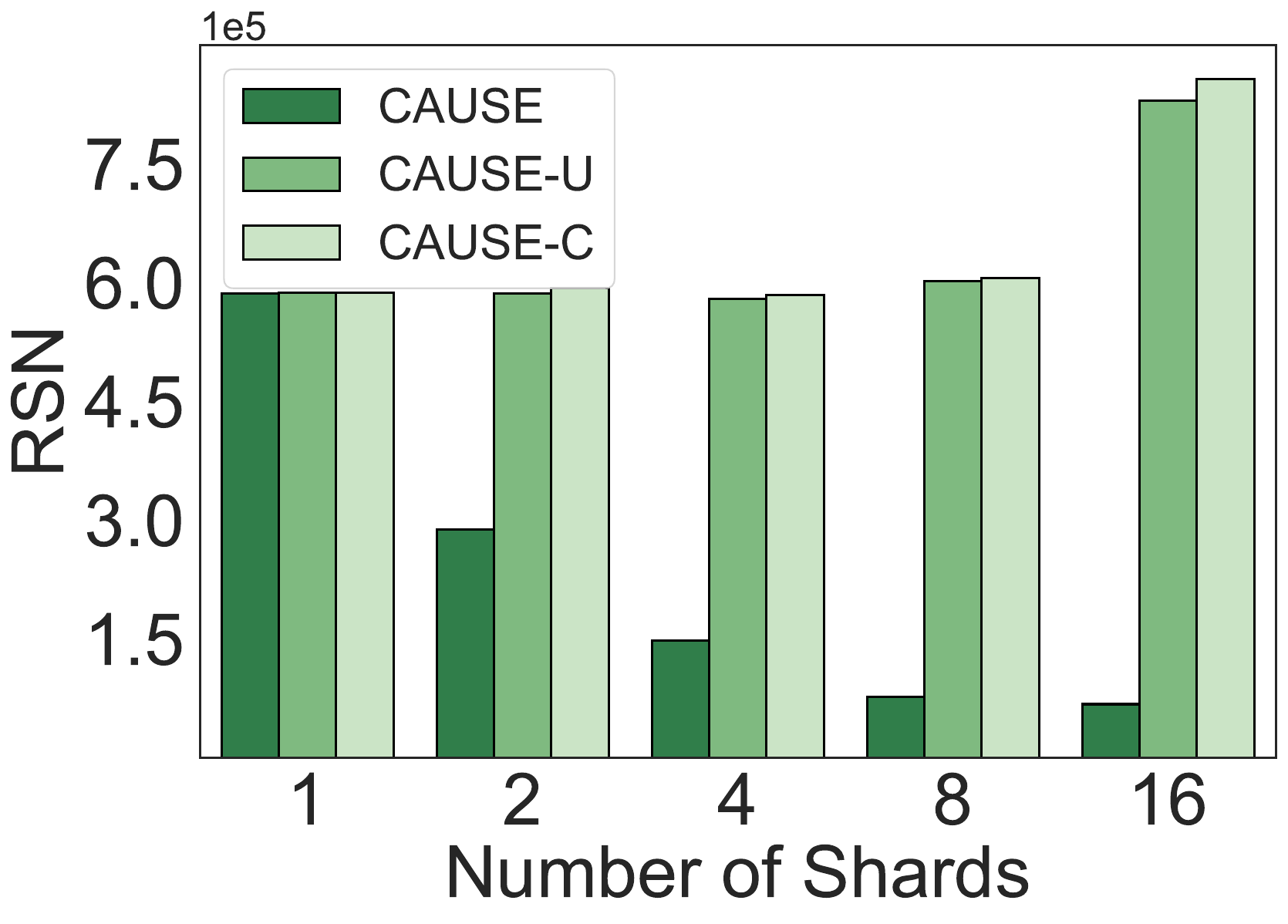}\label{fig:dp_rsn_s}}
    \subfigure[RSN \vs unlearning probabilities.]{
    \includegraphics[width=0.44\linewidth]{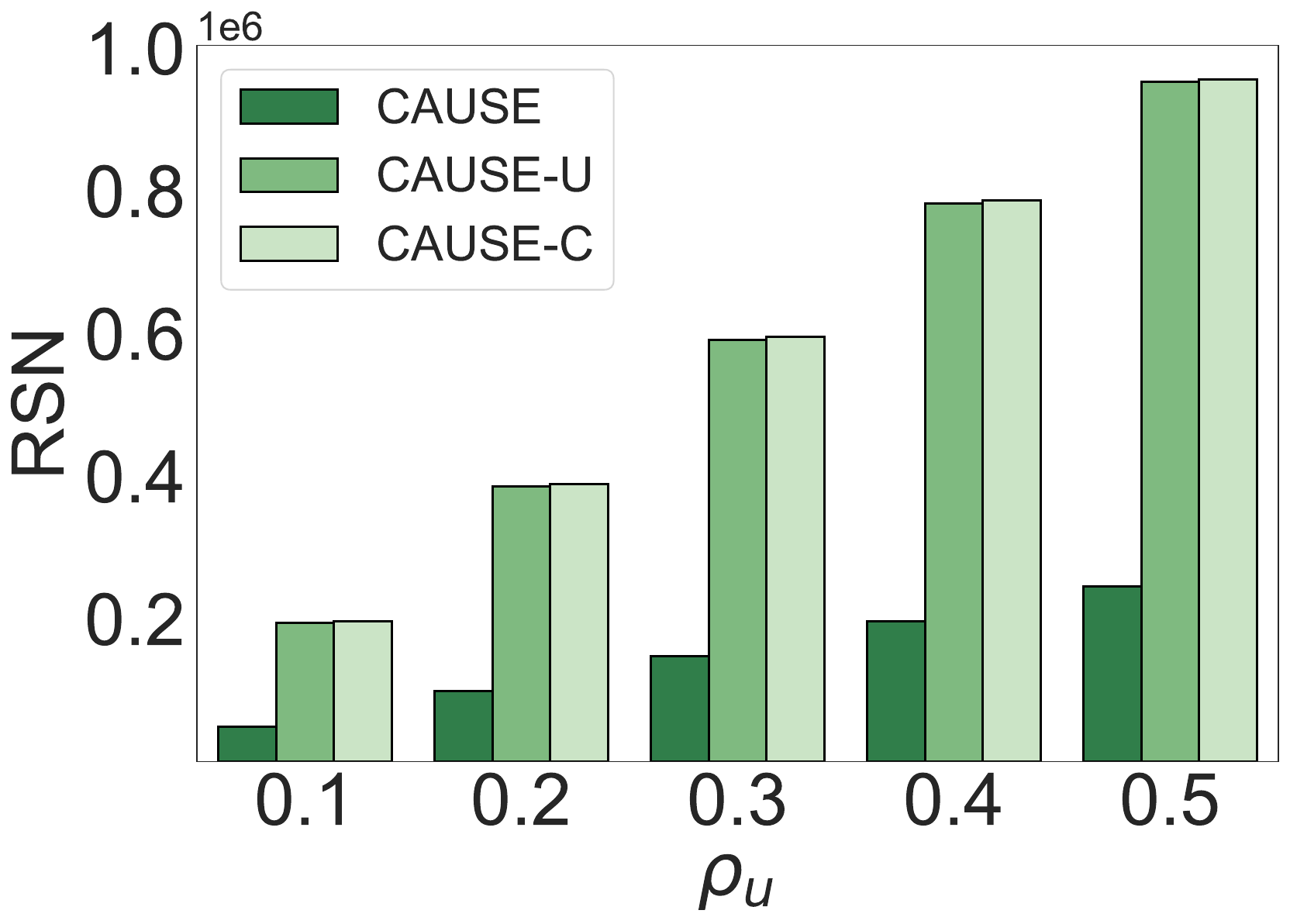}\label{fig:dp_ur}}
    \vspace{-0.5 em}
    \caption{Ablation study on various data partition methods.}
    \vspace{-1. em}
    \label{fig:exp_dp}
\end{minipage}
\vspace{-1. em}
\end{figure}

Figure~\ref{fig:exp_dp} demonstrates the performance of \tool, \tool-U, and \tool-C in terms of the accuracy and the number of retrained samples. We can see that the \tool achieves the highest accuracy and the lowest retrained sample number among the three versions of \tool running on t3.medium in Figure~\ref{fig:dp_s} and Figure~\ref{fig:dp_rsn_s} when $\mathcal{S}=1,2,4,8,16$. Notably, when $\mathcal{S}=1$, all systems perform equally well in terms of accuracy due to the absence of data partitioning in Figure~\ref{fig:dp_s}. However, as $\mathcal{S}$ increases, we observe a relative decline in accuracy: 16.94\% for \tool, 23.00\% for \tool-U, and 45.08\% for \tool-C. In~\ref{fig:dp_rsn_s}, the retrained sample number of \tool decreases with a larger $\mathcal{S}$. However, those of \tool-U and \tool-C increase, aligning with systems using uniform-based and class-based data partition approaches in Figure~\ref{figs:rsn_vs_s}. This highlights the effectiveness of UCDP in maintaining higher accuracy levels despite an increasing number of shards.
Figure~\ref{fig:dp_ur} shows the impact of varying unlearning request probabilities on the number of retrained samples by running \tool, \tool-U, and \tool-C. Again, \tool demonstrates a significant advantage, reducing the number of retrained samples 
compared to \tool-U and \tool-C. This reduction highlights UCDP's efficiency in minimizing the need for extensive retraining.

These findings strongly suggest that UCDP not only surpasses traditional data partitioning methods used in exact unlearning systems but also emphasizes the importance of tailoring machine unlearning components to fit the specific needs of resource-constrained devices. By effectively managing data partitioning, \tool enhances both the accuracy and unlearning efficiency, setting a new benchmark for performance in dynamic EC scenarios.

\section{Related Work and Discussion}

 In this section, we scrutinize the related work and discuss the feasible and possible attacks against CAUSE. 
 
\noindent \textbf{Exact unlearning.}
\label{subsec:background_exact_unlearning}
Exact unlearning is a rigorous process that involves retraining a model from scratch without the unlearned data, ensuring that its impact is completely eradicated~\cite{hu2024duty}. This is achieved by either strategically partitioning the training dataset into non-overlapping shards and training separate models on each shard or by designing a model architecture that allows only parts of the model to be influenced by any given data sample. For instance, the SISA framework (Sharded, Isolated, Sliced, and Aggregated), as proposed by Bourtoule~\etal~\cite{bourtoule2021machine}, embodies the first approach, splitting the dataset into shards to allow for efficient unlearning by retraining only the relevant sub-model. In parallel, Brophy and Lowd's~\cite{brophy2021machine} data removal-enabled forests typify the second approach, enabling efficient unlearning within Random Forests. These methodologies adhere to the concept of certified data erasure, which guarantees that models from which data has been removed are indistinguishable from those never exposed to the data. While existing work~\cite{ginart2019making,schelter2021hedgecut} supports this for simpler models like linear regressions and k-nearest neighbors, it falls short for complex nonlinear models such as deep learning~\cite{izzo2021approximate}. Yan~\etal~\cite{yan2022arcane} propose a similar strategy called ARCANE, which segments datasets by class labels, trains sub-models independently, and utilizes decision aggregation. Thus, ARCANE can efficiently identify which sub-models need retraining based on the class of unlearned data. 
However, existing exact machine learning systems produce a heavy burden on memory resources, since plenty of model parameters have to be stored.

\noindent \textbf{Approximate unlearning.} 
Approximate unlearning is the process of eliminating or mitigating the impact of specific data subsets to be ``forgotten'' from a model that has already been trained, without necessitating a complete retraining of the model scratch~\cite{xu2024machine}. Approximate unlearning seeks to enhance the efficiency of the unlearning process by relaxing the strict requirements for effectiveness and certifiability, as opposed to the resource-intensive and slow process of exact unlearning~\cite{thudi2022necessity}. The strategies employed in approximate unlearning are diversified but can be primarily divided into two categories that are based on the influence function~\cite{guo2019certified,izzo2021approximate} and the gradient of the loss function~\cite{neel2021descent,wu2020deltagrad,thudi2022unrolling}. The first category~\cite{guo2019certified, izzo2021approximate} calculates the impact of the data to be forgotten on the model parameters using the influence function, allowing for the application of a Newton step to exclude this impact and adjust the model accordingly. This method's benefit lies in its simplicity, requiring only a one-step Newton update to eliminate the data's contribution. However, it faces challenges, such as the complexity of computing the inverse Hessian matrix for large-scale deep learning models and the potential inapplicability in scenarios where the training data has been removed or is unavailable. The second category~\cite{neel2021descent,wu2020deltagrad,thudi2022unrolling} focuses on determining the gradients that the data contributed during training. By reintegrating these gradients, the method aims to approximate a model that resembles one retrained from scratch, effectively ``unlearning'' the data. 
However, due to the fact that knowledge of data used in the training process may persist within trained models via existing approximate unlearning systems~\cite{maini2024tofu}, approximate unlearning is not ready to be implemented on resource-constrained devices.

\noindent \textbf{{Discussion.}}
Machine unlearning, originally developed to protect privacy and intellectual property, removes certain undesirable capabilities from general-purpose AI systems. However, as reported at the AI Seoul Summit~\cite{bengio2024international}, current approximate unlearning approaches are vulnerable to adversarial attacks~\cite{hu2024duty, hu2024learn}, and exact unlearning is infeasible due to computational overheads.

\noindent \textbf{\textit{(i)} Feasibility.} We adopt exact unlearning through CAUSE, effectively removing undesirable data from the training model. CAUSE streamlines and accelerates the unlearning process compared to other systems by incorporating UCDP and RCMP. Recognizing the constraints of limited resources on devices, CAUSE employs FiboR and SC to optimize memory usage, improving the efficiency of pinpointing the starting point for unlearning. Experimental evidence, as detailed in \S\ref{subsec:exp_speed} and \S\ref{subsec:exp_scalability}, demonstrates CAUSE's feasibility across various scales involving resource-constrained unlearning.

\noindent \textbf{\textit{(ii)} Possible attacks.} Ideally, the unlearning system should be invulnerable to machine learning attacks such as model inversion attacks or membership inference attacks concerning the unlearned data. As CAUSE is an exact unlearning system, it completely removes and re-trains all related sub-models using a dataset that excludes the unlearned data. Consequently, attackers employing model inversion or membership inference techniques cannot access the original data associated with a specific user who has requested the unlearning. Furthermore, CAUSE enhances the protection of retained data by integrating RCMP, utilizing pruning techniques that are shown to improve data privacy against inversion and membership inference attacks~\cite{chen2018shallowing, yuan2022membership}.

\section{Conclusion}
\label{sec:conclusion}

Machine unlearning on resource-constrained devices continues to become increasingly relevant in light of privacy concerns. In this work, we introduce \tool, a pioneering exact unlearning system specifically for resource-constrained devices such as satellites, edge and mobile devices. \tool consists of four key components, including resource-constrained model pruning, user-centric data partitioning, Fibonacci-based replacement, and shard controller, to enable efficient exact unlearning on resource-constrained devices. Comprehensive experiments are conducted based on three widely-used datasets and we compare \tool against three benchmark exact unlearning systems. Our findings demonstrate that \tool facilitates efficient unlearning and surpasses other systems in accuracy, unlearning speed, and energy consumption. We are confident that \tool can provide valuable insights for advancing future edge unlearning frameworks.

\bibliographystyle{IEEEtranS}
\bibliography{main}

\appendices

\section{Model Pruning Details}
\label{apx:pruning_settings}

Table~\ref{tab:exp_pruning} shows the experimental settings of model pruning experiments in~\ref{subsec:pruning}, where TE is training epochs and RE is retraining epochs. Iterative pruning is selected as the default pruning method, the same as the workflow in Figure~\ref{fig:pruning}.

\begin{table}[ht]
\caption{Experiment Settings (Pruning)}
\label{tab:exp_pruning}
\footnotesize
\centering
\resizebox{\columnwidth}{!}{
\begin{tabular}{cccccccc}
\toprule
\textbf{Baseline} & \textbf{Dataset} & \textbf{GPU Model} & \textbf{Optimizer} & \textbf{Batch Size} & \textbf{TE} & \textbf{RE} \\
\midrule
VGG-16       & CIFAR-10  &      A40 &     SGD & 128 & 150 & 30 \\
ResNet-34    & CIFAR-10  &      A40 &    Adam &  80 &  50 & 20 \\
DenseNet-121 & CIFAR-100 &      A40 & RMSprop & 128 & 100 & 20 \\
MobileNetV2        & CIFAR-10 & RTX 4090 &   Adam & 128 & 50 & 20 \\
\bottomrule
\end{tabular}
}
\vspace{-1 em}
\label{table:parameter_settings}
\end{table}

\section{System Settings}
\label{apx:settings}

The settings of various exact unlearning systems can be found in Table~\ref{tab:model_training_and_pruning_details}.

\begin{table}[ht]
\caption{Training and Pruning Configurations}
\label{tab:model_training_and_pruning_details}
\resizebox{\columnwidth}{!}{%
\begin{tabular}{l|c|c|c|c|c}
\toprule
\textbf{Baselines} & \textbf{\tool} & \textbf{SISA} & \textbf{ARCANE} & \textbf{OMP-70} & \textbf{OMP-95}\\
\midrule
\multirow{2}{*}{Optimizer} & \multirow{2}{*}{Adam} & \multirow{2}{*}{Adam} & SGD  & \multirow{2}{*}{Adam} & \multirow{2}{*}{Adam} \\
 &  &  &  (Momentum 0.9) &  &  \\
\midrule
Initial learning rate & 0.001 & 0.001 & 0.001 & 0.001 & 0.001 \\
\midrule
Pruning method & Iterative & N/A & N/A & One-shot & One-shot \\
\midrule
Pruning ratio & 70\% & N/A & N/A & 70\% & 95\% \\
\bottomrule
\end{tabular}
}
\end{table}

\end{document}